\documentclass[10pt,journal,compsoc]{IEEEtran}
\ifCLASSOPTIONcompsoc
  \usepackage[nocompress]{cite}
\else
  \usepackage{cite}
\fi

\usepackage{epsfig}
\usepackage{graphicx}
\usepackage{amsmath}
\usepackage{amssymb}

\usepackage{enumitem}
\usepackage{multirow}
\usepackage{bbding}
\usepackage{pifont}
\usepackage{wasysym}
\usepackage{tikz}
\usepackage{float}

\ifCLASSINFOpdf
\else
\fi
\def\etal{\emph{et al.}}

\begin{document}
%\title{Bare Demo of IEEEtran.cls for\\ IEEE Computer Society Journals}

% \title{Cross-domain Detection : A Review}
% \title{Unsupervised Domain Adaptive \\ Detection : A Survey}
\title{Unsupervised Domain Adaptation of \\Object Detectors: A Survey}

%\author{Poojan Oza, Vishwanath Sindagi, Vibashan VS and Vishal M. Patel \thanks{Poojan Oza, Vishwanath A. Sindagi, Vibashan VS and Vishal M. Patel are with the department of Electrical and Computer Engineering (ECE) at Johns Hopkins University, Baltimore, MD - 21218. Email-id : \{poza2, vishwanathsindagi, vvishnu2, vpatel36\}@jhu.edu.}
%}

\author{Poojan~Oza,~\IEEEmembership{Member,~IEEE,}
        Vishwanath~A.~Sindagi,~\IEEEmembership{Member,~IEEE,}
        Vibashan~VS,~\IEEEmembership{Member,~IEEE,}
        and~Vishal~M.~Patel,~\IEEEmembership{Senior Member,~IEEE}% <-this % stops a space
\IEEEcompsocitemizethanks{\IEEEcompsocthanksitem Poojan Oza, Vishwanath Sindagi, Vibashan VS, and Vishal M. Patel are with the department of Electrical and Computer Engineering (ECE) at Johns Hopkins University, Baltimore, MD - 21218. E-mail: \{poza2, vishwanathsindagi, vvishnu2, vpatel36\}@jhu.edu.\protect}% <-this % stops an unwanted space
\thanks{Manuscript received Month XX, XXXX; revised Month XX, XXXX.}}


\markboth{Journal of \LaTeX\ Class Files,~Vol.~XX, No.~X, Month~XXXX}%
{Shell \MakeLowercase{\textit{et al.}}: Bare Demo of IEEEtran.cls for Computer Society Journals}



\IEEEtitleabstractindextext{
\begin{abstract}
Recent advances in deep learning have led to the development of accurate and efficient models for various computer vision applications such as classification, segmentation, and detection. However, learning highly accurate models relies on the availability of large-scale annotated datasets. Due to this, model performance drops drastically when evaluated on label-scarce datasets having visually distinct images, termed as domain adaptation problem. There are a plethora of works to adapt classification and segmentation models to label-scarce target dataset through unsupervised domain adaptation. Considering that detection is a fundamental task in computer vision, many recent works have focused on developing novel domain adaptive detection techniques. Here, we describe in detail the domain adaptation problem for detection and present an extensive survey of the various methods. Furthermore, we highlight strategies proposed and the associated shortcomings. Subsequently, we identify multiple aspects of the problem that are most promising for future research. We believe that this survey shall be valuable to  the pattern recognition experts working in the fields of computer vision, biometrics, medical imaging, and autonomous navigation by introducing them to the problem, and familiarizing them with the current status of the progress while providing  promising directions for future research.
\end{abstract}

% Note that keywords are not normally used for peer review papers.
}

\maketitle

\IEEEdisplaynontitleabstractindextext

% \fi
\IEEEpeerreviewmaketitle

\IEEEraisesectionheading{\section{Introduction}\label{sec:introduction}}

\IEEEPARstart{T}{he} success of deep learning has been greatly beneficial for various fields such as natural language processing \cite{vaswani2017attention}, \cite{devlin2019bert}, \cite{brown2020language}, robotics \cite{mnih2013playing}, \cite{schrittwieser2019mastering}, \cite{xiao2019thinking}, computer vision \cite{lecun1998gradient}, \cite{krizhevsky2012imagenet}, \cite{he2016deep}, etc. This is especially evident in the case of computer vision, where majority of the progress can be largely attributed to the advancements in deep convolutional neural networks (DCNN) \cite{lecun1998gradient}. Owing to their learning capacity, DCNN models have achieved state-of-the-art performance in many vision tasks such as object classification (\cite{he2016deep}, \cite{huang2017densely}, \cite{hu2018squeeze}), semantic segmentation (\cite{long2015fully}, \cite{zhao2017pyramid}, \cite{chen2017deeplab}), and object detection (\cite{ren2015faster}, \cite{redmon2016you}, \cite{liu2016ssd}). This has led to DCNN's increased popularity in several real world applications as compared to the classical computer vision techniques. Specifically, deep learning based object detection has become an integral part of many real-world applications ranging from video security/surveillance, augmented reality, autonomous navigation, human computer interface, self-checkout convenience stores. Major advancements like Faster-RCNN \cite{ren2015faster}, You Only Look Once (YOLO) \cite{redmon2016you} and Single Shot Multi-box Detector (SSD) \cite{liu2016ssd} have resulted in significant improvements of detection performance and speed.

\begin{figure*}[ht!]
	\begin{center}
		\includegraphics[width=0.85\linewidth]{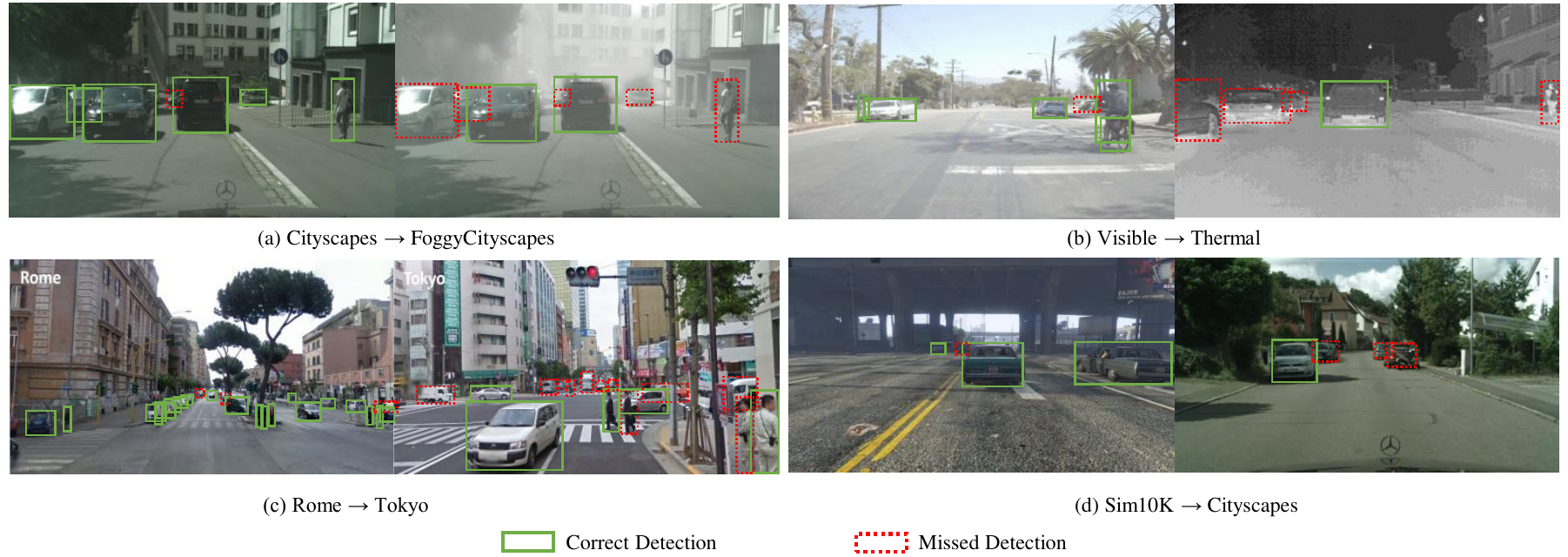}
	\end{center}
	\vskip -15.0pt
	\caption{Visualization of object detection results. Left: Source trained model on source domain, Right: Source trained model on the target domain. (a) The model, trained on Cityscapes dataset, performs well on the Cityscapes images.  However, it fails to detect multiple cars when evaluated on FoggyCityscapes images, which has a domain shift due to fog. (b) Under visible to thermal domain shift, the model fails to detect person and cars in the thermal domain. (c) A detection model trained in Rome, when evaluated on another city such as Tokyo undergoes drastic performance reduction due to differenences in  scene appearances, weather, objects, etc. (d) In the case of Sim10K to Cityscapes, multiple cars are missed in the Cityscapes domain as the model was trained on images captured from the simulated virtual world. These examples show that the detection models  generalize poorly under the domain shift/dataset distribution shift.}
	\label{fig:intro_fig} 
\end{figure*}

\begin{figure}[ht!]
	\begin{center}
		\includegraphics[width=0.65\linewidth]{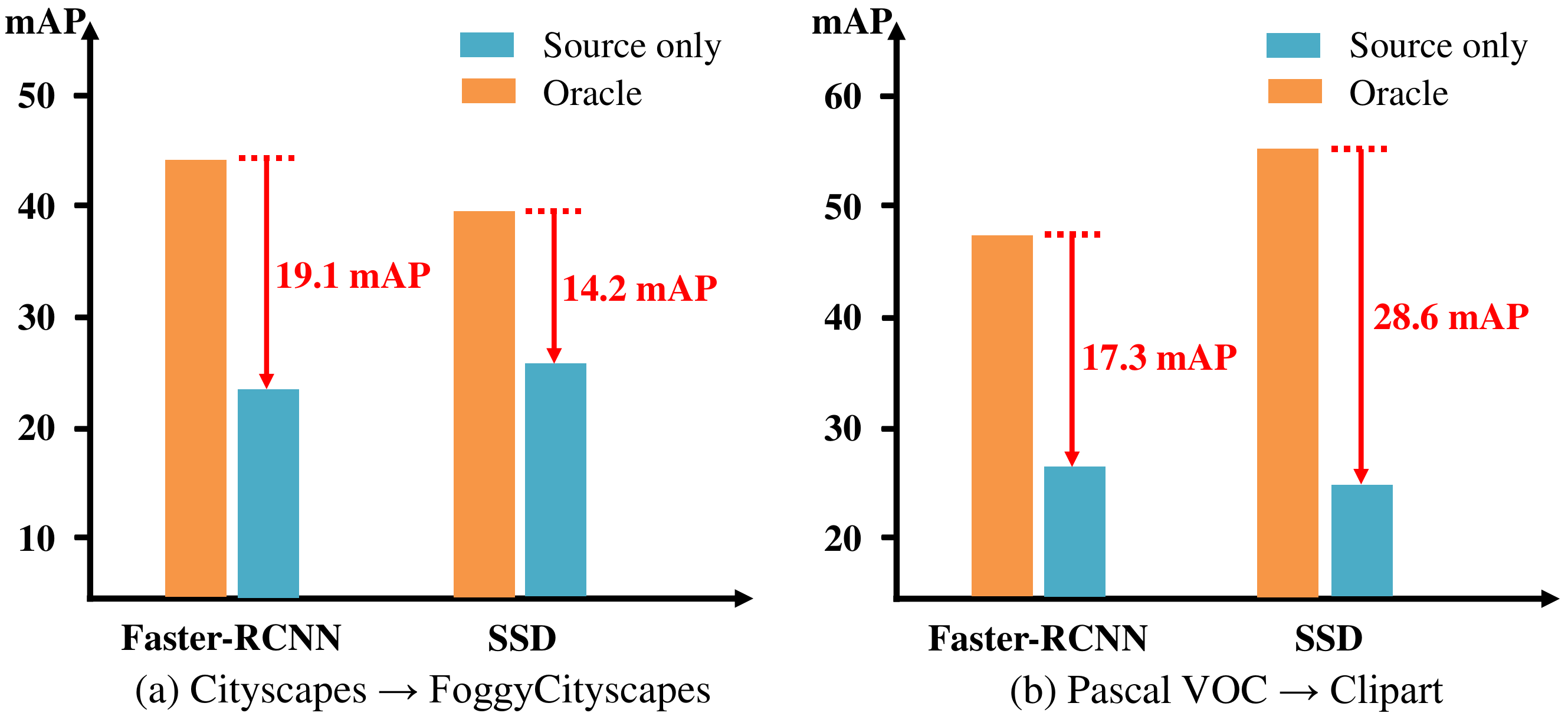}
	\end{center}
	\vskip -15.0pt
	\caption{ Illustration of detector performance; In (a), the model is trained on Cityscapes and evaluated on FoggyCityscapes and in (b), the model is trained on Pascal VOC and evaluated on Clipart. We can observe a significant drop in the performance of the detector when there is a distribution shift in the training and test data.}
	\label{fig:src_oracle}
\end{figure}

It is important to note that most DCNN models need to be trained in a supervised fashion, which has been made possible due to the availability of large datasets having thousands of images annotated with ground-truth labels \cite{deng2009imagenet}, \cite{everingham2010pascal}, \cite{lin2014microsoft}. However, one of the major drawbacks is the poor generalization capability of DCNN models to visually distinct images compared to the training images. For instance, a detection model trained with a dataset collected in Rome may not necessarily perform well on images from Tokyo due to the changes in the appearance of scenes/objects and/or weather between them, as illustrated in Fig.~\ref{fig:intro_fig}(c). A similar example is shown for cases such as sunny to foggy weather (Fig.~\ref{fig:intro_fig}(a)), visible to thermal (Fig.~\ref{fig:intro_fig}(b)), and synthetic to real-world (Fig.~\ref{fig:intro_fig}(d)). Fig.~\ref{fig:src_oracle} shows quantitatively the performance drop of different deep learning based object detectors that are trained on one particular dataset, when evaluated on different datasets. This problem where models, trained on one particular dataset (also known as source dataset), do not generalize well to a dataset that has a different distribution (also known as target dataset) is commonly referred to as \textit{domain shift} or \textit{distribution shift} in the literature \cite{ben2010theory}, \cite{DA_Patel_SPM15}, \cite{ganin2016domain}.  

A straightforward approach to solving this distributional shift problem is annotating the target dataset images with ground-truth detection labels. However, this might prove to be infeasible considering that the labor cost of the annotation process is prohibitively expensive for all visually distinct conditions. To circumvent this issue, many methods rely on the principles of unsupervised domain adaptation \cite{ben2010theory}, \cite{DA_Patel_SPM15} which involves training the DCNN model with both label-rich source dataset and label-scarce target dataset having visually distinct appearance. Techniques \cite{ganin2016domain}, \cite{hoffman2016fcns}, \cite{chen2018domain} for domain adaptive training with source and unlabeled target datasets have demonstrated improved generalization capabilities, resulting in improved performance on the visually distinct target domain. The unsupervised domain adaptation has been extensively studied for the task of classification \cite{shrivastava2014unsupervised}, \cite{sun2016return}, \cite{ganin2016domain}, \cite{saito2017asymmetric}, \cite{hoffman2018cycada}, \cite{tzeng2017adversarial}, \cite{saito2018maximum}, \cite{lee2019sliced}, \cite{chang2019domain}, \cite{roy2019unsupervised}, \cite{long2016unsupervised}, \cite{saito2018adversarial}, \cite{lee2019drop}, \cite{murez2018image}, \cite{sun2019unsupervised}, \cite{volpi2018adversarial}, \cite{hu2018duplex}, \cite{kurmi2019attending}, and semantic segmentation \cite{hoffman2016fcns}, \cite{tsai2018learning}, \cite{chen2017no}, \cite{sankaranarayanan2018learning}, \cite{zou2018unsupervised}, \cite{chen2018road}, \cite{luo2019taking}, \cite{tsai2019domain}, \cite{licontent}, \cite{tsai2019domain}, \cite{du2019ssf}, \cite{chang2019all}, \cite{li2019bidirectional}, \cite{paul2020domain}, \cite{zou2018unsupervised}, \cite{hong2018conditional}, \cite{zhang2021prototypical}. 

\begin{figure*}[ht!]
	\begin{center}
		\includegraphics[width=0.95\linewidth]{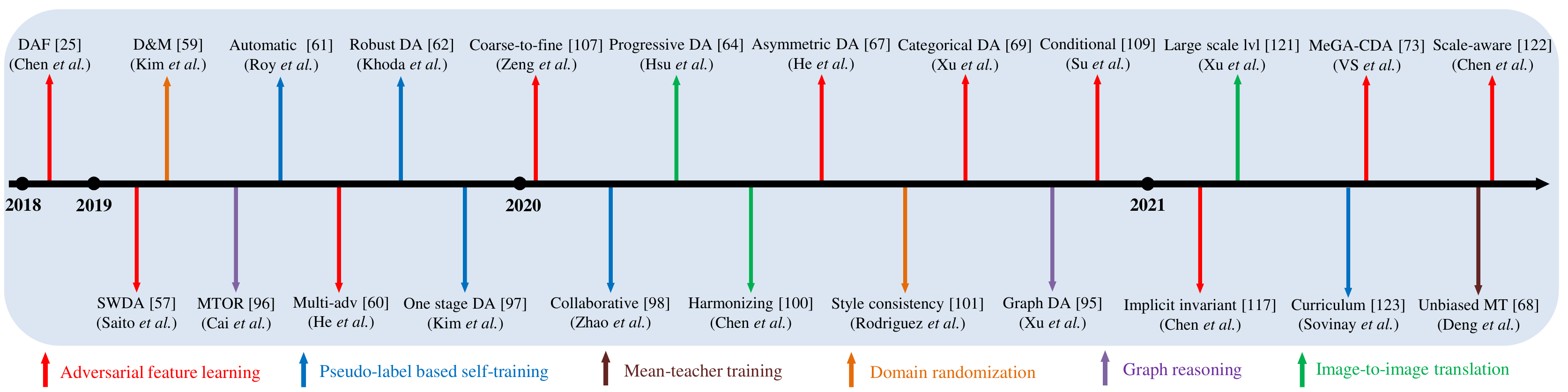}
	\end{center}
	\vskip -15.0pt
\caption{A timeline of the key papers related to the domain adaptation of object detectors published over the recent years. These methods are broadly categorized into six classes: Adversarial feature learning, Pseudo-label based self-training, Graph-reasoning, Image-to-image translation, Domain randomization and Mean-teacher training. }
\label{fig:timeline} 
\end{figure*}

However, unlike classification (image-level prediction) and semantic segmentation (pixel-level prediction), the object detection task involves bounding-box localization and the bounding-box-level category prediction task. This poses unique challenges while addressing unsupervised domain adaptation for object detection models. This has sparked an interest in addressing unsupervised domain adaptation for object detection with many novel approaches proposed very recently \cite{chen2018domain}, \cite{saito2019strong}, \cite{zhu2019adapting}, \cite{kim2019diversify}, \cite{he2019multi}, \cite{roychowdhury2019automatic}, \cite{khodabandeh2019robust}, \cite{xie2019multi}, \cite{hsu2020progressive}, \cite{tzeng2018splat}, \cite{zhuang2020ifan}, \cite{he2020domain}, \cite{deng2021unbiased}, \cite{xu2020exploring}, \cite{pan2020multi}, \cite{hsu2020every}, \cite{sindagi2020prior}, \cite{vs2021mega}, \cite{li2020free}, \cite{liang2020domain}, \cite{guan2021uncertainty}, \cite{sun2021multi}, \cite{zhang2021detr}, \cite{scheck2020unsupervised}, \cite{yang2020channel}. A timeline of some of the key papers proposed in the recent past is shown in  Fig.~\ref{fig:timeline}.

While there exist multiple survey papers that extensively review domain adaptation techniques both classification \cite{DA_Patel_SPM15}, \cite{wang2018deep}, \cite{zhuang2020comprehensive} and semantic segmentation \cite{toldo2020unsupervised}, \cite{zhao2020review}, to the best of our knowledge,  there is no comprehensive survey of unsupervised domain adaptation of object detectors. Although Li \etal \cite{li2020deep} attempt to review some of the  domain adaptive object detection literature, their discussions are limited and lack a comprehensive comparison of existing methods. This motivates us to present a comprehensive literature analysis of all the domain adaptive object detection methods proposed in the past few years, along with detailed discussions and comprehensive comparison. The major contributions of this work are  summarized as follows:
\begin{enumerate}
	\item As opposed to the existing survey\cite{li2020deep}, we provide a more detailed and thorough discussion on all the existing works (to the best of our knowledge)  on domain adaptive object detection. We also define a taxonomy of the various works in literature. Furthermore, we discuss the preliminaries of the relevant topics like object detection and domain adaptation in an attempt to make the paper self-sufficient and useful to the readers who are not familiar with  these concepts.
	\item We present a comprehensive comparison of the existing methods on all publicly available datasets used in the literature for unsupervised domain adaptive object detection with thorough discussions on the methods and detailed comparison of the performances along with their respective experimental settings.
	\item Finally, we identify potential research directions that might prove beneficial for the researchers working in the area in order to further advance the state-of-the-art.
\end{enumerate}

\section{Preliminaries}\label{sec:prelim}
In this section, we provide an introduction to two of the important aspects related to domain adaptive object detection, i.e., object detection  and unsupervised domain adaptation. We formally set up the problem and notations that are used throughout the paper.

\subsection{Object detection}\label{subsec:od}
Over the years, deep convolutional neural network based object detectors have demonstrated exceptional improvements in performance on a variety of datasets  and have become an integral part of various computer vision applications. There are a variety of surveys \cite{liu2020deep}, \cite{zhao2019object}, \cite{zou2019object} on the topic covering wide range of techniques proposed over the past decade for object detection. The most popular frameworks for object detection are Faster-RCNN \cite{ren2015faster}, You Only Look Once (YOLO) \cite{redmon2016you} and Single Shot Multi-box Detector (SSD) \cite{liu2016ssd}. The majority of domain adaptive object detection works are based on the Faster-RCNN and a few others use SSD. Other recent frameworks include, Fully Convolutional One Stage (FCOS) Object Detection \cite{tian2019fcos} and DEtection TRansformer (DETR) \cite{carion2020end}. However, these frameworks have been only scarcely used for the domain adaptive object detectors. In what follows, we briefly describe the widely used detection frameworks in the domain adaptive detection literature, i.e., Faster-RCNN and SSD.
%Since majority of works in domain adaptive object detection rely on Faster-RCNN as their base framework, in this section we will also focus on brief overview for the same. However, we do note that YOLO and SSD frameworks are closely similar to Faster-RCNN with minor design changes made to make the model more time-efficient during inference.

\begin{figure}[b!]
	\begin{center}
		\includegraphics[width=0.95\linewidth]{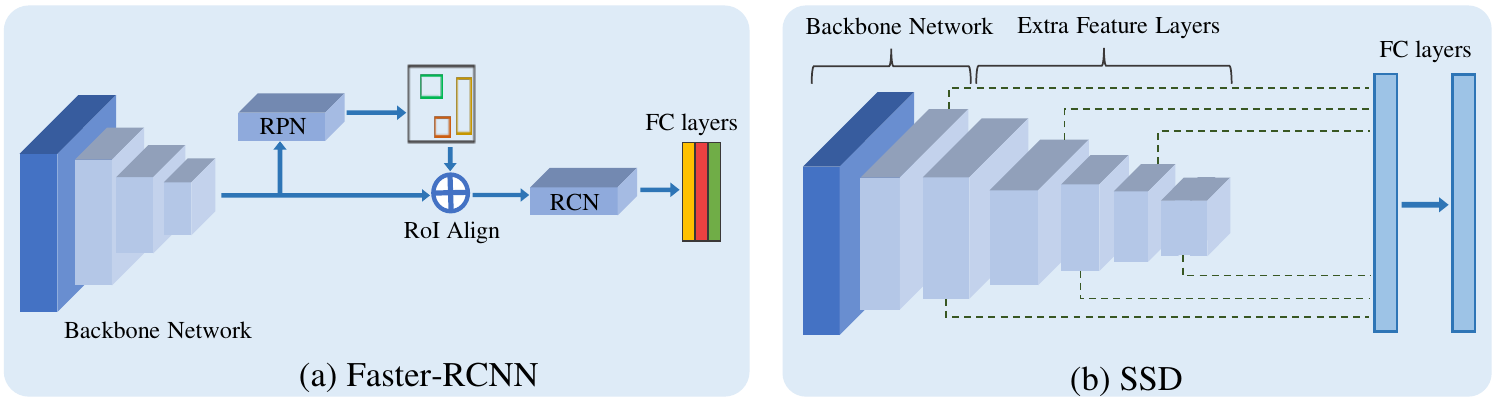}
	\end{center}
	\vskip -15.0pt
	\caption{ Illustration of popular detection frameworks: (a) Faster-RCNN \cite{ren2015faster}, (b) Single Shot Multi-Box Detector (SSD) \cite{liu2016ssd}.}
	\label{fig:det_frameworks} 
\end{figure}

\subsubsection{Faster-RCNN}\label{subsubsec:frcnn}
The Faster-RCNN framework, proposed by Ren \etal \cite{ren2015faster}, follows a two-stage object detection approach and it  consists of three major components: \textit{1)} a backbone CNN, \textit{2)} a Region Proposal Network (RPN), and \textit{3)} a Region-of-Interest (RoI) based classifier (RCN). Fig.~\ref{fig:det_frameworks}(a) shows an overview of the Faster-RCNN architecture. Consider a dataset, $\mathcal{D}=\{X^i, Y^i\}_{i=1}^{N}$, having $N$ images, with each image $X^i$ with ground-truth annotation $Y^i$. Here, the ground-truth annotation $Y^i$ denotes both bounding boxes and respective object categories in the corresponding image $X^i$. As shown in Fig.~\ref{fig:det_frameworks}(a), an input image ($X^i$) is forwarded through the backbone network resulting in a set of feature maps. These feature maps are then fed to the RPN network which generates a set of candidate object proposals. The RPN network relies on  pre-defined anchor boxes of multiple sizes and aspect ratios in order to effectively learn to generate the candidate proposals. Subsequently, each proposal is then transformed into fixed-size features using RoI-pooling. Finally, the pooled features are then forwarded through the RCN, which predicts the category label for each candidate proposal in addition to refining its bounding box. For training the RPN candidate, a category-agnostic binary label (of being an object or not) is assigned to each anchor. The $j^{th}$ anchor is assigned a label, denoted as $y^j_b \in \{0, 1\}$, as positive (or $1$) if it has the highest Intersection over Union (IoU) overlap with one of the ground-truth boxes or if it has an IoU overlap higher than 0.7 with any of the ground-truth boxes in the corresponding image. Similarly, a negative label (or $0$) is assigned to the anchor if IoU ratio is lower than 0.3 for all ground-truth boxes. The RPN is then tasked to perform a binary classification to identify whether the candidate bounding box proposal corresponds to one of the objects in the image and learn the offset between the ground-truth bounding box, denoted as $\textbf{b}^j$, and respective anchor box to get final bounding box prediction, denoted as $\tilde{\textbf{b}}^j$. The offset learning is supervised with the help of a regression loss applied on the bounding box parameters. Both these losses are combined together to obtain the final loss for region proposal network as shown below:
\setlength{\belowdisplayskip}{1pt} \setlength{\belowdisplayshortskip}{1pt}
\setlength{\abovedisplayskip}{1pt} \setlength{\abovedisplayshortskip}{1pt}
\begin{equation}\label{eq:rpn}
\begin{split}
\mathcal{L}_{rpn} \ = \ \frac{1}{N_{b}} \sum_{j} &\mathcal{L}^{bce}_{rpn}(y^j_b, \ p^j_b) \\
& + \ \lambda_{rpn} \frac{1}{N_{bbox}} \sum_{j} {p}^{j}_b \mathcal{L}^{reg}_{rpn}(\textbf{b}^j, \ \tilde{\textbf{b}}^{j}),
\end{split}
\end{equation}
where $j$ is the index of an anchor box in the mini-batch and $p^j_b$ is the probability assigned to the respective anchor box being an object. The loss, $\mathcal{L}_{rpn}^{reg}$, computes the smooth-L1 distance between the given ground truth bounding box and the predicted bounding box $\tilde{\textbf{b}}^j$. Both $\textbf{b}^j$ and $\tilde{\textbf{b}}^j$ are vectors having four bounding box parameters, namely center x-coordinate, center y-coordinate, height and width to represent a bounding box. Also, $bce$ denotes binary cross entropy loss and $reg$ denotes regression loss, which is smooth L1 loss for Faster-RCNN \cite{ren2015faster}. Here, $\mathcal{L}^{bce}_{rpn}$ is normalized with the size of mini-batch and $\mathcal{L}^{reg}_{rpn}$ is normalized with number of bounding box locations $N_{bbox}$.
% \textcolor{red}{Here, $\lambda_{rpn}$ is introduced to balance the $RPN$ regression and classification losses. The offset to obtain final bounding box prediction is calculated by matching the predicted offset with difference between anchor box parameters and ground-truth bounding box parameters. These difference and offset are calculated as:}
% \begin{equation}\label{eq:bbox_offset}
% \begin{aligned}
% \Delta b_{\mathrm{x}} &=\left(x-x_{\mathrm{a}}\right) / w_{\mathrm{a}}, \quad \Delta b_{\mathrm{y}}=\left(y-y_{\mathrm{a}}\right) / h_{\mathrm{a}}, \\
% \Delta b_{\mathrm{w}} &=\log \left(w / w_{\mathrm{a}}\right), \quad \Delta b_{\mathrm{h}}=\log \left(h / h_{\mathrm{a}}\right), \\
% \Delta \tilde{b}_{\mathrm{x}} &=\left(\tilde{x}-x_{\mathrm{a}}\right) / w_{\mathrm{a}}, \quad \Delta \tilde{b}_{\mathrm{y}}=\left(\tilde{y}-y_{\mathrm{a}}\right) / h_{\mathrm{a}}, \\
% \Delta \tilde{b}_{\mathrm{w}} &=\log \left(\tilde{w} / w_{\mathrm{a}}\right), \quad \Delta \tilde{b}_{\mathrm{h}}=\log \left(\tilde{h} / h_{\mathrm{a}}\right),
% \end{aligned}
% \end{equation}
% where $\Delta b_{\mathrm{x}}, \Delta b_{\mathrm{y}}, \Delta b_{\mathrm{h}}, \Delta b_{\mathrm{w}}$ denotes the difference of anchor box parameters $x_a, y_a, h_a, w_a$ and ground-truth bounding box parameters, $x, y, h, w$. Similarly, $\Delta \tilde{b}_{\mathrm{x}}, \Delta \tilde{b}_{\mathrm{y}}, \Delta \tilde{b}_{\mathrm{h}}, \Delta \tilde{b}_{\mathrm{w}}$ denotes the predicted offset to get final bonding box prediction parameters, $\tilde{x},\tilde{y},\tilde{h},\tilde{w}$.

Next, the RCN network is trained to perform classification of RoI-pooled features using cross entropy loss with $K+1$ class classification, denoted as $\mathcal{L}^{ce}_{rcn}$. Here, $K$ denotes the number of categories in the dataset and an additional class to represent the background category. Additionally, the RCN is also tasked to predict the bounding box offset through regression loss similar to the RPN network
\begin{equation}\label{eq:rcn}
\begin{split}
\mathcal{L}_{rcn} \ = \ \frac{1}{N_{b}} \sum_{j} &\mathcal{L}^{ce}_{rcn}(y^j_c, \ \textbf{p}^j) \\
& + \ \lambda_{rcn} \frac{1}{N_{bbox}} \sum_{j} {p}^{j}_b \mathcal{L}^{reg}_{rcn}(\textbf{b}^j, \ \tilde{\textbf{b}}^{j}),
\end{split}
\end{equation}
where  $y^j_c \in \{1,\dots,K+1\}$ denotes the ground-truth category label of $j^{th}$ RoI-pooled feature and $\textbf{p}^j$ is the predicted probability vector denoting probabilities assigned for all $K+1$ categories. The loss $\mathcal{L}^{ce}_{rcn}$ denotes cross entropy loss and $\mathcal{L}^{reg}_{rcn}$ loss is same as the one used for the RPN network. 

The overall loss function used to train the entire Faster-RCNN network  is trained is defined as:
\begin{equation}\label{eq:loss_det}
\mathcal{L}^{frcnn}_{det} \ = \ \mathcal{L}_{rpn} \ + \ \mathcal{L}_{rcn}.
\end{equation}

More details regarding the anchor boxes, bounding box regression losses, training procedure, and architecture can be found in the \cite{ren2015faster}.

\subsubsection{Single Shot Multi-Box Detector (SSD)}\label{subsubsec:ssd}

Liu \etal \cite{liu2016ssd} proposed a single shot object detection framework which consists of forwarding the image through a single stage as opposed to two stages in the Faster-RCNN detector. Fig.~\ref{fig:det_frameworks}(b) illustrates the SSD detection architecture. By following this approach, SSD eliminates the need for an object proposal stage, making it simpler and computationally efficient as compared to the Faster-RCNN approach. The SSD framework employs  VGG16 as the backbone network which is used for extracting feature map of size $H\times W$ from an input image $X$. For each feature map location, SSD discretizes the output space of the bounding boxes into a set of default bounding boxes. A convolutional layer is added that for each feature map location predicts a score for a category or offsets relative to the default box coordinates. The set of default boxes contain bounding boxes of multiple pre-defined aspect ratios and scales to match any object shape in the image better. Furthermore, SSD combines predictions from feature maps at multiple scales to better handle the object scales with respect to the image. 

Once the model predictions are available, they are matched with the ground-truth box and category to preform an end-to-end training with regression and classification loss. The regression loss used in SSD is a smooth L1 loss, denoted here as $L_1^{s}$. For each location in the feature map, a default box is matched with a ground-truth bounding box. The final bounding box prediction is computed by adding the predicted offset to the default boxes and the regression loss is computed to correct the offsets based on the matched ground-truth bounding box. The matching strategy is to find default box which has best jaccard overlap with the ground-truth bounding box and then matching default boxes any ground-truth having jaccard overlap higher than 0.5. For a given $i^{th}$ predicted bounding box $\tilde{\textbf{b}}^i$ and matched $j^{th}$ ground-truth bounding box $\textbf{b}^j$, the corresponding label is defined as $y^{ij}\in \{0, 1\} = 1$, the regression loss is given as:
\begin{equation}\label{eq:ssd_reg}
\begin{array}{c}
\mathcal{L}_{reg}=\sum\limits_{i=1}^{HW} \sum\limits_{j=1}^{N_b} y^{ij} L_1^{s}(\tilde{\textbf{b}}^i, \textbf{b}^j), \\
\end{array}
\end{equation}
where $N_b$ denotes number of ground-truth bounding box per image. Both $\tilde{\textbf{b}}^i, \textbf{b}^j$ are bounding box vector having center $x-y$ location and height $(h)$ and width $(w)$. For each predicted bounding box, the classification loss is computed over $K+1$ categories as shown below:
\begin{equation}\label{eq:ssd_cls}
\mathcal{L}_{cls}=-\sum\limits_i \sum\limits_{c=1}^{K+1} \textbf{y}^i_c \log(\textbf{p}^i_c),
\end{equation}
where $\textbf{y}^i_c\in\{1,\dots,K+1\}$ denotes one-hot vector indicating the category label respective predicted bounding boxes and $\textbf{p}^i_c$ is the corresponding prediction probability vector. Specifically, $\textbf{p}^i_c$ denotes the probability of $i^{th}$ bounding box belonging to $c^{th}$ category. As earlier, there are $K$ categories in the dataset and $K+1^{th}$ label denotes the background class. The final detection loss is a combination of both regression and classification losses and is defined as follows:
\begin{equation}\label{eq:ssd_det}
\mathcal{L}^{ssd}_{det} \ = \ \mathcal{L}_{reg} \ + \ \mathcal{L}_{cls}.
\end{equation}
In the case where there are no predicted bounding boxes that can be matched with one of the ground-truth bounding boxes, the regression loss is set to zero. More details regarding the default boxes, box matching algorithm bounding box regression losses, training procedure, and architecture details can be found in \cite{liu2016ssd}.

%%%%%%%%%%%%%%%%%%%%%%%%%%%%%%%%%%%%%%%%%%%%%%%%%%%%%%%%%%%%%%%%% Block comment

\subsection{Domain adaptation}\label{subsec:da}

In the domain adaptation problem, we consider two domains, namely source and target, denoted as $\mathcal{S}$ and $\mathcal{T}$, respectively. The source and target domains are assumed to have different data distributions, i.e., $\text{P}_{\mathcal{S}} \neq \text{P}_{\mathcal{T}}$. Most domain adaptation formulations consider that the source dataset is label-rich, while the target dataset is label-scarce in nature \cite{DA_Patel_SPM15}. Multiple variations of this formulation that are commonly studied in the literature include, \textit{semi-supervised} \cite{daume2010frustratingly, saito2019semi, xu2019d}, \textit{weakly-supervised} \cite{inoue2018cross}, and \textit{unsupervised} domain adaptation \cite{ben2010theory, chen2018domain, saito2019strong, xu2020cross, cai2019exploring, vs2021mega, sindagi2020prior}. In the context of object detection, the \textit{semi-supervised domain adaptation} formulation assumes that source domain is fully labeled with bounding box annotations and corresponding category labels and only a subset of the target domain samples are fully annotated with bounding box and respective category labels. \textit{Weakly-supervised domain adaptation} formulation assumes that source domain is fully annotated and all target domain samples have binary annotations indicating the presence/absence of any category and no bounding box annotations. Lastly, the \textit{unsupervised domain adaptation} formulation assumes that source domain is fully annotated while no annotations are available for target domain. Among these formulations, the unsupervised formulation is more practical and challenging. Further, the solutions obtained for this formulation can be easily adopted to address the semi-supervised and weakly-supervised domain adaptation tasks as well. For these reasons, we mainly focus on reviewing works that address unsupervised domain adaptation for object detection. In what follows, we formally define the unsupervised domain adaptation formulation and provide a brief overview of the same.

\subsubsection{Unsupervised Domain adaptation}\label{subsubsec:uda}
Let us denote the source dataset as, $ \ \mathcal{S}=\{X^i_s, Y^i_s\}_{i=1}^{N_s}$, and it consists of $N_s$ number of images. Here, $X^i_s$ denotes $i^{th}$ image and $Y^i_s$ denotes the corresponding bounding box annotations with category label. Similarly, let us denote the target dataset as, $\mathcal{T}=\{X^i_t\}_{i=1}^{N_t}$ having $N_t$ number of target domain images with no ground-truth annotations. Ben \etal \cite{ben2010theory} proposed a framework to perform domain adaptation for the given setup, i.e., labeled source dataset and unlabeled target dataset, with theoretical upper bounds on the target performance. Ben \etal \cite{ben2010theory} designed a $\mathcal{H}\Delta\mathcal{H}$-distance to measure the divergence between two sets of samples that have different data distributions, as is the case for the domain adaptation problem. Let us consider an arbitrary source domain image $X_s \in \mathcal{S}$ and an arbitrary target domain image $X_t \in \mathcal{T}$. Furthermore, let us consider a domain discriminator denoted as, $D : X \rightarrow \{0, 1\}$, that takes in any image $X \in \{\mathcal{S} \cup \mathcal{T}\}$ and predicts the domain of the input image. classifies source domain image $X_s \in \mathcal{S}$ as label $0$, and target domain image $X_t \in \mathcal{T}$ as label $1$. Considering $\mathcal{H}$ to be a set of possible domain discriminators, the $\mathcal{H}\Delta\mathcal{H}$-distance can be defined as follows:
\begin{equation}\label{eq:h_div}
\begin{split}
d_{\mathcal{H}\Delta\mathcal{H}}(\mathcal{S}, \mathcal{T})= &2 \underset{(D, D') \in \mathcal{H}^2}{\operatorname{sup}} \Big| \underset{X \sim \mathcal{S}}{\mathbf{E}}[D(X) \neq D'(X)] \\
& - \underset{X \sim \mathcal{T}}{\mathbf{E}}[D(X) \neq D'(X))] \Big|,
\end{split}
\end{equation}
where $\mathbf{E}_{X \sim \mathcal{S}}$ and $\mathbf{E}_{X \sim \mathcal{T}}$ denotes the expected domain classification errors over the source and target domain dataset, respectively. More precisely, the Eq.~\ref{eq:h_div} measures the divergence by the disagreement of the hypothesis sampled from $\mathcal{H}$. The ideal joint hypothesis is defined as:
\begin{equation}\label{eq:joint_hyp}
D^* = \underset{D \in \mathcal{H}}{\operatorname{argmin}} \operatorname{Err}_{\mathcal{S}}(D^*)+\operatorname{Err}_\mathcal{T}(D^*). 
\end{equation}

Here, the terms $\operatorname{Err}_{\mathcal{S}}$ and $\operatorname{Err}_{\mathcal{T}}$ denote the expected prediction errors on the source and target domain data samples, respectively. This distance is often used in the domain adaptation literature to measure the adaptability between any give source and target domain datasets. Ben \etal \cite{ben2010theory} present a theorem that further defines the upper bound on the target error as:
\begin{equation}\label{eq:upper_bound}
\begin{split}
\forall D \in \mathcal{H}, \ &\operatorname{Err}_{\mathcal{T}}(D) \leq \\
&\operatorname{Err}_{\mathcal{S}}(D) + \frac{1}{2} d_{\mathcal{H}\Delta\mathcal{H}}(\mathcal{S}, \mathcal{T}) + Const.
\end{split}
\end{equation}
We can note from the Eq.~\ref{eq:upper_bound}, the target error is upper bounded by three terms, namely expected prediction error on the source domain, domain divergence denoted in Eq.\ref{eq:h_div}, and few constant terms. More details regarding both Eq.~\ref{eq:h_div} and Eq.~\ref{eq:upper_bound} are provided in \cite{ben2010theory}. A majority of the domain adaptation works in the literature rely on this formulation and focus on minimizing the upper bound on the target error by reducing the domain divergence between the source and target domain. In what follows, we discuss the different strategies used in the literature to address this specifically for the task of object detection.
% \left(1-\min_{D \in \mathcal{H}}\left(\operatorname{Err}_{\mathcal{S}}(D({X_s}))+\operatorname{Err}_{\mathcal{T}}(D({X_t}))\right)\right),

\begin{figure}[ht!]
	\begin{center}
		\includegraphics[width=0.95\linewidth]{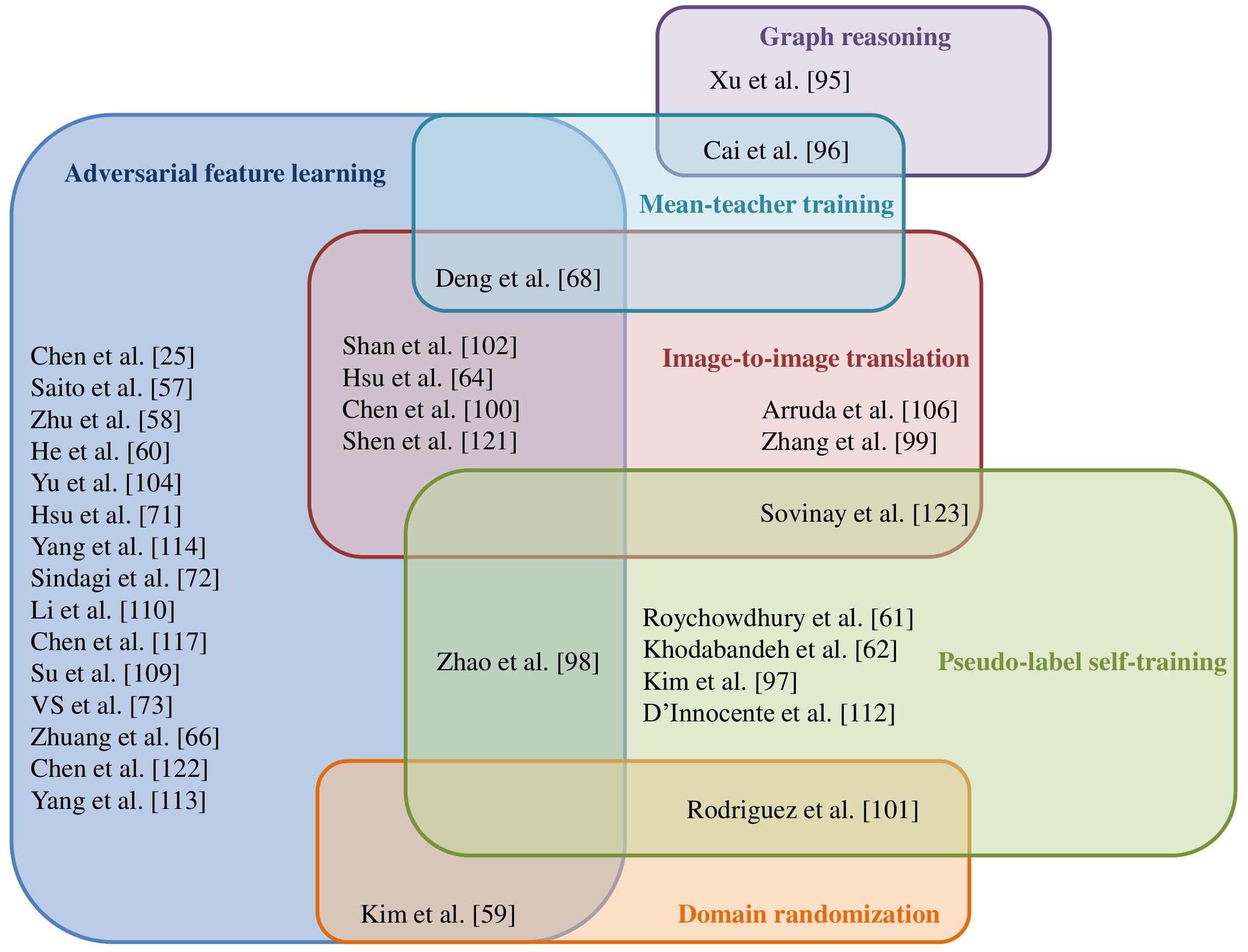}
	\end{center}
	\vskip -15.0pt
	\caption{The figure provides a Venn diagram of domain adaptive object detection methods. Each method falls into a class of adaptation approach (discussed in Sec.~\ref{sec:methods}), listed in the set representing respective techniques. Some of the work fall into more than one class of approaches, listed in the overlap region of set of respective class of adaptation techniques.}
	\label{fig:venndia} 
\end{figure}

\section{Methods}\label{sec:methods}

As discussed earlier, a variety of methods \cite{chen2018domain}, \cite{saito2019strong}, \cite{zhu2019adapting}, \cite{kim2019diversify}, \cite{he2019multi}, \cite{roychowdhury2019automatic}, \cite{khodabandeh2019robust}, \cite{xie2019multi}, \cite{hsu2020progressive}, \cite{tzeng2018splat}, \cite{zhuang2020ifan}, \cite{he2020domain}, \cite{deng2021unbiased}, \cite{xu2020exploring}, \cite{pan2020multi}, \cite{hsu2020every}, \cite{sindagi2020prior}, \cite{vs2021mega}, \cite{li2020free}, \cite{liang2020domain}, \cite{guan2021uncertainty}, \cite{sun2021multi}, \cite{zhang2021detr}, \cite{scheck2020unsupervised}, \cite{yang2020channel} have been proposed for the task of domain adaptive object detection. Based on a meticulous review of these approaches, we categorize them into the following classes.\\

\noindent\textit{1)} \textbf{Adversarial feature learning}: This class of adaptation approach performs an adversarial training of object detector model with the help of a domain discriminator. The training follows a gradient reversal layer based feature learning proposed by Ganin \etal \cite{ganin2016domain}. Specifically, the detector model is trained to produce features that fool the domain discriminator, while domain discriminator is tasked to correctly classify the features as source/target domain. This results in detector model producing domain invariant features which are useful to perform detection in target domain. Many methods in the literature utilize this strategy to adapt detectors to target domain \cite{chen2018domain}, \cite{saito2019strong}, \cite{sindagi2020prior}, \cite{vs2021mega}, \cite{hsu2020every}.\\

\noindent\textit{2)} \textbf{Pseudo-label based self-training}: Many works in the literature \cite{inoue2018cross}, \cite{roychowdhury2019automatic}, \cite{khodabandeh2019robust}, \cite{kim2019self}, \cite{zhao2020collaborative}  utilize highly confident predictions by source-trained detector model to train the it on the target. Since confident predictions on target domain have higher chance of being correct, such training strategy progressively makes detector model better on the target.\\

\noindent\textit{3)} \textbf{Image-to-image translation}: The image-to-image translation based strategy utilize an unpaired image-translation model to translate target image to a source-like image or vice versa. This reduces distribution shift in the visual domain and makes it easier for detector to perform well on the source-like target images. Many work in the literature utilize such approach for improving detector performance on the target domain \cite{zhang2019cycle}, \cite{hsu2020progressive}, \cite{chen2020harmonizing}, \cite{rodriguez2019domain}.\\

\noindent\textit{4)} \textbf{Domain randomization}: Another interesting way to improve the detector performance on target domain is to devoid all source-style bias from the model. Domain randomization strategy creates multiple stylized version of source domain data to train the detector model such that the model is not biased towards any one style and generalizes better on the target domain. Some works in the literature such as  \cite{kim2019diversify}, \cite{rodriguez2019domain} follow this strategy.\\

\renewcommand{\arraystretch}{0.8}
\setlength{\tabcolsep}{4.5pt}
\begin{table*}[ht!]
	\centering
	\caption{List of different domain adaptive object detection approaches. }
	\resizebox{1\linewidth}{!}{
		\label{tab:method_list}
		\begin{tabular}{|l|c|c|c|c|}
			\hline
			\textbf{Method}                                                                                           & \textbf{Detection framework}               & \textbf{Type}                                                                                             & \textbf{Publication}                 & \textbf{Year} \\ \hline
			Faster-RCNN in the wild \cite{chen2018domain}                               & Faster-RCNN                 & Adversarial feature learning                                                                    & Chen \etal, CVPR            & 2018          \\ \hline
			Cross-domain weakly supervised adaptation \cite{inoue2018cross}             & SSD                         & Pseudo-label based self-training                                                               & Inoue \etal, CVPR           & 2018          \\ \hline
			Strong weak distribution alignment \cite{saito2019strong}                   & Faster-RCNN                 & Adversarial feature learning                                                                    & Saito \etal, CVPR           & 2019          \\ \hline
			Selective cross-domain alignment \cite{zhu2019adapting}                     & Faster-RCNN                 & Adversarial feature learning                                                                    & Zhu \etal, CVPR             & 2019          \\ \hline
			Diversify and match \cite{kim2019diversify}                                 & Faster-RCNN                 & \begin{tabular}[c]{@{}c@{}}Domain randomization,               \\ Adversarial feature learning                            \end{tabular}       & Kim \etal, CVPR             & 2019          \\ \hline
			Automatic adaptation from unlabeled videos \cite{roychowdhury2019automatic} & Faster-RCNN                 & Pseudo-label based self-training                                                               & Roychowdhury \etal, CVPR    & 2019          \\ \hline
			Mean teacher with object relations \cite{cai2019exploring}                  & Faster-RCNN                 & \begin{tabular}[c]{@{}c@{}}Graph-reasoning,\\ Mean-teacher training\end{tabular}                                                       & Cai \etal, CVPR             & 2019          \\ \hline
			Multi-adversarial adaptation \cite{he2019multi}                             & Faster-RCNN                 & Adversarial feature learning                                                                    & He \etal, ICCV              & 2019          \\ \hline
			Robust learning from noisy labels \cite{khodabandeh2019robust}              & Faster-RCNN                 & Pseudo-label based self-trainig                                                                & Khodabandeh \etal, ICCV     & 2019          \\ \hline
			Self-training for one-stage detector \cite{kim2019self}                     & SSD                         & Pseudo-label based self-training                                                               & Kim \etal, ICCV             & 2019          \\ \hline
			Multi-level adaptation \cite{xie2019multi}                                  & Faster-RCNN                 & Adversarial feature learning                                                                    & Xie \etal, ICCV Workshop    & 2019          \\ \hline
			Pixel and feature adaptation \cite{shan2019pixel}                           & Faster-RCNN                 & \begin{tabular}[c]{@{}c@{}}Image-to-image translation,               \\ Adversarial feature learning\end{tabular}                                   & Shan \etal, Neurocomputing  & 2019          \\ \hline
			Adapting from synthesis to reality \cite{xu2019adversarial}                 & SSD                         & Adversarial feature learning                                                                    & Xu \etal, IEEE Access       & 2019          \\ \hline
			Improving localization \cite{yu2019dalocnet}                                & Faster-RCNN                 & Adversarial feature learning                                                                    & Yu \etal, IEEE Access       & 2019          \\ \hline
			Cycle-consistent adaptation \cite{zhang2019cycle}                           & Faster-RCNN                 & Image-to-image translation                                                                 & Zhang \etal, IEEE Access    & 2019          \\ \hline
			Cross-domain scene text \cite{chen2019cross}                                & Faster-RCNN                 & Adversarial feature learning                  & Chen \etal, ICNIP           & 2019          \\ \hline
			Cross domain detection image translation \cite{arruda2019cross}             & Faster-RCNN                 & Image-to-image translation                                                                 & Arruda \etal, IJCNN         & 2019          \\ \hline
			Graph-induced prototype alignment \cite{xu2020cross}                        & Faster-RCNN                 & Graph-reasoning                                                              & Xu \etal, CVPR              & 2020          \\ \hline
			Coarse-to-fine adaptation \cite{zheng2020cross}                             & Faster-RCNN                 & Adversarial feature learning                                                                    & Zheng \etal, CVPR           & 2020          \\ \hline
			Harmonizing transferability and discriminability \cite{chen2020harmonizing} & Faster-RCNN                 & \begin{tabular}[c]{@{}c@{}} Image-to-image translation,               \\ Adversarial feature learning\end{tabular}   & Chen \etal, CVPR            & 2020          \\ \hline
			Cross-domain document object detection \cite{li2020cross}                   & Faster-RCNN                 & Adversarial feature learning                                                                    & Li \etal, CVPR              & 2020          \\ \hline
			Categorical regularization \cite{xu2020exploring}                           & Faster-RCNN                 & Adversarial feature learning                                                                    & Xu \etal, CVPR              & 2020          \\ \hline
			Prior-based detector adaptation \cite{sindagi2020prior}                     & Faster-RCNN                 & Adversarial feature learning                                                                    & Sindagi \etal, ECCV         & 2020          \\ \hline
			Every pixel matters \cite{hsu2020every}                                     & FCOS \cite{tian2019fcos}    & Adversarial feature learning                                                                    & Hsu \etal, ECCV             & 2020          \\ \hline
			Collaborative training \cite{zhao2020collaborative}                         & Faster-RCNN                 & \begin{tabular}[c]{@{}c@{}}Pseudo-label based self-training,               \\ Adversarial feature learning\end{tabular}                                   & Zhao \etal, ECCV            & 2020          \\ \hline
			Conditional normalization network \cite{su2020adapting}                     & Faster-RCNN                 & Adversarial feature learning                                                                    & Su \etal, ECCV              & 2020          \\ \hline
			Spatial attention pyramid network \cite{li2020spatial}                      & Faster-RCNN                 & Adversarial feature learning                                                                    & Li \etal, ECCV              & 2020          \\ \hline
			Asymmetric tri-way training \cite{he2020domain}                             & Faster-RCNN                 & Adversarial feature learning                                                                    & He \etal, ECCV              & 2020          \\ \hline
			Dual multi-label prediction \cite{zhao2020adaptive}                         & Faster-RCNN                 & Adversarial feature learning                                                                    & Zhao \etal, ECCV            & 2020          \\ \hline
			One-shot cross-domain adaptation \cite{d2020one}                            & Faster-RCNN                 & Pseudo-label based self-training                                                               & D'Innocente \etal, ECCV     & 2020          \\ \hline
% 			Style-mining for one-shot adaptation \cite{luo2020adversarial}              & Faster-RCNN                 & \begin{tabular}[c]{@{}c@{}}Pseudo-label based self-training,               \\ Domain randomization\end{tabular}                                           & Luo \etal, NeurIPS          & 2020          \\ \hline
			Progressive adaptation \cite{hsu2020progressive}                            & Faster-RCNN                 & \begin{tabular}[c]{@{}c@{}}Image-to-image translation,               \\ Adversarial feature learning\end{tabular}                                   & Hsu \etal, WACV             & 2020          \\ \hline
			Multi-scale robust discrimination \cite{pan2020multi}                       & Faster-RCNN                 & Adversarial feature learning                                                                    & Pan \etal, WACV             & 2020          \\ \hline
			Object detection via style consistency \cite{rodriguez2019domain}           & SSD                         & \begin{tabular}[c]{@{}c@{}}Domain randomization,               \\ Pseudo-label based self-training\end{tabular}                                     & Rodriguez \etal, BMVC       & 2020          \\ \hline
			Image-instance full alignment network \cite{zhuang2020ifan}                 & Faster-RCNN                 & Adversarial feature learning                                                                    & Zhuang \etal, AAAI          & 2020          \\ \hline
			Free lunch for source-free adaptation \cite{li2020free}                     & Faster-RCNN                 & Pseudo-label based self-training                                                               & Li \etal, AAAI              & 2020          \\ \hline
			Forward-backward cyclic adaptation \cite{yang2020unsupervised}              & Faster-RCNN                 & Adversarial feature learning                                                                    & Yang \etal, ACCV            & 2020          \\ \hline
			Domain invariant region proposal \cite{yang2020domain}                      & Faster-RCNN                 & Adversarial feature learning                                                                    & Yang \etal, ICME            & 2020          \\ \hline
			Uncertainty-aware distributional alignment \cite{nguyen2020domain}          & Faster-RCNN                 & Adversarial feature learning                                                                    & Nguyen \etal, ICM            & 2020          \\ \hline
			Region proposal oriented adaptation \cite{alqasir2020region}                & Faster-RCNN                 & Adversarial feature learning                                                                    & Alqasir \etal, ACIVS        & 2020          \\ \hline
			Cross-device OCT lesion detection \cite{yang2020unsupervised}               & Faster-RCNN                 & Adversarial feature learning                                                                    & Yang \etal, ISBI            & 2020          \\ \hline
			Memory-guided category-wise adaptation \cite{vs2021mega}                    & Faster-RCNN                 & Adversarial feature learning                                                                    & VS \etal, CVPR              & 2021          \\ \hline
			Implicit invariant one-stage network \cite{chen20213net}                    & SSD                         & Adversarial feature learning                                                                                                                                                 & Chen \etal, CVPR            & 2021          \\ \hline
 			Unbiased mean-teacher \cite{deng2021unbiased}                               & Faster-RCNN                 & \begin{tabular}[c]{@{}c@{}}Mean-teacher training, \\ Image-to-image translation, \\ Adversarial feature learning\end{tabular}                                                                    & Deng \etal, CVPR            & 2021          \\ \hline
 			Domain-specific suppression  \cite{wang2021domain}             & Faster-RCNN                          & Adversarial feature learning      & Wang \etal, CVPR            & 2021          \\ \hline
 			Augmented feature alignment  \cite{wang2021afan}             & Faster-RCNN                          & \begin{tabular}[c]{@{}c@{}}Image-to-image translation, \\ Adversarial feature learning\end{tabular}      & Wang \etal, TIP            & 2021          \\ \hline
			Instance-invariant progressive disentanglement \cite{wu2019instance}        & Faster-RCNN                 & \begin{tabular}[c]{@{}c@{}}Adversarial feature learning,               \\ Graph-reasoning \end{tabular}                                                               & Wu \etal, TPAMI             & 2021          \\ \hline
			Large-scale instance-level image-to-image translation \cite{shen2021cdtd}   & Faster-RCNN                 & \begin{tabular}[c]{@{}c@{}}Image-to-image translation,\\ Adversarial feature learning\end{tabular}                                               & Shen \etal, IJCV            & 2021          \\ \hline
			Scale-Aware Domain Adaptive Faster RCNN \cite{chen2021scale}   & Faster-RCNN                 & Adversarial feature learning                                                   & Chen \etal, IJCV            & 2021          \\ \hline
			Curriculum self-paced learning \cite{soviany2021curriculum}                 & Faster-RCNN                 & \begin{tabular}[c]{@{}c@{}}Pseudo-label based self-training,\\ Image-to-image translation\end{tabular}                                                    & Sovinay \etal, CVIU         & 2021         \\ \hline
			Adaptive transformer-based detector \cite{zhang2021detr}                    & DETR \cite{zhang2021detr}   & Adversarial feature learning                                                                    & Zhang \etal, archived       & ---          \\ \hline
		\end{tabular}
		%\abovedisplayskip
	}
\end{table*}

\noindent\textit{5)} \textbf{Mean-teacher training}: Mean-teacher is an effective way to utilize unlabeled data to improve model generalization by progressively training a detector model in a student-teacher framework. This motivated a few works in the literature \cite{deng2021unbiased}, \cite{cai2019exploring} to explore mean-teacher training to adapt detector model by utilizing unlabeled target domain.\\

\noindent\textit{6)} \textbf{Graph reasoning}: Some works in the literature \cite{cai2019exploring}, \cite{xu2020cross}, \cite{wu2019instance} exploit the inter-object and intra-object relationships that exist in detection dataset. These object relations are modeled through graphs which help the detector on target domain by training to enforce the same object relations.\\

Fig.~\ref{fig:venndia} illustrates the various categories of approaches. A comprehensive list of the existing approaches is presented in Table~\ref{tab:method_list}.  In what follows, we discuss the key papers related to the respective categories in the aforementioned list.

%  \textcolor{blue}{Some of these strategies are inspired by the classification/segmentation domain adaptation methods and include object detection-specific information.} 

%%%%%%%%%%%%%%%%%%%%%%%%%%%%%%%%%%%%%%%%%%%%%%%%%%%%%%%%%%%%%%%%%%%%% ADVERSARIAL FEATURE LEARNING %%%%%%%%%%%%%%%%%%%%%%%%%%%%%%%%%%%%%%%%%%%%%%%%%%%%%%%%%%%%%%%%%%%%%

\subsection{Adversarial feature learning}\label{subsec:adv_feat_learning}

\subsubsection{Adversarial training through gradient reversal}\label{subsubsec:ganin}
\begin{figure}[h!]
	\begin{center}
		\includegraphics[width=0.99\linewidth]{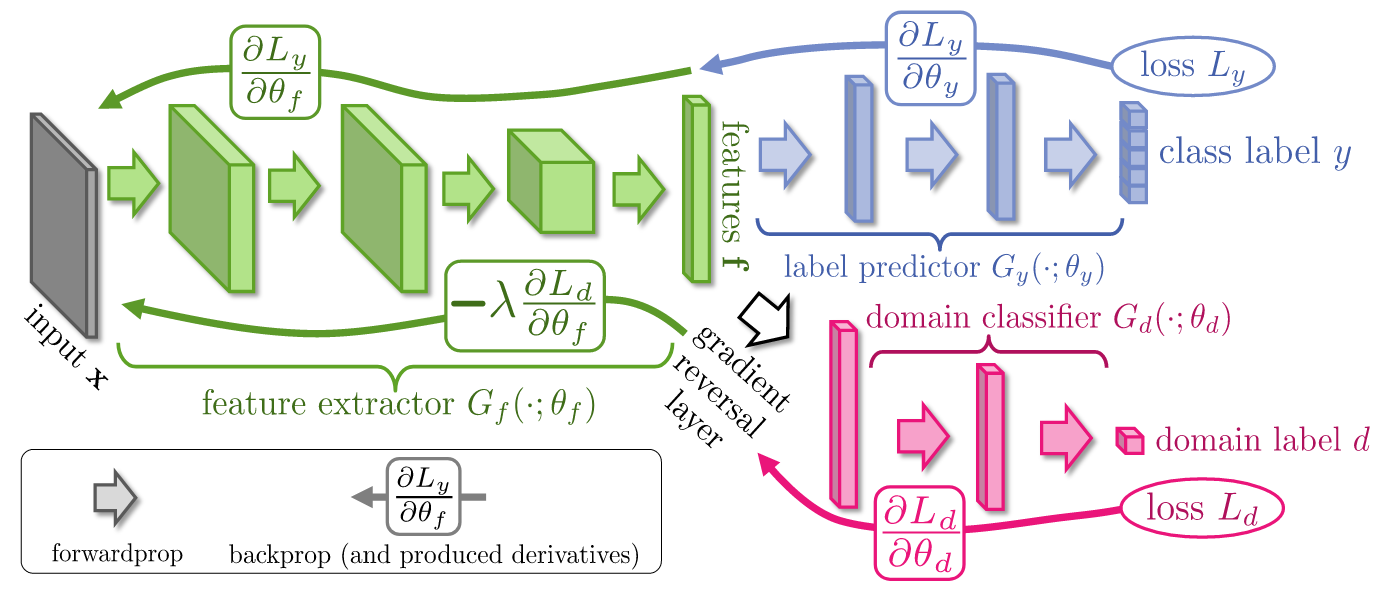}
	\end{center}
	\vskip -15.0pt
	\caption{Domain adaptation by backpropagation as proposed in \cite{ganin2016domain} with the example of classification task.}
	\label{fig:ganin}  
\end{figure}

The adversarial feature learning is built on the theory proposed by Ben \etal \cite{ben2010theory} (see Sec.~\ref{subsubsec:uda} for details). Specifically, the overall strategy involves minimizing the upper bound given in  Eq.~\ref{eq:upper_bound} by directly minimizing the $\mathcal{H}\Delta\mathcal{H}$-distance. As we can notice from $\mathcal{H}\Delta\mathcal{H}$-distance given in Eq.~\ref{eq:h_div}, this distance is inversely proportional to the error rate of the domain classifier $D$. The goal in a domain adaptation scenario is to reduce this distance, i.e., increase the domain classifier error. Ganin \etal \cite{ganin2016domain} exploited this and proposed  a novel gradient reversal approach to train any neural network model for domain adaptation. The overall goal is to achieve a domain invariant feature space of a backbone neural network with the help of a neural network-based domain classifier. Suppose we denote a domain classifier network as $D$ and the backbone feature extractor network as $F$. In that case, the feature extractor network also tries to increase the domain classifier loss. The network $F$ tries to minimize the task-specific loss (classification/segmentation/detection loss) and maximize the domain classification loss in the overall training pipeline. The network $D$ is trained to minimize domain classification loss. In addition to the task-specific loss, an additional loss involving domain classification is added. This loss is termed as adversarial loss \cite{ganin2016domain} and it can be written as:
\begin{equation}\label{eq:ganin_adv}
\max _{F} \min _{D \in \mathcal{H}}\left\{\operatorname{\textbf{E}}_{\mathcal{S}}(D)+\operatorname{\textbf{E}}_{\mathcal{T}}(D)\right\},
\end{equation}
where $\mathcal{H}$ denotes the hypothesis space for the domain classifier and $F$ is the feature extractor network. $\operatorname{\textbf{E}}_{\mathcal{S}}(D)$ and $\operatorname{\textbf{E}}_{\mathcal{T}}(D)$ denote the expected domain classification error over source and target domain, respectively. Eq.~\ref{eq:ganin_adv} is implemented with the help of a gradient reversal layer which is applied before the input to the domain classifier as shown in Fig.~\ref{fig:ganin}. The gradient reversal layer during feed-forward acts as an identity function and the gradients are multiplied with $-1$ during backpropagation. In effect, this forces feature extractor $F$ to maximize the domain classification loss while minimizing the task-specific loss resulting in the domain invariant feature space as proven by Ben \etal \cite{ben2010theory}. All the methods are utilizing this strategy to adapt a detector model using labeled source and unlabeled target domain fall under the \textit{adversarial feature learning} category.

\subsubsection{Domain adaptive detection via adversarial training}\label{subsubsec:det_adv_train}

Chen \etal \cite{chen2018domain} were among the first to formulate and address the problem of  domain adaptive  object detection. Fig.~\ref{fig:dafaster} illustrates an overview of their approach. Given the problem setup discussed in Sec.~\ref{subsubsec:uda} with a source $\mathcal{S}$ and target $\mathcal{T}$ domain, the method utilizes Faster-RCNN detection framework and proposes to utilize gradient reversal training at multiple stages of the detection framework. Specifically, given a backbone network $F_b$, RCN network $F_{rcnn}$, and RPN network $F_{rpn}$ in the detection model, they apply adversarial training at both the \textit{image-level} features which are extracted from backbone $F_b$ and the \textit{instance-level} features  that are extracted from the RCN network, $F_{rcnn}$. Let the features extracted from the backbone be denoted as $F_b(X)\in\mathbb{R}^{C \times H \times W}$, where $C$ denotes the number of channels, $H$ and $W$ denote the height and width of the feature map, respectively. Furthermore, the discriminator networks for adversarial training at {image-level} and {instance-level} are denoted as $D_{img}$ and $D_{inst}$. The {image-level} domain classification loss used to perform adversarial training at the image-level is then defined as:
\begin{figure}[b]
	\begin{center}
		\includegraphics[width=1.00\linewidth]{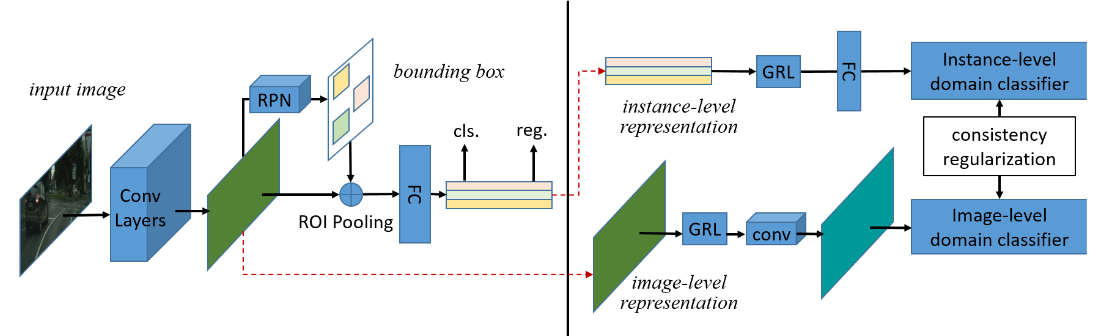}
	\end{center}
	\vskip -15.0pt
	\caption{Domain adaptive object detection in the wild \cite{chen2018domain} applies adversarial feature learning at image-level and instance-level with gradient reversal applied to detection backbone feature map and pooled features of the Faster-RCNN model. To regularize the adaptation, a consistency regularization is applied to make both image-level and instance-level domain classifier in sync with each other.}% Image taken from \cite{chen2018domain}.
	\label{fig:dafaster} 
\end{figure}
\begin{equation}\label{eq:dafaster_img}
\begin{split}
\mathcal{L}_{img}&=-\sum\limits_{i,h,w}[y^i_d \log D_{img}(F_b(X^i)^{(h,w)}) \ + \ \\
&(1-y^{i}_d) \log (1-D_{img}(F_b(X^i)^{(h,w)}))],
\end{split}
\end{equation}
where $i$ indicates $i^{th}$ image in the batch, $h\in\{1,\dots,H\}$, $w\in\{1,\dots,W\}$, and $F_b(X^i)^{(h,w)}$ denotes the feature of size $1\times C$ at location $(h,w)$ in the feature map. The output of the discriminator network $D_{img}$ is a  probability score indicating whether the given image is from source domain or target domain. Here, $y^i_d\in\{0, 1\}$ denotes the domain label which is $0$, when $X^i \in \mathcal{S}$ and $1$, when $X^i \in \mathcal{T}$. Similarly, the instance-level domain classification loss used to perform adversarial training at the instance level is defined as:
\begin{equation}\label{eq:dafaster_inst}
\begin{split}
\mathcal{L}_{inst}&=-\sum\limits_{i,j}[y^i_d \log D_{inst}(\textbf{f}^{pooled}_{ij}) \ + \ \\
&(1-y^{i}_d) \log (1-D_{inst}(\textbf{f}^{pooled}_{ij}))],
\end{split}
\end{equation}
where $\textbf{f}^{pooled}_{ij}$ indicates the RoI-pooled feature of size $1 \times d$ from the $j^{th}$ proposal of the image $X^i$. Since, both image and instance domain discriminators are trained independently and for any image $X^i$ should result in same domain prediction, Chen \etal \cite{chen2018domain} introduce a regularization that enforces consistency across predictions of the domain discriminators. This consistency regularization is defined as:
\begin{equation}
\begin{split}
&\mathcal{L}_{consistency}=\\
&\sum\limits_{i,j}\|\frac{1}{N^i_{act}}\sum\limits_{h, w} D_{img}(F_b(X^i)^{(h,w)})-D_{inst}(\textbf{f}^{pooled}_{ij})\|_{2},
\end{split}
\end{equation}
where $N^i_{act}$ indicates number of activations in the feature map $F_b(X^i)$ of a given image $X^i$ and $\|\cdot\|_2$ indicates $L2$-norm. Let us denote the overall detection model as $F=\{F_b, F_{rpn}, F_{rcnn}\}$. Combining all these loss functions, the final training objective is given as:
\begin{equation}\label{eq:dafaster_final_1}
\max_{\ D_{inst}, \ D_{img}} \min_{F} \ \mathcal{L}^{frcnn}_{det} - \lambda \ (\mathcal{L}_{img} + \mathcal{L}_{inst}),
\end{equation}
\begin{equation}\label{eq:dafaster_final_2}
\min_{F, \ D_{inst}, \ D_{img}} \lambda \ \mathcal{L}_{consistency},
\end{equation}
where $\lambda$ is a trade-off parameter used to balance adversarial and consistency loss. The detection and the discriminator networks are trained with final objective, Eq.~\ref{eq:dafaster_final_1} and Eq.~\ref{eq:dafaster_final_2}. To summarize, the detection network $F$ aims to minimize the detection loss and maximize the image-level and instance-level domain classification loss. The discriminator networks aim to minimize the domain classification loss. Note that both the detector and the discriminators aim to minimize the consistency loss. The detection loss is applied only to the source data since target data does not have any label annotations. The domain classification loss and consistency regularization are applied on both labeled source and unlabeled target data.
\begin{figure}[b]
	\begin{center}
		\includegraphics[width=0.85\linewidth]{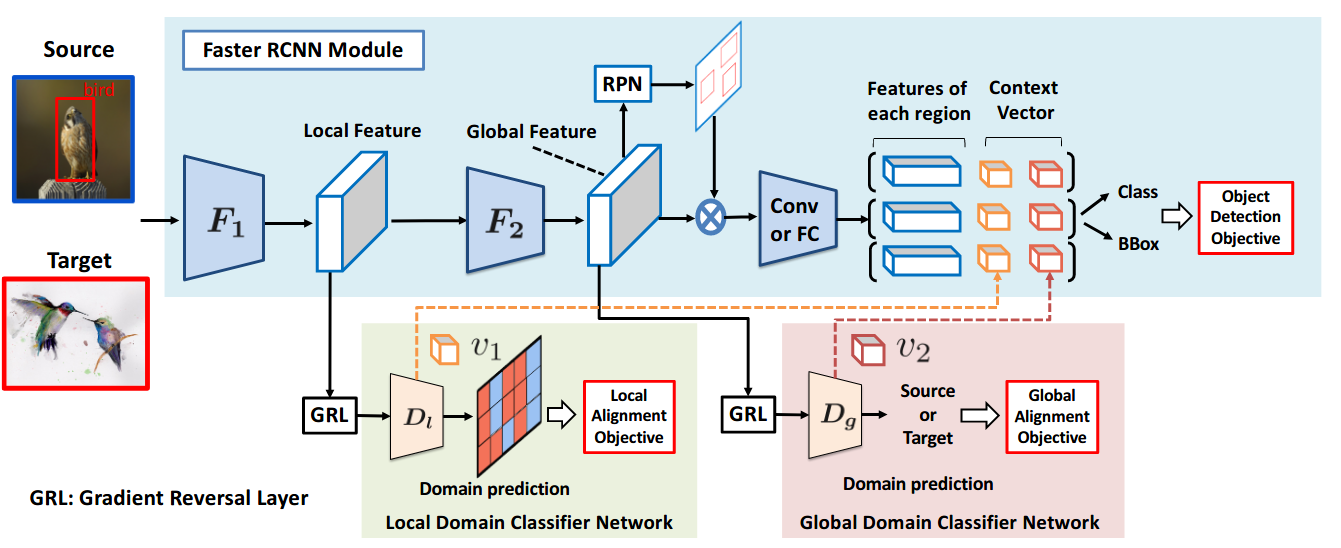}
	\end{center}
	\vskip -15.0pt
	\caption{Strong weak alignment for adaptive object detection \cite{saito2019strong}. One discriminator is applied at local-level (early layers of backbone) to perform a strong feature alignment via gradient reversal layer and another discriminator performs a weak alignment on the global-level (final layers of backbone). Furthermore, the features learned by the discriminator are fed to the classification network to provide domain context to improve the performance further.}
	\label{fig:swda} 
\end{figure}

Saito \etal \cite{saito2019strong} argue that directly applying the gradient reversal at multiple levels in the backbone network is not necessarily optimal. Since shallower convolutional layers of the backbone capture local information, directly applying the adversarial loss would be beneficial in learning domain invariant local features between source and target domain data. This type of alignment is to as \textit{strong alignment} in their work. Further, they reason that since the deeper layers in the backbone capture global information, performing a similar strong alignment would not be optimal. For example, when source and target domain data are sampled from different countries/cities, the number of objects in a scene, object co-occurrence, scene layout, etc can be very different. Consider the case when the source domain image contains only one object, whereas the target image contains multiple objects, performing strong alignment would likely increase the risk of misalignment. To tackle this issue, \cite{saito2019strong} modified the adversarial loss at the global-level (later convolutional layers of backbone network) by replacing the traditionally used binary cross-entropy loss with focal loss \cite{lin2017focal}. This strategy in \cite{saito2019strong} is referred to as \textit{weak-alignment}. The overall approach then utilizes both the local \textit{strong alignment} and global \textit{weak-aliment} to reduce the domain gap between source and target domain data, resulting in increased detection performance for target images. Let the detection backbone network $F_b$ be divided into two sub-networks that are cascaded together, namely global feature extractor $F_g$ and local feature extractor $F_l$ such that, $F_b(X) = F_g(F_l(X))$. The adversarial loss for strong alignment is defined as:
\begin{equation}
\begin{array}{l}
\mathcal{L}_{local_{s}}=\frac{1}{N_{s} H W} \sum\limits_{i=1}^{N_{s}} \sum\limits_{w=1}^{W} \sum\limits_{h=1}^{H} \|D_{l}\left(F_{l}\left(X^{i}_{s}\right)\right)^{(h,w)}\|^{2}, \\
\mathcal{L}_{local_{t}}=\frac{1}{N_{t} H W} \sum\limits_{i=1}^{N_{t}} \sum\limits_{w=1}^{W} \sum\limits_{h=1}^{H}\|1-D_{l}\left(F_{l}\left(X^{i}_{t}\right)\right)^{(h,w)}\|^{2}, \\
\mathcal{L}_{local}=\frac{1}{2}\left(\mathcal{L}_{local_{s}}+\mathcal{L}_{local_{t}}\right),\\
\end{array}
\end{equation}
where $D_{l}$ denotes the local domain discriminator, $N_s$ and $N_t$ are the number of source and target examples respectively, $\mathcal{L}_{local_{s}}$ and $\mathcal{L}_{local_{t}}$ denote the source and target domain classification loss respectively, and $F_{l}(X^{i}_{s}), F_{l}(X^{i}_{s}) \in \mathbb{R}^{C \times H \times W}$ are source and target local feature maps respectively. Denoting the global discriminator network as $D_g$, the weak alignment loss is defined as:
\begin{equation}
\begin{aligned}
&\mathcal{L}_{{global}_{s}}=-\frac{1}{N_{s}} \sum\limits_{i=1}^{N_{s}}\mathrm{FL}(D_{g}\left(F_b\left(X^{i}_{s}\right)\right),\\
&\mathcal{L}_{{global}_{t}}=-\frac{1}{N_{t}} \sum\limits_{i=1}^{N_{t}}\mathrm{FL}(1-D_{g}\left(F_b\left(X^{i}_{t}\right)\right),\\
&\mathcal{L}_{global}=\frac{1}{2}\left(\mathcal{L}_{global_{s}}+\mathcal{L}_{global_{t}}\right),
\end{aligned}
\end{equation}
where $\mathcal{L}_{{global}_{s}}$ and $\mathcal{L}_{{global}_{t}}$ are weak global alignment loss applied on source and target domain images, respectively. $F_b(\cdot)$ extracts the global feature, and $D_g(\cdot)$ denotes the probability of the global feature being from the source domain or target domain. $\mathrm{FL}$ denotes the focal loss defined as:
\begin{equation}
\mathrm{FL}(p)=-(1-p)^{\gamma} \log(p),
\end{equation}
where $\gamma$ is a focal loss parameter that controls the weight on hard-to-classify examples \cite{lin2017focal}. More specifically, the value of $\gamma>1$ will assign low loss values for the easy samples and high loss for hard samples, thereby focusing on the hard samples while training. As a result, while performing adversarial training with gradient reversal layer at the global level, hard target samples will be given more focus and easy to classify target examples will not be forced to align with the source domain. The paper demonstrated adaptation in the Faster-RCNN framework with this strong and weak alignment strategy and specifically showed the significance of weak alignment at the global level. Additionally, to further improve the performance of local and global domain discriminators, $D_l$ and $D_g$ are concatenated with RCN features to add domain-specific context information for object classification, as shown in Fig.~\ref{fig:swda}. The final objective of the network is defined as:
\begin{equation}
\max_{D_{g}, \ D_{l}} \min_{F} \ \mathcal{L}_{det} \ - \ \lambda \ ( \mathcal{L}_{global} + \mathcal{L}_{local}),
\end{equation}
where $\lambda$ is a trade-off hyper-parameter and $F$ denotes the entire detection network.

\begin{figure}[h!]
	\begin{center}
		\includegraphics[width=1.0\linewidth]{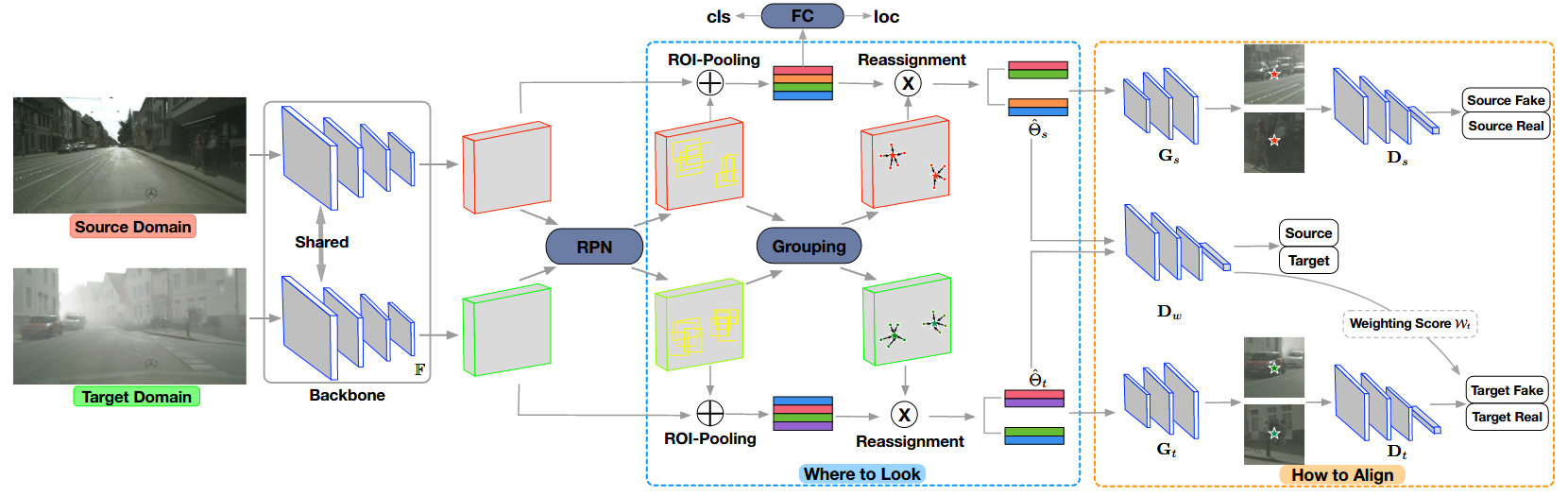}
	\end{center}
	\vskip -15.0pt
	\caption{Selective cross-domain alignment method \cite{zhu2019adapting}. First, the region of interest-based grouping strategy is used to identify the discriminative regions. Next, the discriminative regions are used to perform a weighted alignment using generative adversarial training.}
	\label{fig:selective_da} 
\end{figure}

A major drawback of the methods discussed so far is that they try to utilize the entire feature map to perform the alignment. However, a more optimal approach would be to perform the feature alignment on regions corresponding to objects in detection. Zhu \etal \cite{zhu2019adapting} base their method on this observation and selectively align the features of source and target domain data by mining the regions that are discriminative. For this, the authors exploited the region proposal network of the Faster-RCNN detection framework. Their method is divided into two parts, namely ``where to look'' and ``how to align", as illustrated in  Fig.~\ref{fig:selective_da}. In  the ``where to look'' stage, region proposals generated by the RPN network are mined to find groups within the feature maps. To overcome the noisy proposals of the RPN network, \textit{K-means} clustering is performed using the center coordinates of the proposals. The cluster centers obtained through the \textit{K-means} are used as grouped regions. Based on these mined groups, a fixed number of features are reassigned to each group (i.e. cluster). Instead of performing alignment with gradient reversal layer on these grouped features, a generative adversarial network \cite{goodfellow2014generative} based strategy is used to perform indirect feature alignment through generation. A similar strategy has been shown to work well in the case of classification \cite{hu2018duplex}, \cite{sankaranarayanan2018generate}. Specifically, they use the features in a group to reconstruct the corresponding patch of the original image by performing within-domain (i.e. $s\rightarrow s$, $t\rightarrow t$) and cross-domain (i.e. $t\rightarrow s$, $s\rightarrow t$) patch reconstructions. Denoting domain-specific generators as $G_s$ and $G_t$ and domains specific discriminators as $D_s$ and $D_t$ corresponding to the source and target domain, respectively, the patch reconstruction adversarial joint loss is defined as:
\begin{equation}
\mathcal{L}^{joint}_{adv}= \mathcal{L}_{D_s} \ + \ \mathcal{L}_{D_t} \ + \ \mathcal{L}_{G_{s}} \ + \ \mathcal{L}_{G_{t}} \ + \ \mathcal{L}_{F}.
\end{equation}

Here, the loss functions $\mathcal{L}_{D_s}$ and $\mathcal{L}_{D_t}$ are within domain losses that try to minimize the patch real/fake classification loss by predicting real input as real and fake input as fake. In contrast, for the generator network losses $\mathcal{L}_{G_{s}}$ and $\mathcal{L}_{G_{t}}$ then try to fool the discriminator networks by forcing them to identify fake as real and real as a fake within the same domain. The detection model $F$ aims to minimize the cross-domain patch reconstruction loss, i.e., classifying a fake source as a real target and a fake target as a real source. Similar strategies are used in \cite{hu2018duplex}, \cite{sankaranarayanan2018generate} for classification task. By minimizing the cross-domain patch reconstructions, the network $F$ will learn a domain invariant feature space resulting in a reduced gap between source and target domain. This strategy closely follows the gradient reversal training explained earlier. However, instead of directly applying adversarial loss on the feature space, the loss is applied to the reconstructed patch images. Additionally, a weight estimation network $D_w$ is also added to balance source and target domain adversarial losses. The weight estimation network is trained with a binary cross-entropy loss. Denoting the source and target domain RoI-pooled feature groups as $\tilde{\textbf{f}}_s$ and $\tilde{\textbf{f}}_t$ respectively, the weight estimation loss is defined as:
\begin{equation}
\mathcal{L}_w=\log(D_w(\tilde{\textbf{f}}_s)) + \log(1-D_w(\tilde{\textbf{f}}_t)).
\end{equation}
The primary benefit of using weight estimation network is that it weighs the alignment loss based on how similar the target patches look to the source domain patches. If we were to denote the output of the weight estimation network for the target domain pooled features as $\lambda_t$, the final objective for the method can be written as:
\begin{equation}
\begin{aligned}
\min_{F, G_s, G_t, D_{w}} \max_{D_t, \ D_s} \ & \mathcal{L}^{frcnn}_{det}  + \lambda_t  (\mathcal{L}_{D_t}  +  \mathcal{L}_{G_{t}}  + \mathcal{L}_{F}) \\
&+ (\mathcal{L}_{D_s} + \mathcal{L}_{G_{s}}) + \mathcal{L}_{w}.
\end{aligned}
\end{equation}

The network training is performed in a stage-wise manner by updating supervised detection loss, discriminator loss, weight loss, cross domain and within domain generation loss separately.

\subsubsection{Weighted adversarial feature alignment}\label{subsubsec:weighted_adv}

\begin{figure}[h!]
	\begin{center}
		\includegraphics[width=1.00\linewidth]{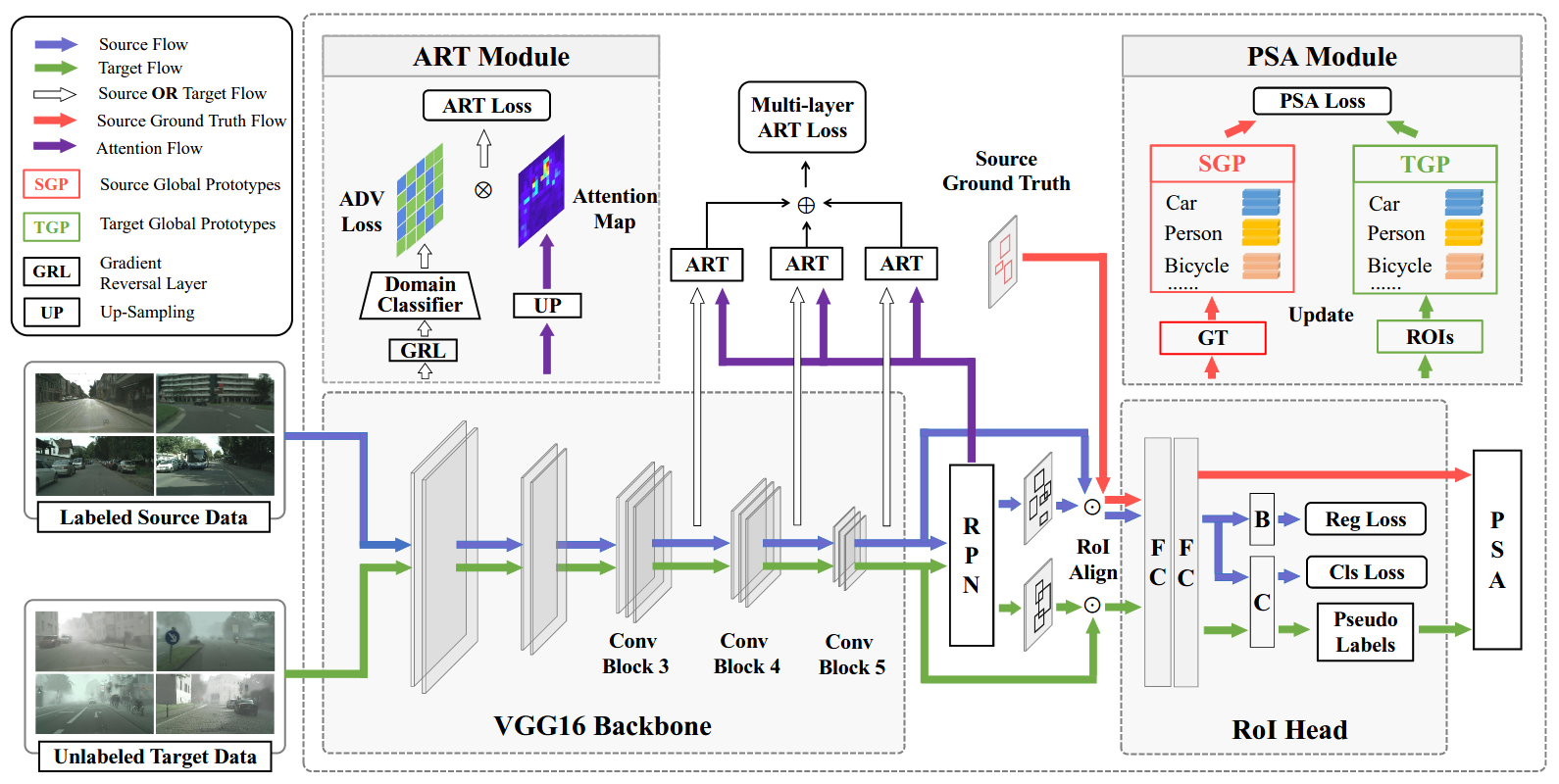}
	\end{center}
	\vskip -15.0pt
	\caption{The coarse-to-fine adaptation \cite{zheng2020cross} proposes a multi-level alignment by applying gradient reversal-based adversarial loss at multiple layers in the detection backbone network. An attention mechanism is introduced that weights the adversarial loss based on the activation strength at any particular location in the respective feature map. This loss weighting mechanism is called as Attention-based Regional Transfer (ART) in the figure. Furthermore, a Prototype Similarity Alignment (PSA) loss is introduced that aligns the object prototype features of both source and target domain to regularize the alignment.}% Image taken from \cite{zheng2020cross}.
	\label{fig:c2f_det} 
\end{figure}

Most of the work in the domain adaptive detection literature that are based on adversarial feature learning focus on improving the gradient reversal training in order to obtain better feature alignment. Various methods such as \cite{zheng2020cross}, \cite{xu2020exploring}, \cite{sindagi2020prior}, \cite{hsu2020every}, \cite{zhao2020adaptive}, \cite{vs2021mega} build on the approaches discussed earlier and broadly follow the strategy of introducing a module that can control the gradient reversal information flow through loss weighting in addition to using a regularization technique to complement the proposed weighting module. For example, Zheng \etal \cite{zheng2020cross} applies domain classifier at multiple levels of a detection backbone network to perform adversarial feature learning with gradient reversal layer, as shown in Fig.~\ref{fig:c2f_det}. These adversarial losses are then multiplied by weights extracted from the backbone. In order to obtain the weights, the final feature map of the backbone is averaged over the channel dimension to obtain an attention map highlighting regions that potentially have an object. These weights are then used to modulate the adversarial loss.  This strategy is referred to as Attention-based Region Transfer (ART). Furthermore, with the help of source ground-truth bounding boxes and predicted proposals for the target domain, class-specific prototypes are learned through feature averaging. At each step, both source and target  prototypes of each class are aligned through prototype similarity loss.  This strategy is referred to as Prototype Similarity Alignment (PSA). The ART loss helps gradient reversal training remove any noisy information coming from non-object regions resulting in better alignment, whereas prototype alignment maintains the semantic consistency while adapting to the target domain. 

\begin{figure}[h!]
	\begin{center}
		\includegraphics[width=1.00\linewidth]{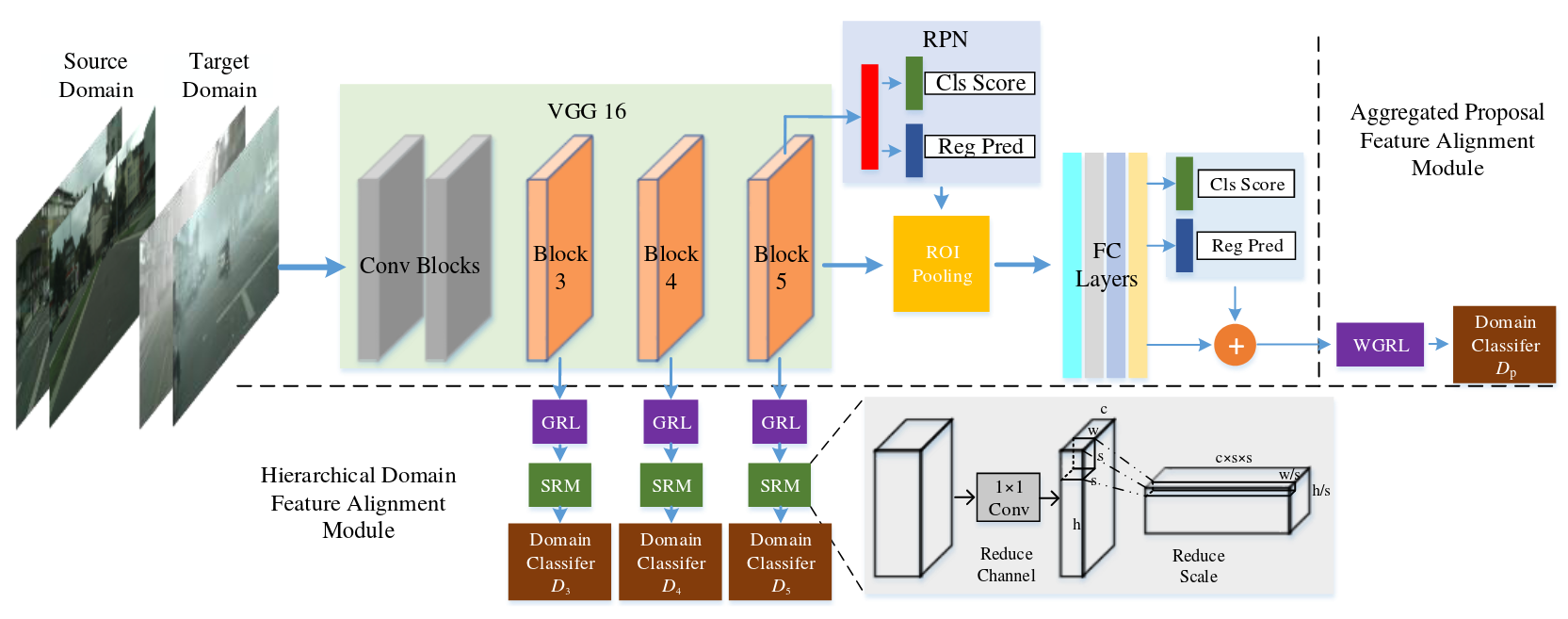}
	\end{center}
	\vskip -15.0pt
	\caption{Multi-adversarial Faster-RCNN \cite{he2019multi} extends the adversarial training strategy of \cite{chen2018domain} by performing image-level gradient reversal training at multiple layers of the detector backbone network. Additionally, it performs hierarchical alignment where the feature maps across the channel dimension are reduced before being forwarded through the domain classifier. Furthermore, it also performs instance-level gradient reversal training and the adversarial loss is weighted with the prediction probabilities obtained from the detection network for each respective RoI-pooled feature as illustrated in the image with Weighted Gradient Reversal Layer (WGRL) loss.}
	\label{fig:multi_adv} 
\end{figure}

Similarly, He \etal \cite{he2019multi} proposed a Multi-Adversarial Faster-RCNN (MAF) approach that extends the work by \cite{chen2018domain}. The first change in MAF is to apply image-level alignment at multiple layers of the backbone network. Additionally, the feature maps are reduced in the channel dimension to align the aggregated information rather than individual components. MAF also includes instance-level alignment but unlike \cite{chen2018domain}, it weights the instance adversarial loss with the prediction probabilities produced by the RCN network. Fig.~\ref{fig:multi_adv} illustrates the overall approach of MAF.

\begin{figure}[h!]
	\begin{center}
		\includegraphics[width=1.00\linewidth]{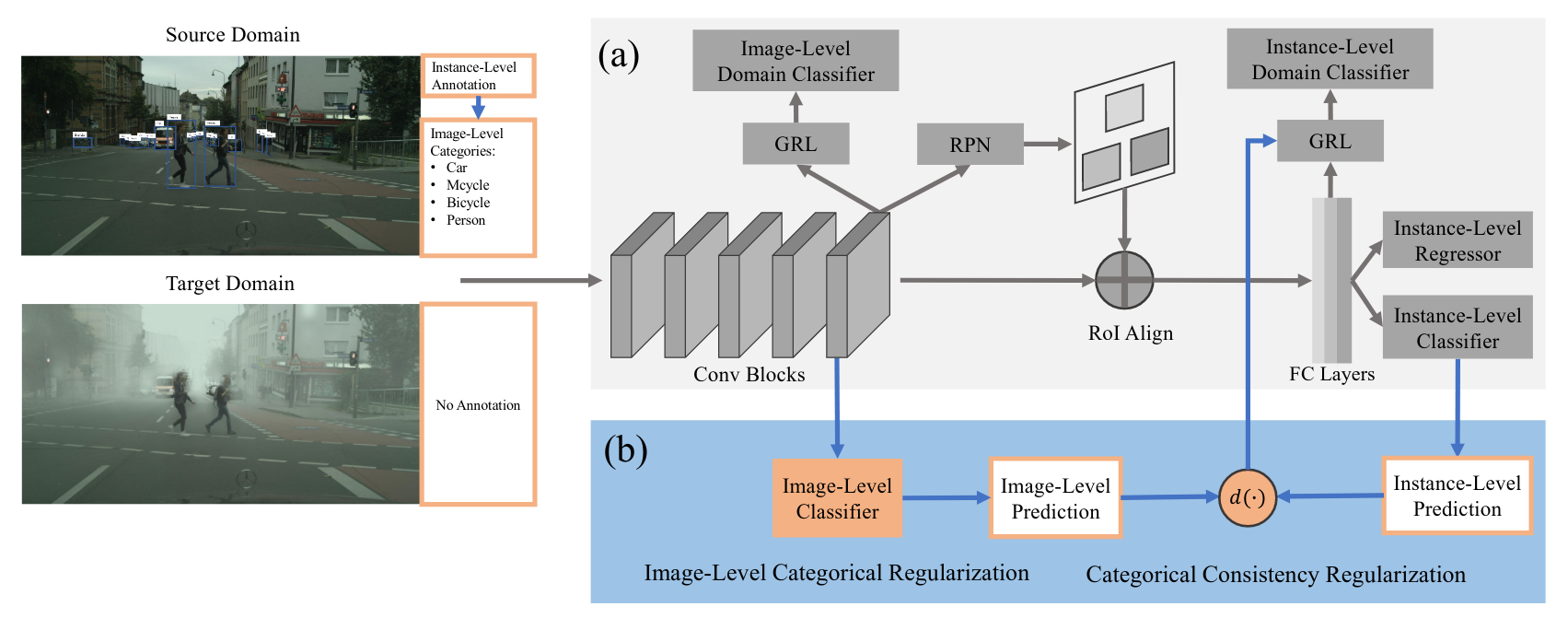}
	\end{center}
	\vskip -15.0pt
	\caption{The categorical regularization approach \cite{xu2020exploring} explores the use of multi-label classification on source domain supervised data along with the adversarial feature learning. The multi-label classification can extract weakly localized object maps that can help guide the adversarial training. The method proposes two additional losses, namely Image-level Categorical Regularization (ICR) and Consistency Categorical Regularization (CCR), that can be applied to any adversarial feature learning pipeline to enhance the performance.}
	\label{fig:cate_reg} 
\end{figure}

Xu \etal \cite{xu2020exploring} a proposed categorical regularization strategy that can be combined with adversarial feature learning based approaches to further enhance the feature alignment between source and target domain. The categorical regularization strategy utilizes instance-level annotations from the labeled source domain data and adds a multi-label classification loss in addition to the detection losses as illustrated in Fig.~\ref{fig:cate_reg}. Such a strategy helps extract weak localization of objects in the feature maps through a multi-label classifier. The weak localization map is then used to gate image-level domain classification loss to block unnecessary information while highlighting the object regions. This image-level weighting through weak localization is termed in the paper as Image-level Categorical Regularization (ICR). Subsequently, the instance-level regularization is applied to the RoI-pooled features. Specifically, the instance-level domain classification loss is weighted using the difference between the RCN prediction probability and the multi-label classification probability corresponding to the category of the respective pooled feature. This regularization is known as the Categorical Consistency Regularization (CCR) loss. Xu \etal showed the effectiveness of this categorical regularization by modifying both \cite{chen2018domain} and \cite{saito2019strong} alignment strategy with the addition of both ICR and CCR loss functions.

Zhao \etal \cite{zhao2020adaptive} combined the multi-label classification with the weak global alignment \cite{saito2019strong}. Specifically, a multi-label classifier is additionally trained along with the detection network with the help of source labeled data. The probability scores predicted by the multi-label classifier are used to condition the domain discriminator at the final layer of the backbone network. The conditioning mechanism takes in both the feature map extracted from the backbone and the multi-label probability vector indicating the probability of all $K$ objects being in the image, using a multi-linear mapping function. The weak global alignment is similar to the \cite{saito2019strong}, i.e., employing focal loss to perform global feature alignment. To regularize the feature alignment training further, the distance between renormalized prediction probability score vector from multi-label classifier and prediction probability score of RCN network is minimized via symmetric Kullback-Leibler divergence.

Sindagi \etal \cite{sindagi2020prior} considered adverse weather conditions as a special case of domain adaptation. They argue that in the case of adverse weather, well-defined models are available that mathematically formulate how the camera captures such conditions. Using these models, they extract weather-specific information termed as ``prior''. These \textit{priors} are then used to modify the conventional domain classification loss into a prior-prediction loss. The training follows a similar strategy of performing gradient reversal layer-based alignment using prior prediction loss. That is, the backbone network of the detection model tries to maximize the prior prediction loss, while the prior prediction network tries to minimize it. This strategy is termed as \textit{prior adversarial training} in their paper and is shown to be effective, especially in the case of hazy and rainy weather conditions.

In contrast to existing methods, Hsu \etal \cite{hsu2020every} do not utilize Faster-RCNN detection framework and instead they follow FCOS \cite{tian2019fcos} one-stage detection framework that has centerness loss in addition to object classification and bounding box regression losses for supervised detection. Hsu \etal \cite{hsu2020every} specifically exploit the FCOS framework to propose a center-aware feature alignment strategy. Since FCOS is trained with the centerness loss on labeled source data, it provides a coarse prediction of center-focused heatmap indicating the location of the objects. They utilize this center-information to create a class-agnostic map of objects and gate it with the final feature map of the backbone network. Both global and center-aware discriminators are applied on the respective feature maps to perform domain classification. The alignment is performed through gradient reversal layer-based adversarial training where the domain discriminators are responsible for center-aware feature alignment while rejecting any noisy information of the background with more focus on the prominent parts of the object instances. Similar to the other approaches, the feature alignment is performed at multiple levels of the backbone network.

\begin{figure}[h!]
	\begin{center}
		\includegraphics[width=1.00\linewidth]{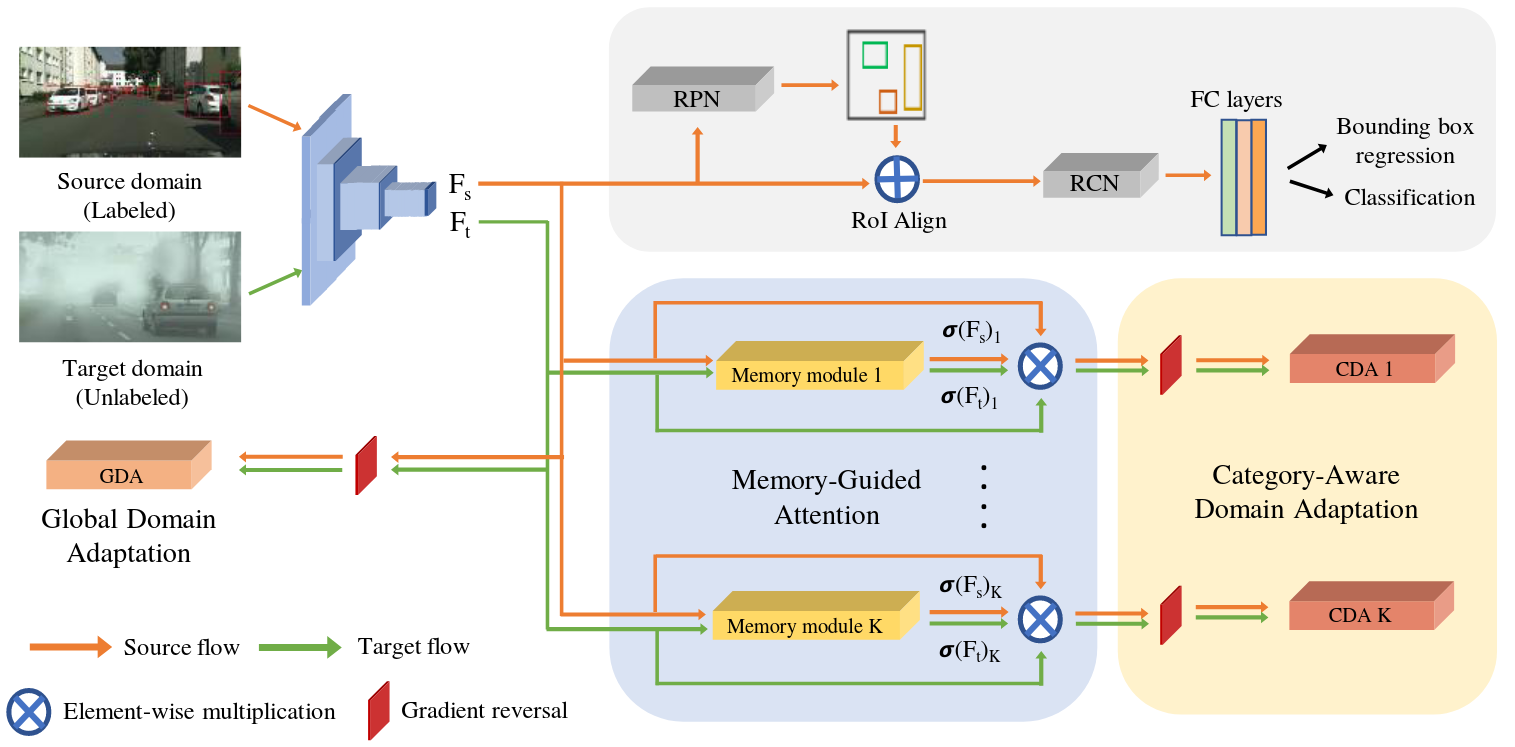}
	\end{center}
	\vskip -15.0pt
	\caption{Memory guided category wise adaptation \cite{vs2021mega} proposes a strategy to perform class-wise alignment between source and target domain. A memory module is utilized corresponding to each category that is iteratively updated to encode class-information. These memory modules are later used to create attention maps of respective categories. Through a global domain classifier and multiple category-specific domain classifier category-aware feature alignment is performed between source and target images.}
	\label{fig:mega_cda} 
\end{figure}

VS \etal \cite{vs2021mega} pointed out that existing adversarial learning-based approaches are prone to the category-wise \textit{negative transfer} problem. Training with gradient reversal and domain classifier only ensures that features become domain invariant. However, when both source and target domain contains multiple categories, which is often the case, there are chances of misalignment across different categories of target and source features. VS \etal \cite{vs2021mega} address this issue of negative transfer with by proposing memory-guided attention for category-aware domain adaptation (MeGA-CDA).  Their key idea is to utilize $K+1$ category-specific domain discriminators for aligning each category separately. However, the label information is unavailable for the target dataset, which prohibits the usage of category-specific domain discriminators. To overcome this, the authors introduce a memory module in the feature map extracted by the backbone network and produces attention maps that attend to the respective category features while blocking information from the rest of the categories. Fig.~\ref{fig:mega_cda} illustrates their overall approach of using memory modules for generating category-specific information and its subsequent usage in class-specific domain alignment. The memory module is trained jointly with the other components of the network, and this end-to-end training ensures that the category-specific attention map is progressively improved over the training process.  Note that they also use the category agnostic domain alignment that is similar to the strong alignment strategy of \cite{saito2019strong}. This ensures that the overall feature maps are aligned between source and target domain, while the additional category-specific discriminators prevent the negative transfer by performing the class-wise alignment. 

Let $F_b$ as the backbone network of the detection model, $X_s \sim \mathcal{S}$ and $X_t \sim \mathcal{T}$ be any arbitrary source and target domain image respectively, the typical weighted domain classification loss ($\mathcal{L}^{weighted}_{adv}$) used for most of the methods discussed in this section can be formulated as:
\begin{equation}
\begin{array}{l}
    \mathcal{L}^{weighted}_{adv}\left(X_{s}, X_{t}\right)= \\
\begin{aligned}
\sum_{h=1}^{H} & \sum_{w=1}^{W}  y_{d} \cdot \sigma_{img_s}^{(h, w)} \cdot \left(1-{D}\left(\sigma_{feat_s}^{(h, w)} \cdot F_b(X_s)^{(h, w)}\right)\right)^{2} \\
+&\left(1-y_{d}\right) \cdot \sigma_{img_t}^{(h, w)} \cdot \left({D}\left(\sigma_{feat_t}^{(h, w)} \cdot F_b(X_t)^{(h, w)}\right)\right)^{2},
\end{aligned}
\end{array}
\end{equation}
where $D$ is the image-level (in some cases instance-level) domain classifier, $y_{d} \in \{0, 1\}$ is domain label which is $1$ for source and $0$ for target domain images, $\sigma_{img_t}, \sigma_{img_s} \in \mathbb{R}^{H \times W}$ are the image-level weights applied on the domain classification loss, $\sigma_{feat_t}, \sigma_{feat_s} \in \mathbb{R}^{H \times W}$ are the feature-level weights applied to mask-out irrelevant information and highlight important features that aid source and target domain alignment. Depending on the method, $\sigma_{img}$ or $\sigma_{feat}$ are set to identity, and either image-level or feature level weights are applied for the loss computations. Methods such as\cite{zheng2020cross}, \cite{he2019multi}, \cite{xu2020exploring}, \cite{sindagi2020prior}, \cite{zhao2020adaptive} utilize only $\sigma_{img}$ weights to appropriately balance the adversarial loss. Whereas other methods such as \cite{hsu2020every}, \cite{vs2021mega} apply only feature-level weights $\sigma_{feat}$. The contribution of each method comes from the way either $\sigma_{img}$ and $\sigma_{feat}$ is modeled. For example, Xu \etal \cite{xu2020exploring} models $\sigma_{img}$ with the help of multi-label classification probability vector and VS \etal \cite{vs2021mega} models $\sigma_{feat}$ with category-specific memory modules.

\subsubsection{Additional methods}\label{subsubsec:additional_methods}

\begin{figure}[h!]
	\begin{center}
		\includegraphics[width=0.99\linewidth]{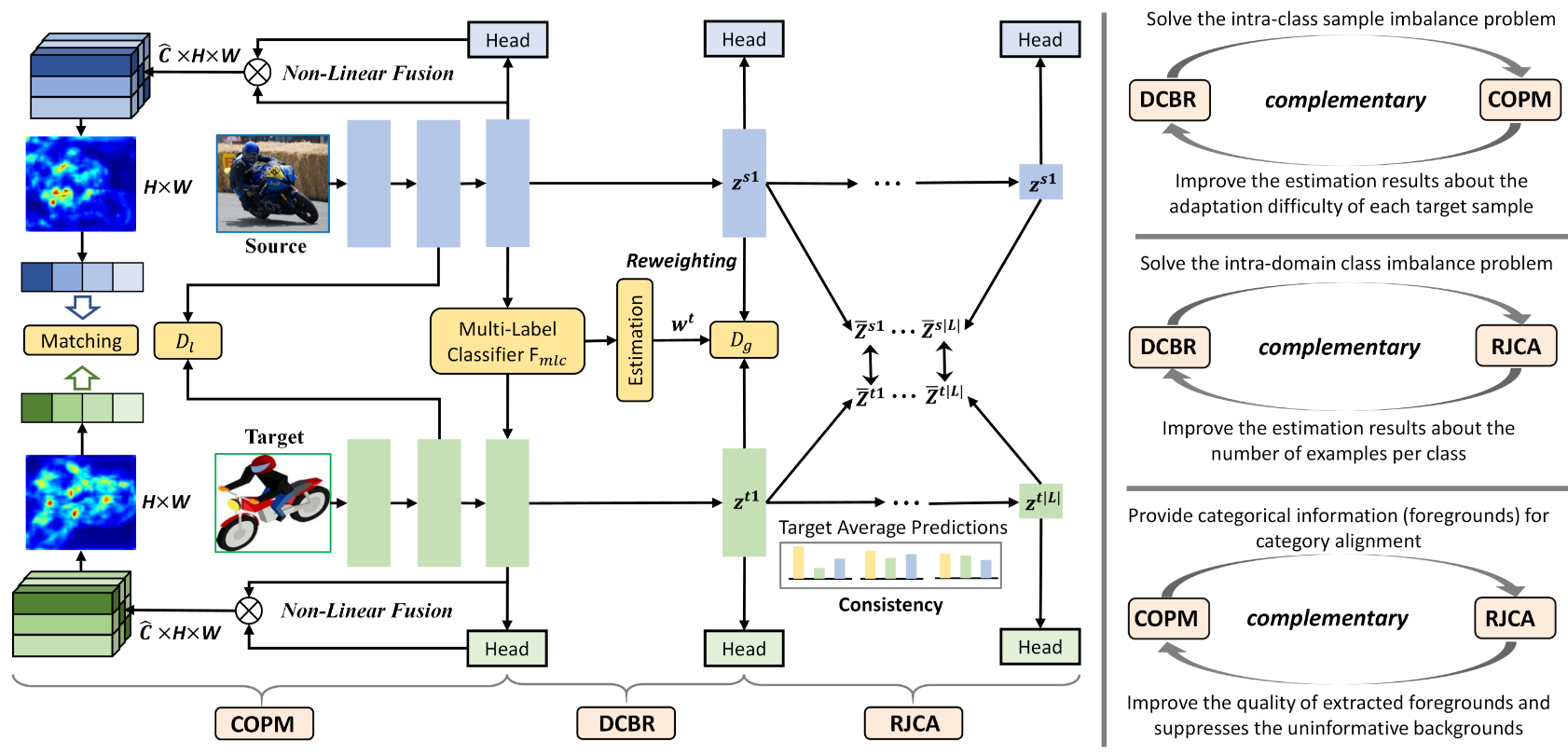}
	\end{center}
	\vskip -15.0pt
	\caption{Implicit instance variation network (I3Net) \cite{chen20213net} involves four major components, \textit{1)} pixel-level and image-level feature alignment through gradient reversal layer, \textit{2)} category-aware object-pattern matching (COPM) through non-linear fusion, and \textit{3)} dynamic class-balanced re-weighting (DCBR) through multi-label classifier, \textit{4)} regularization through joint category alignment (RJCA). All these modules are combined together to adapt an SSD framework based one-stage detection model for the target domain data.}
	\label{fig:i3net} 
\end{figure}

Chen \etal \cite{chen20213net} provided a strategy to align source and target domain features for SSD detection framework, termed as I3Net. Their main strategy is similar to that of multi-level discriminators  \cite{chen2018domain}, \cite{saito2019strong}, \cite{sindagi2020prior}, \cite{he2019multi}, \cite{zheng2020cross} that perform alignment at pixel-level and image-level as shown in Fig.~\ref{fig:i3net}. Furthermore, the domain loss used for training the image level discriminator, is weighted by the probability obtained from a multi-label classifier, similar to \cite{xu2020exploring}. Additionally, I3Net enforces object pattern matching by exploiting the SSD architecture, which predicts probability at each feature map location extracted from the backbone network. These category-specific probability maps are then matched between source and target images for respective categories to enforce consistency between source and target activations. The category features are further improved by minimizing intra-class distance and maximizing inter-class distance with a margin between category-specific prototypes. The prototypes are calculated using category-wise probability patterns obtained by SSD detection framework and are updated through the exponential moving average.

\begin{figure}[h!]
	\begin{center}
		\includegraphics[width=0.99\linewidth]{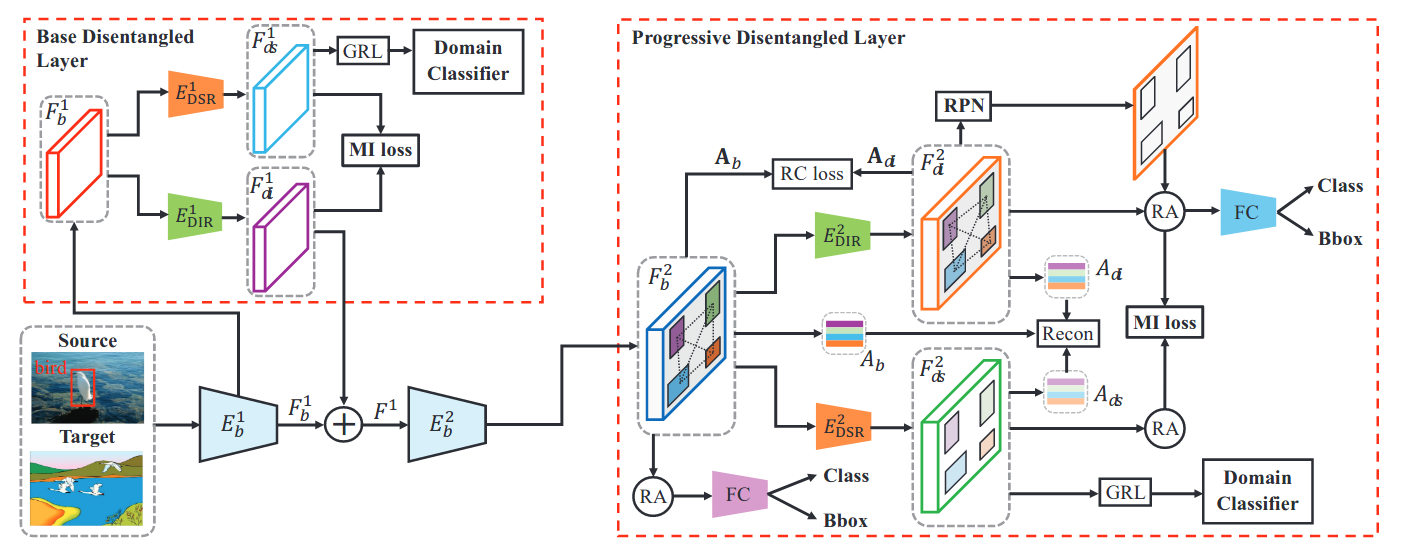}
	\end{center}
	\vskip -15.0pt
	\caption{Instance-invariant adaptation of object detector \cite{wu2019instance} introduces gradient reversal training at image-level and instance-level similar to \cite{chen2018domain}. Two disentanglement layers are proposed to achieve instance-invariance, which minimizes the Mutual Information (MI) loss between the domain-specific and domain-agnostic features. To further improve the disentanglement, a Relation Consistency (RC) loss is used by constructing an object relation graph. The overall model training is done in three stages resulting in an instance-invariant detection model that is better suited for the target domain data.}
	\label{fig:progressive_disentanglement} 
\end{figure}

Wu \etal \cite{wu2019instance} points out that feature-level or pixel-level alignment strategies suffer from the risk of neglecting the instance-level object characteristics. To achieve this, the features learned through the training are required to be disentangled into domain-invariant and domain-specific parts. Wu \etal \cite{wu2019instance} proposed a progressive disentanglement strategy that performs stage-wise training of Faster-RCNN. As shown in Fig.~\ref{fig:progressive_disentanglement}, both image-level and instance-level domain classifiers are employed with gradient reversal layer similar to \cite{chen2018domain}. Mutual Information (MI) loss is applied to the separated features to disentangle domain-invariant from domain-specific features. The MI loss formulation is borrowed from the Mutual Information Neural Estimation (MINE) approach \cite{belghazi2018mine}. Specifically, MINE-based MI loss transforms the intractable mutual information maximization into a tractable binary classification objective that can be trained in an end-to-end manner. To further regularize the training, object-relational graph is created with the help of domain-invariant features for each input image, by considering the original and domain-invariant feature maps. Since both features are extracted from the same image, they contain the same objects at the respective locations. Hence, a object Relational Consistency (RC) loss is enforced to maintain inter-class relationships across both feature maps.

There are several subsequent works such as \cite{zhuang2020ifan}, \cite{guan2021uncertainty}, \cite{xie2019multi}, \cite{hong2018conditional}, \cite{li2020spatial}, \cite{pan2020multi}, \cite{yang2020domain}, \cite{chen2021scale} that utilize the adversarial feature learning strategies discussed in this section. Most of these approaches address cross-domain detection for the application of autonomous driving/surveillance. Some of the notable works that address the issue of domain shift in other applications include Yang \etal \cite{yang2020unsupervised} which tackles the detector adaptation for medical tasks with the help of gradient reversal layer-based adversarial training for performing OCT lesion detection. In other tasks, Chen \etal \cite{chen2019cross} proposed adversarial feature training similar to \cite{chen2018domain} for adapting scene text detection models from synthetic data to outdoor settings, and Li \etal \cite{li2020cross} utilized a multi-level feature alignment applied at multiple convolutional layers of the backbone network to train a detector for cross-domain detection on documents. Additionally, Li \etal \cite{li2020cross} established a benchmark for the task of cross-domain detection in document space.
Furthermore, there are many other interesting works available as pre-print  \cite{liu2020domain}, \cite{fu2020deeply}, \cite{liu2020wqt}, \cite{liang2020domain}, \cite{yang2020channel}, \cite{salzmann2021attention}, \cite{nguyen2021incremental}. Amongst different types of domain adaptive detection techniques, adversarial feature learning is the most popular approach with different variations available in the literature.

% add later \cite{scheck2020unsupervised}
% i2i - 
% plst -  
% dr - 
% mt - 
%%%%%%%%%%%%%%%%%%%%%%%%%%%%%%%%%%%%%%%%%%%%%%%%%%%%%%%%%%%%%%%%%%%%% SELF-TRAINING PSEUDO_LABELS %%%%%%%%%%%%%%%%%%%%%%%%%%%%%%%%%%%%%%%%%%%%%%%%%%%%%%%%%%%%%%%%%%%%%

\subsection{Pseudo-label based self-training}\label{subsec:self_training_pseudo_labels}

Self-training of object detectors on the target domain with pseudo label-based supervision is one of the simplest approaches to adapt the source-trained model to the target domain. Given a source pre-trained detection model, $F^{src}$, which is trained on the annotated source domain dataset $\mathcal{S}=\{X_s^i, Y_s^i\}_{i=1}^{N_s}$, it is used to generate pseudo-labels on the target domain data $\mathcal{T} = \{X_t^i\}_{i=1}^{N_t}$. The pseudo-labels, $\tilde{Y}_t^i = F^{src}(X_t)$, along with the images from the  target dataset  forms a new dataset, $\tilde{\mathcal{T}} = \{X_t^i, \tilde{Y}_t^i\}_{i=1}^{N_t}$. Here, $\tilde{Y}_t^i = F^{src}(X_t)$ denotes pseudo-labels obtained from a source-trained detector model $F^{src}$ for $i^{th}$ target domain image $X_t^i$. Self-training typically involves using these pseudo-labels to re-train the network on the target data. In reality, the pseudo-labels are potentially noisy and often incorrect. Hence, directly training the model with these pseudo-annotations can potentially lead to a situation where the errors keep getting reinforced into the network. A filtering strategy is employed to overcome this issue, which involves removing annotations that are not confident. Most of the existing works revolve around developing complex and accurate filtering techniques to deal with the noise present in the pseudo-labels. Note that this training strategy with pseudo-labels is simple yet very effective as it attempts to directly minimize the target domain prediction error, unlike adversarial feature learning where the strategy is to minimize an upper bound over the target prediction error.  

\begin{figure}[h!]
	\begin{center}
		\includegraphics[width=0.99\linewidth]{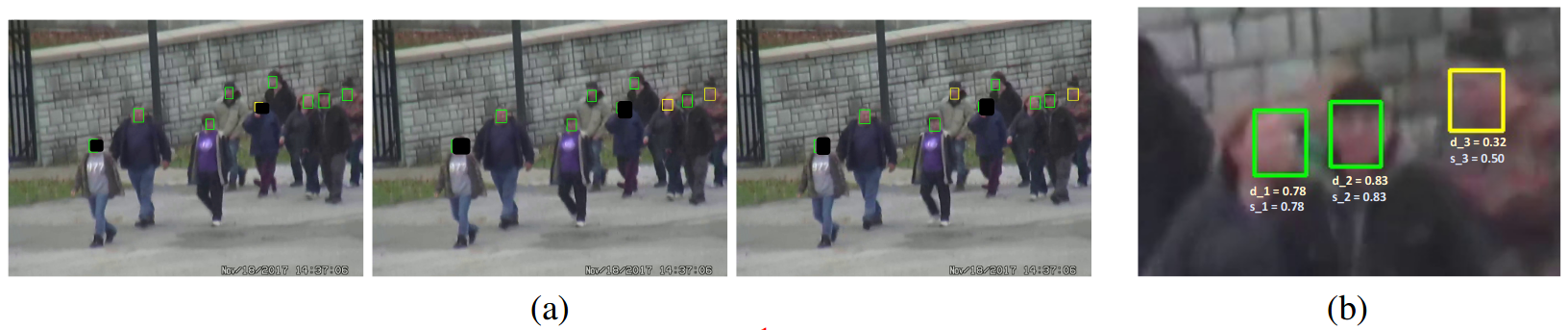}
	\end{center}
	\vskip -17.5pt
	\caption{ Self-training by exploiting temporal consistency of video data \cite{roychowdhury2019automatic} (a)  Green boxes are the detection obtained by applying source-trained detector model. In contrast, yellow bounding boxes are obtained through tracking the bounding boxes across video, exploiting temporal consistency between adjacent frames. (b) Green boxes are high-confidence pseudo-labels and kept as is, whereas yellow boxes tracked by exploiting temporal frame consistency are assigned soft labels, i.e. instead of assigning hard category labels, they are softened through weighted addition with confidence.}
	\label{fig:auto_video} 
\end{figure}

Roychowdhry \etal \cite{roychowdhury2019automatic} proposed a self-training based approach, where they compute pseudo-labels automatically using video data. Specifically, they exploit temporal consistency between adjacent frames to track detections by the source-trained model as illustrated in Fig.~\ref{fig:auto_video}. This helps in mining more pseudo-labels which the detection model might have missed. Consequently, the pseudo-labels for the target dataset now contain predictions from the detection network and the tracker.  It is trained on both labeled source dataset and pseudo-labeled target dataset for adapting the detector to target data. Irrespective of the way it was collected (i.e. from detector model or via tracking), the target domain pseudo-labels are assigned the same category label, i.e., $y^c_t=1$ for positive class and $y^c_t=0$ for negative class. However, there are two types within the positive pseudo-labels: predictions extracted from the detection model and pseudo-labels extracted using the tracker. Both of them are assigned a score value to calculate an interpolated labels. The score assignment is given as:
\begin{equation}\label{eq:hard_labels}
s_t=\left\{\begin{array}{ll}
p_t, & \text{if pseudo-label from detector} \\
\theta, & \text{if pseudo-label from tracker}
\end{array}\right.
\end{equation}
Here, $s_t$ is the score assigned to all pseudo-labels, $p_t$ is probability obtained from detector model, $\theta$ is confidence assigned to tracker pseudo-labels where regardless of their low confidence, the score is raised up to this constant value to emphasize their importance during training. Using the score value assigned to each pseudo-label, a soft-label is obtained by interpolating the label with score values. The interpolated labels are computed as follows:
\begin{equation}
\tilde{y}^{c}_t=\lambda s_{t}+(1-\lambda) y^{c}_t,
\end{equation}
where $\lambda$ is interpolation parameter and $y^c_t$ is hard-labels given in by the Eq.~\ref{eq:hard_labels} for $i^{th}$ bounding box. Finally, the detector model is trained on both source and target domain data with both ground-truth and pseudo-labels. For any image $X^i$ the detection loss is given as:
\begin{equation}
\mathcal{L}^i_{det}=\{\begin{array}{ll}
\mathcal{L}^{frcnn}_{det}(Y^i_s, p^i_s), & \text { if } X^i \in \mathcal{S} \\
\mathcal{L}^{frcnn}_{det}(\tilde{Y}^i_t, p^i_t), & \text { if } X^i\in \mathcal{T}
\end{array},
\end{equation}
where $\mathcal{L}^{frcnn}_{det}$ is Faster-RCNN detection loss, $Y_s$ denotes annotation corresponding to respective source domain image that contains both bounding box and category label, and $\tilde{Y}_t$ denotes pseudo-label corresponding to respective target domain image containing both bounding box information and interpolated category label $\tilde{y}^c_t$. As the training progresses, tracker-based pseudo-labels' emphasis helps in improving adaptation to target domain data.
%Note that their approach is demonstrated only on single object datasets for vehicle detection or face detection task. 

\begin{figure}[h!]
	\begin{center}
		\includegraphics[width=0.99\linewidth]{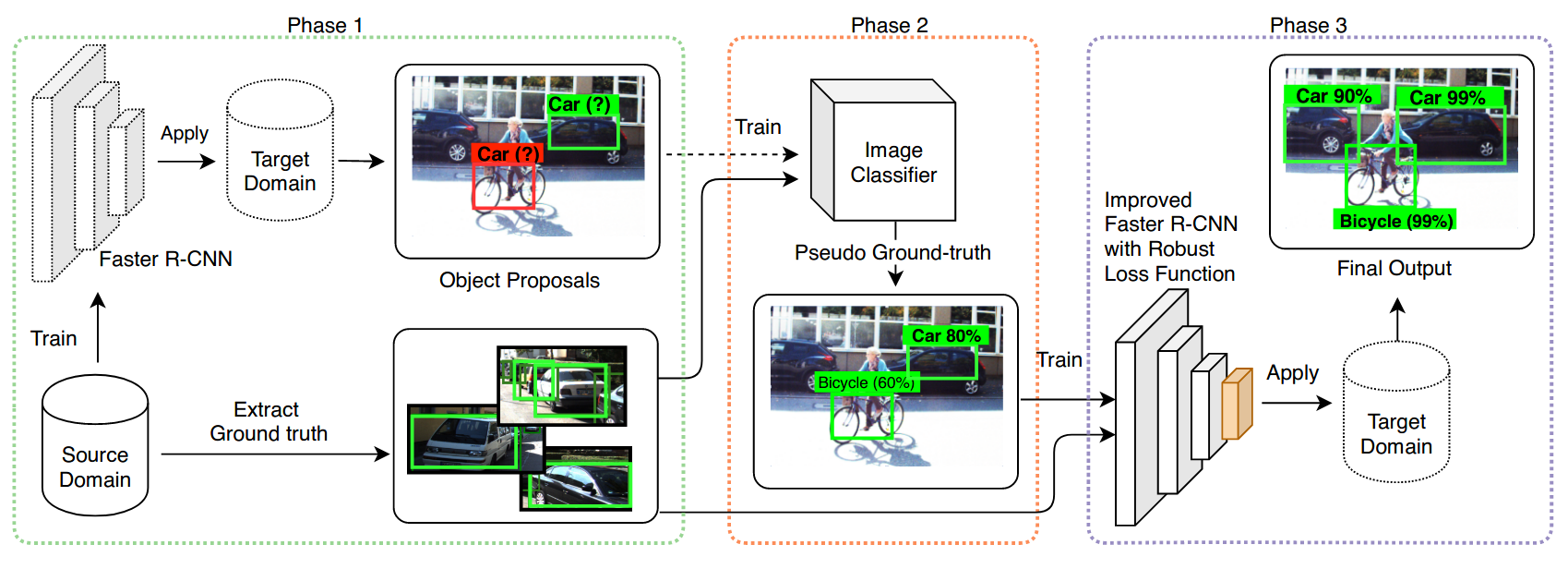}
	\end{center}
	\vskip -15.0pt
	\caption{Robust Faster-RCNN \cite{khodabandeh2019robust} utilizes the source-trained model to obtain noisy labels on the target domain dataset. A classification module is used to update the noisy labels and improve their quality. Finally, the detection model is trained on the target domain data with the help of a robust loss function that incorporates the uncertainty to account for possible mistakes present in the noisy pseudo-labels obtained through the source-trained model.}
	\label{fig:robust_learning} 
\end{figure}

Although mining pseudo-labels by exploiting the temporal consistency of video data is an effective strategy to obtain more annotations, it relies heavily on the availability of video data which is not always possible. Hence, it is important to develop single-image based pseudo-label training strategy. Khodabandeh \etal \cite{khodabandeh2019robust} proposed a method to specifically address the noise present in the pseudo-labels for the single-image target dataset. Their approach, termed as Robust Faster-RCNN, follows a three-phase training strategy. All phases are as illustrated with block diagrams corresponding to each phase in Fig.~\ref{fig:robust_learning}. In the first phase, the detection network is trained with a supervised detection loss $\mathcal{L}_{det}^{frcnn}$ using a source domain dataset that has access to ground-truth annotations. In the second phase, the source-trained detector model is used to obtain pseudo-labels for the target domain dataset. Subsequently, the pseudo-labels are further refined using an additional classifier network that is pre-trained on a large-scale classification dataset. The refinement strategy utilizes both detector model prediction and classifier network predictions. With the help of refined labels, phase three involves training the detector model with a newly designed loss that accounts for potential noise in the pseudo-labels and helps the detector learn better. Let us consider, $\textbf{p}_c$ and $\textbf{p}_c^{img}$ be the probability vector of a prediction from detector model and pre-trained image-classification network, respectively. Furthermore, let $\textbf{s}_c$ and $\textbf{s}_c^{img}$ denote the logits (score vectors) of a prediction from detector model and pre-trained image-classification network, respectively. Also, $\tilde{\textbf{b}}^{pseudo}$ and $\bar{\textbf{b}}^{current}$ denote the bounding box pseudo-label collected in the first phase and current bounding box prediction by the detector model, respectively. The Robust Faster-RCNN \cite{khodabandeh2019robust} training utilizes these predictions to refine the pseudo-annotations, which are then used as supervision to train the detection model on the target data. The following equations describe the refinement process:
\begin{equation}\label{eq:refine}
\begin{array}{ll}
\textbf{p}^{refined}_{c} \ = \ \operatorname{softmax}((\textbf{s}_{c}+ \ \alpha \ \textbf{s}_{img}) /(1+\alpha)), \\
\textbf{b}^{refined} \ = \ (( \ \tilde{\textbf{b}}^{ \ current} \ + \ \alpha \ \bar{\textbf{b}}^{ \ pseudo} \ ) /(1+\alpha)),
\end{array}
\end{equation}
where $\alpha$ is a hyper-parameter that controls the trade-off between the two terms. $\alpha$ starts with a large value and it is gradually decreased to smaller values as the third phase training progresses \cite{khodabandeh2019robust}. The theoretical formulation that lead to the refinement equations Eq.~\ref{eq:refine} can be found  in \cite{khodabandeh2019robust}. Training with refined annotations help the detector model counter the noise in the pseudo-labels much better.  

\begin{figure}[b!]
	\begin{center}
		\includegraphics[width=1.00\linewidth]{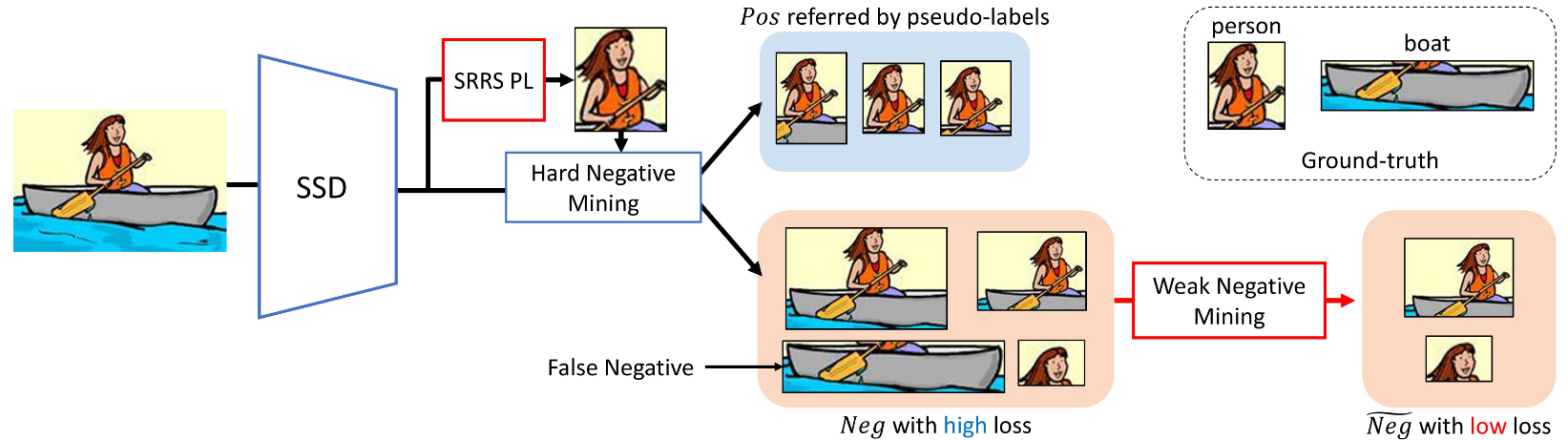}
	\end{center}
	\vskip -15.0pt
	\caption{Self-training one-stage detector \cite{kim2019self} utilizes a pseudo-label based self-training to adapt SSD based detector model to target domain data. The pseudo-labels for training are generated following an eligibility criteria termed as, Supporting Region-based Reliable Score (SRRS) in the paper. Additionally, hard negative mining is used to get positive and negative examples which are later used to perform self-training of the SSD-based detector model.}
	\label{fig:st_ssd} 
\end{figure}

Previous methods to perform self-training were based on the Faster-RCNN framework and relied on subsequent assumptions to handle the noise in the obtained pseudo-labels. Kim \etal \cite{kim2019self} proposed a strategy that focused on performing self-training of one-stage object detector models, specifically addressing it for the SSD-based model. The authors perform hard negative mining of pseudo-labels followed by a weak negative mining strategy as shown in Fig.~\ref{fig:st_ssd}. During the hard mining phase, a threshold $\delta$ is utilized to filter out the false-negative bounding box proposals based on their IoU with final bounding box prediction. Weak mining phase further improves the pseudo-labels by reducing the risk of false positives with the help of assigning instance-level scores calculated using Support Region-based Reliable Score $(\operatorname{SRRS})$. The score is calculated by a set of RoIs predicted by the detection model that satisfies the hard mining threshold $\delta$. If there are $N_{support}$ RoIs that satisfy the $\delta$-IoU threshold corresponding to the any detector bounding box prediction $\tilde{\textbf{b}}$ having confidence score of $\tilde{p}_b$, the $\operatorname{SRRS}$ score corresponding to the respective bounding box prediction can be calculated as:
\begin{equation}
\operatorname{SRRS}(\tilde{\textbf{b}})=\frac{1}{N_{support}} \sum\limits_{i=1}^{N_{support}} \text{IoU}(\textbf{b}_{i}, \ \tilde{\textbf{b}}) \cdot p_b.
\end{equation}
Based on the $\operatorname{SRRS}$ scores, the pseudo-labels are further filtered with another threshold $\upsilon$ to reduce the false positives. They also employ an additional regularization, termed in the paper as Adversarial Background Score Regularization (ABSR). This regularization utilizes the gradient reversal layer discussed in Sec.~\ref{subsubsec:ganin}; however, it is only applied on the target domain images. Without this regularization, the detector model risks incorrect prediction with high confidence. The addition of ABSR restricts the classifier sub-network of SSD to produce such overconfident predictions by avoiding alignment to non-transferable background regions. Together with pseudo-labels mined through hard and weak mining phase and the regularization loss, the detector model is trained in an end-to-end fashion. The self-training is performed by minimizing  $\mathcal{L}^{ssd}_{det}$ as discussed in Sec.~\ref{subsubsec:ssd}, where the loss is computed using the mined pseudo-labels.

Apart from these methods, there are many other interesting works available as pre-print \cite{yu2019unsupervised}, \cite{munir2021sstn}, \cite{wang2021robust} which utilize pseudo-label based self-training strategy for adapting detection model to target domain.

%%%%%%%%%%%%%%%%%%%%%%%%%%%%%%%%%%%%%%%%%%%%%%%%%%%%%%%%%%%%%%%%%%%%% IMAGE-TO-IMAGE TRANSLATION %%%%%%%%%%%%%%%%%%%%%%%%%%%%%%%%%%%%%%%%%%%%%%%%%%%%%%%%%%%%%%%%%%%%%

\subsection{Image-to-image translation}\label{subsec:image_to_image_translation}

\begin{figure}[h!]
	\begin{center}
		\includegraphics[width=0.95\linewidth]{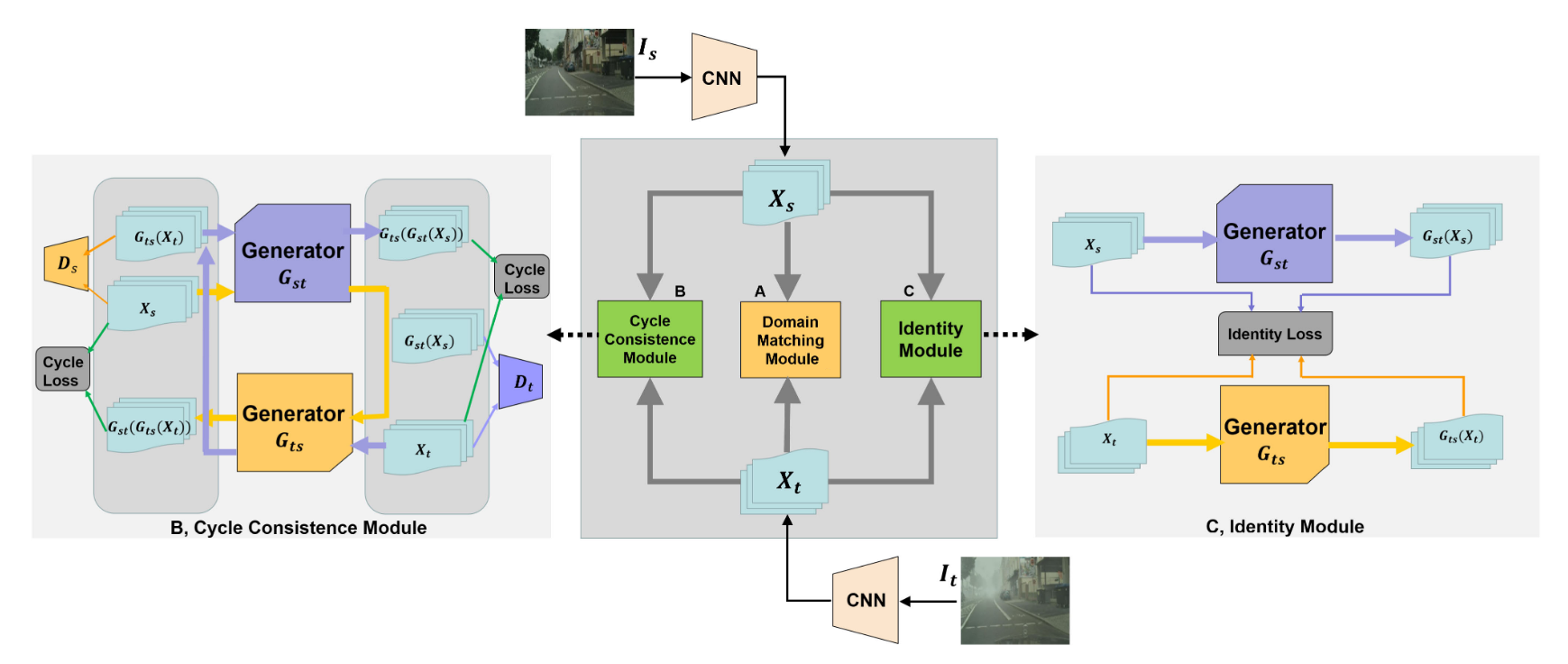}
	\end{center}
	\vskip -15.0pt
	\caption{Zhang \etal \cite{zhang2019cycle} extended the domain adaptive Faster-RCNN proposed in \cite{chen2018domain} with the addition of Cycle-GAN based image-to-image translation module at the input-level. The image-to-image translation module training is shown in the figure and domain matching module corresponds to the feature alignment strategy proposed in \cite{chen2018domain}.}
	\label{fig:ca_faster} 
\end{figure}

As discussed earlier, the primary issue in unsupervised visual domain adaptation is that the target domain images are visually very distinct from the source domain. This causes a large gap in the feature space of the source-trained detector network, resulting in poor performance on the target domain. The method falling under adversarial feature learning (discussed in Sec.~\ref{subsec:adv_feat_learning}) and pseudo-label based self-training (discussed in Sec.~\ref{subsec:self_training_pseudo_labels})  attempt to learn a feature representation that is more suited to perform detection on the target domain images. In contrast to feature alignment approaches, image-to-image translation-based methods to mitigate the domain gap at the input level. One of the most popular strategies used by image-to-image translation-based adaptation strategy is to use an unpaired image-to-image translation algorithm like Cycle-GAN \cite{zhu2017unpaired, yi2017dualgan}, UNpaired Image-to-image Translation \cite{liu2017unsupervised}, Multi-modal UNIT \cite{huang2018multimodal}. Methods that utilize the image-to-image translation-based approach often utilize additional strategies like self-training or adversarial feature learning to further boost the target detection performance. Image-to-image translation helps in bridging the gap at the input level while feature level adversarial alignment ensures a shared feature space for the target and source domain. Zhang \etal \cite{zhang2019cycle} specifically utilize Cycle-GAN to learn a mapping function between source and target domain images, in a method termed as Cycle-consistent Adaptive Faster-RCNN (CA-FRCNN) and illustrated in Fig.~\ref{fig:ca_faster}. Their approach extends the DA-Faster approach proposed in \cite{chen2018domain} (discussed in Sec.~\ref{subsubsec:det_adv_train}) by adding an image-to-image translation at the input level while adding other losses like gradient reversal at an instance and image-level intact. As shown in their experiments \cite{zhang2019cycle}, the addition of an image-translation module further enhances the performance.

\begin{figure}[h!]
	\begin{center}
		\includegraphics[width=0.95\linewidth]{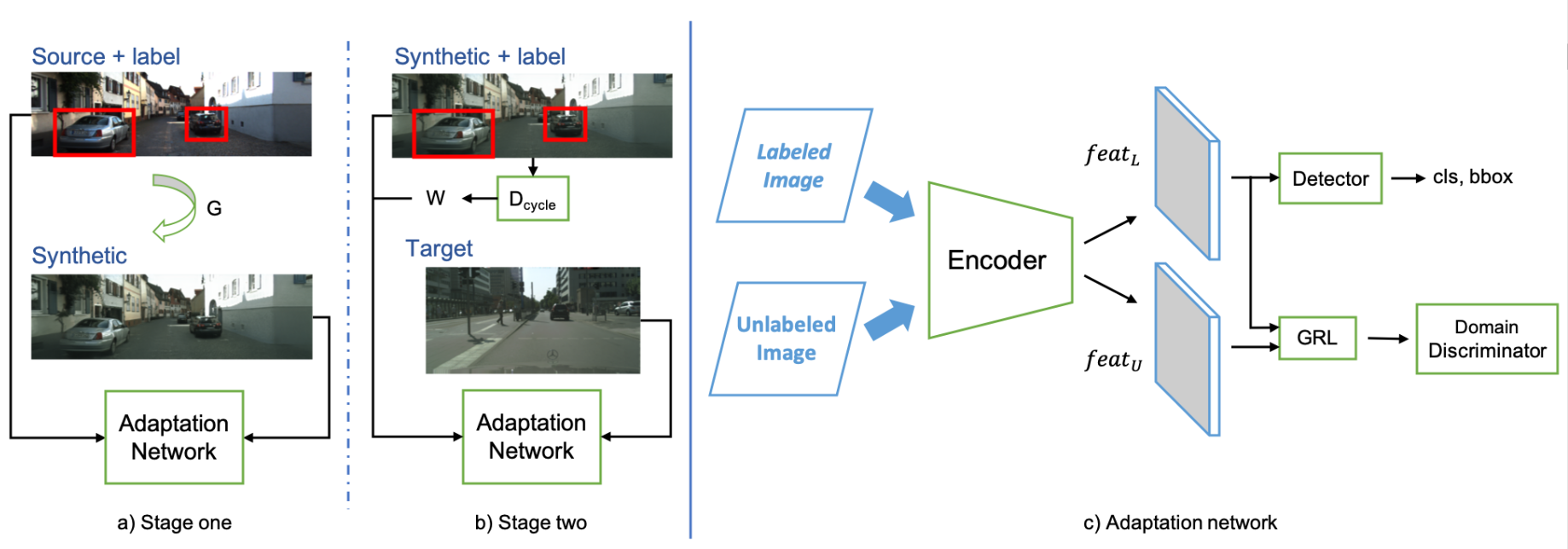}
	\end{center}
	\vskip -15.0pt
	\caption{Hsu \etal \cite{hsu2020progressive} proposed a progressive adaptation strategy that tries to reduce the gap between source and target domain at both input-level and feature-level. Specifically, before the detector training for the target domain, Hsu \etal \cite{hsu2020progressive} trains a Cycle-GAN model that can map source images to the target domain and vice versa. The Cycle-GAN model is then applied to target domain images to get translated images. Additionally, the gradient reversal layer-based learning strategy is used that further cuts the gap in the feature domain between the source domain and translated target domain, as shown in the figure.}
	\label{fig:progressive_da} 
\end{figure}

\begin{figure}[t!]
	\begin{center}
		\includegraphics[width=0.95\linewidth]{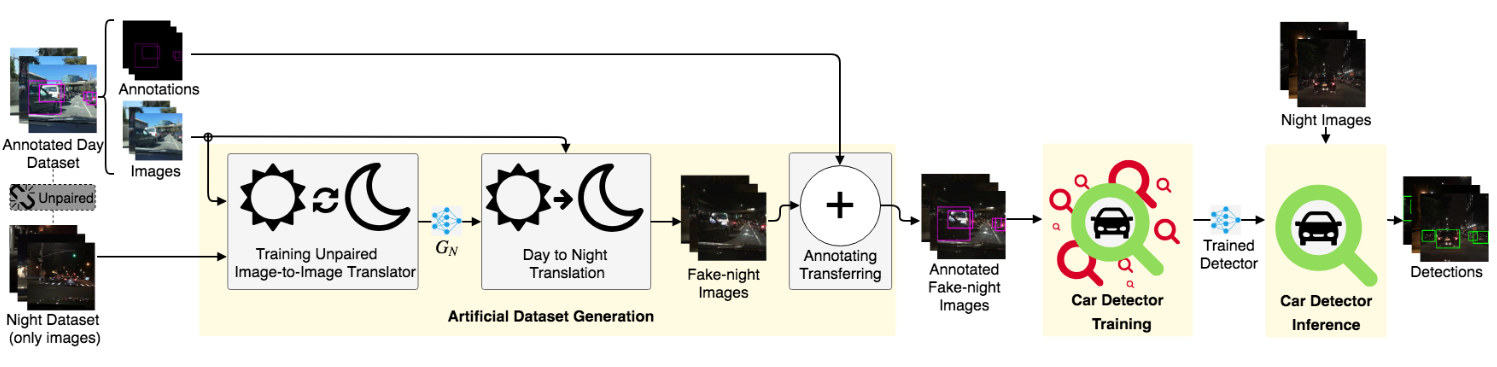}
	\end{center}
	\vskip -15.0pt
	\caption{Arruda \etal \cite{arruda2019cross} utilized image-to-image translation based strategy to adapt detector model from daylight scenes to nighttime. As illustrated in the figure, a Cycle-GAN based module is used to translate daylight images to a fake nighttime one. The daylight images have ground-truth annotations and can train the detector with fake-nighttime in a supervised fashion. This results in a better adapted detector model for the nighttime scenario.}
	\label{fig:day_to_night} 
\end{figure}

Hsu \etal \cite{hsu2020progressive} proposed a progressive adaptation strategy where they follow a similar strategy of using image-level translation. As shown in Fig.~\ref{fig:progressive_da}, a Cycle-GAN based image-to-image translation module is utilized to translate target images into source-like images. Then with the help of the gradient reversal layer, the residual domain gap is further reduced in the feature space. Furthermore, with progressive adaptation \cite{hsu2020progressive},  they showed that applying gradient reversal at only image-level is sufficient for adaptation rather than applying both image-level and instance-level losses. Arruda \etal \cite{arruda2019cross} utilized the image translation module to specifically address the domain gap between daylight (source domain) and night-time (target domain) data, as shown in Fig.~\ref{fig:day_to_night}. Once the image translation module is ready, it is used to translate all daylight data to create a fake night-time dataset. Since ground-truth annotations are available for the daylight images, they can be used to supervise the detector model with fake night-time images as input. Such training will help learn a detector model that is better suited for the night-time scenario as compared to a model trained with fully supervised daylight images.

\begin{figure}[h!]
	\begin{center}
		\includegraphics[width=0.95\linewidth]{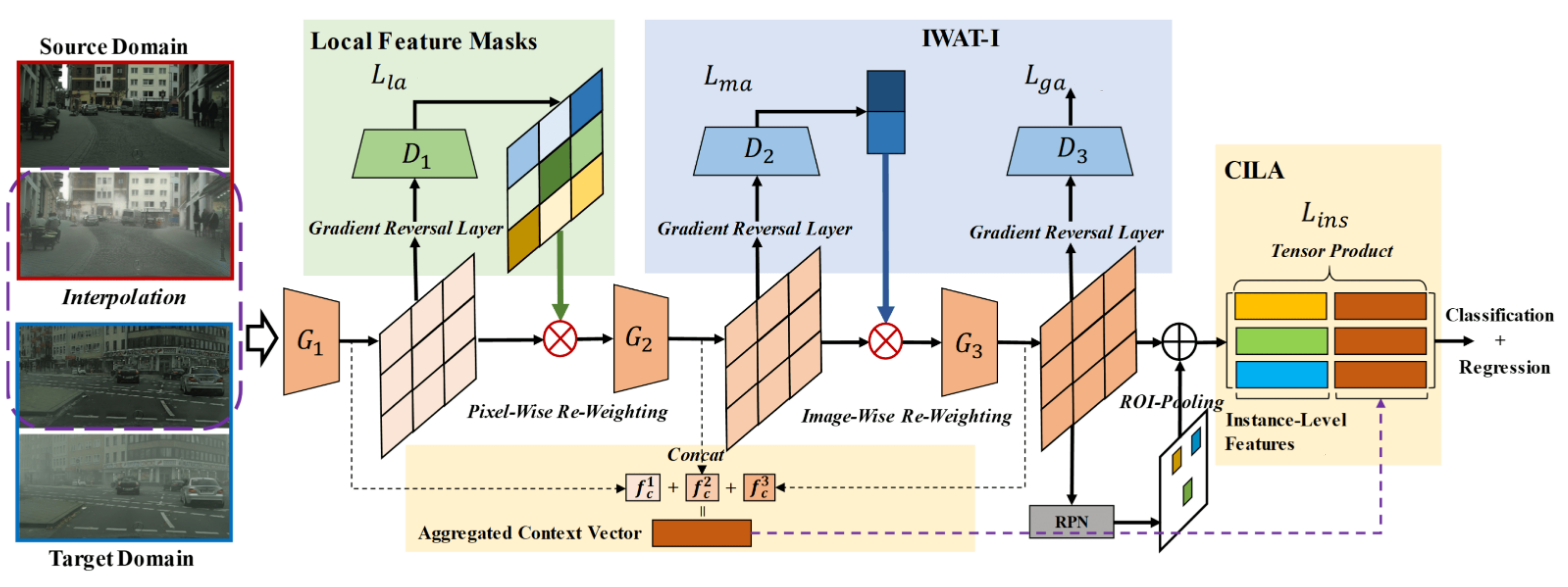}
	\end{center}
	\vskip -15.0pt
	\caption{Chen \etal \cite{chen2020harmonizing} proposed a Cycle-GAN \cite{zhu2017unpaired} based pre-processing step is introduced to cut the style gap between source and target domain images. Both source and target domain images are translated into ``interpolated domain'' to fill-in the input-level domain gap. Furthermore, they proposed harmonizing and transferability based adaptation of object detector that adopts multiple previously proposed strategies such as image-level and instance-level gradient reversal training similar to \cite{chen2018domain}, context vectors from discriminators \cite{saito2019strong}, and multiple image-level feature alignment. Here, IWAT-I denotes Importance Weighted Adversarial Training with Interpolated inputs, CILA denotes Context-aware Instance Alignment module, ${L}_{la}$ denotes pixel-wise adaptation loss, ${L}_{ma}$ and ${L}_{ga}$ denote image-wise masked and global adaptation loss.}
	\label{fig:harmonizing} 
\end{figure}

Chen \etal \cite{chen2020harmonizing} combined several of the previously proposed strategies along with image-to-image translation ideas in their approach termed as Harmonizing Transferability Calibration Network (HTCN). The proposed approach is illustrated in Fig.~\ref{fig:harmonizing}. First, a Cycle-GAN based model learns a mapping between the source and the target domain. HCTN utilizes this image-translation module to create source-like target images and target-like source images, also termed as ``interpolated images''. The ``interpolated images'' helps HTCN fill in the distribution gap between domains and consequently reduce the  source-bias of the decision boundaries. Additionally, HTCN employs local alignment loss similar to the one used in \cite{saito2019strong} for strong local alignment (see discussed in Sec.~\ref{subsec:adv_feat_learning}). Unlike other approaches based on image-to-image translation idea that directly utilize feature map outputs of a network, HCTN uses output of the local discrimination network to weigh the feature map of ``interpolated images'' forwarded to the subsequent layers. This strategy is termed as Importance Weighted Adversarial Training with Interpolated images (IWAT-I) and helps avoid negative transfer while promoting positive transfer by appropriately weighting the feature maps. The key motivation behind IWAT-I is that not all ``interpolated images'' have equal transferability and hence appropriate weights depending on the individual images are needed that can highlight the transferable regions while suppressing the noise in the feature map. This strategy closely follows the weighted adversarial training discussed in Sec.~\ref{subsubsec:weighted_adv}. Furthermore, HCTN employs two additional discriminators to perform gradient reversal-based alignment of both masked features and global features at the output of the detection backbone network. The adversarial loss utilized in these two cases follows the formulation used in \cite{saito2019strong} for global alignment; however, HCTN uses binary cross-entropy loss as compared to focal loss used in \cite{saito2019strong}. Lastly, they perform context aware instance-level feature alignment that concatenates the features learned in the previous three discriminators with RoI-pooled features of detection network utilizing the formulation proposed in \cite{chen2018domain} for instance-level adaptation. This strategy is termed as Context-aware Instance-Level Alignment (CILA). By combining all of these losses, HCTN captures the transferable regions in the source, target and ``interpolated image'' domains to fully utilize the useful information for adapting images translated by Cycle-GAN.  Shen \etal \cite{shen2021cdtd} utilize MUNIT \cite{huang2018multimodal} based Image-to-image translation to extend the method proposed in \cite{zhu2019adapting} (discussed in Sec.~\ref{subsubsec:det_adv_train}). Specifically, MUNIT-based translation module is learned by learning to map both whole images (image-level translation) and object regions (instance-level translation) between source and target domains. Furthermore, Shen \etal \cite{shen2021cdtd} also learns a style code bank to obtain an object, background and global style vectors containing rich spatial information to aid the translation network. In addition to the MUNIT-based image translation module, discriminator networks are applied at multiple layers to perform gradient reversal-based adversarial training. Further, another set of domain classification networks are trained without any gradient reversal layer. The features learned by these domain classifiers are concatenated with RoI-pooled features to improve the classification. Shen \etal \cite{shen2021cdtd} also noted that detaching these domain classification networks to restrict the gradient flow from the main detection pipeline further improves the detection performance. Other interesting works \cite{tzeng2018splat}, \cite{li2020unsupervised}, \cite{abramov2020keep}, \cite{sun2021multi} utilizing image-to-image translation strategy can be found as pre-print.

% \begin{figure}[H]%IJCV (springer)
% 	\begin{center}
% 		\includegraphics[width=0.90\linewidth]{ ctdt}
% 	\end{center}
% 	\vskip -15.0pt
% 	\caption{night to day.}
% 	\label{fig:ctdt} 
% \end{figure}

%%%%%%%%%%%%%%%%%%%%%%%%%%%%%%%%%%%%%%%%%%%%%%%%%%%%%%%%%%%%%%%%%%%%% DOMAIN RANDOMIZATION %%%%%%%%%%%%%%%%%%%%%%%%%%%%%%%%%%%%%%%%%%%%%%%%%%%%%%%%%%%%%%%%%%%%%

\subsection{Domain randomization}\label{subsec:domain_randomization}

In the case of image-to-image translation based methods, the primary focus is to learn a mapping between source and target domain and reducing the domain gap at the input-level. In case a perfect mapping function is available that can translate target domain images to source and vice versa, the detector model can perfectly adapt to target domain without requiring any annotations. However, in practice, such mapping function are not necessarily accurate and hence even after image translation, there might still exist a domain gap between original images and translated images, e.g., source images and translated target images or vice versa. Due to this, most image-to-image translation-based approaches utilize an additional gradient reversal based feature alignment or pseudo-label based self-training. To overcome this inability of learning accurate mapping  function between source and target domain, domain randomization applies several transformations on the input image to derive multiple new domains. The goal for detection model is to then consider these new domains during training and learn feature representations that are invariant to all the domains including source and target. This strategy enforces strong constraints on feature representations of detection model as the network has to find features within images that remain invariant across a wide variety of image transformations. However, to achieve this, additional strategies such as adversarial feature learning or self-training are required. 

In summary, image-to-image translation based methods aim to learn intermediate stages where input-level domain gap is reduced to improve model performance. Whereas, domain randomization synthetically generates domains that consist of multiple distinct styles (including source and target domain style) to ensure that the detection models learn features that are useful for detection regardless of style of the input images. Let us consider a source domain dataset $\mathcal{S}=\{X^i_s, Y^i_s\}_{i=1}^{N_s}$ and target domain dataset $\mathcal{T}=\{X^i_t\}_{i=1}^{N_t}$. Domain randomization creates multiple new datasets derived from the given source domain dataset that consist of multiple distinct styles. Let us denote these source-derived stylized dataset as, $ \ \mathcal{S}^{style_m}=\{X^i_{s_m}, Y^i_s\}_{i=1}^{N_{s}}$, where $m \in \{1,\dots, M\}$ denotes $m^{th}$ style and $M$ denotes total number of unique styles. The stylization process ensures that the contents of the image are not changed and hence the object category and location information is preserved through the stylization process. All the stylized datasets are derived from source domain and hence, the $i^{th}$ stylized image $X^i_{s_m}$ can share the ground-truth annotations of corresponding $i^{th}$ image from the original source dataset $X^i_s$, for any $m \in \{1,\dots, M\}$ and $i \in \{1,\dots, N_s\}$.

% Since synthetically generated datasets are derived from source domain, same annotations can be used 

\begin{figure}[h!]
	\begin{center}
		\includegraphics[width=0.95\linewidth]{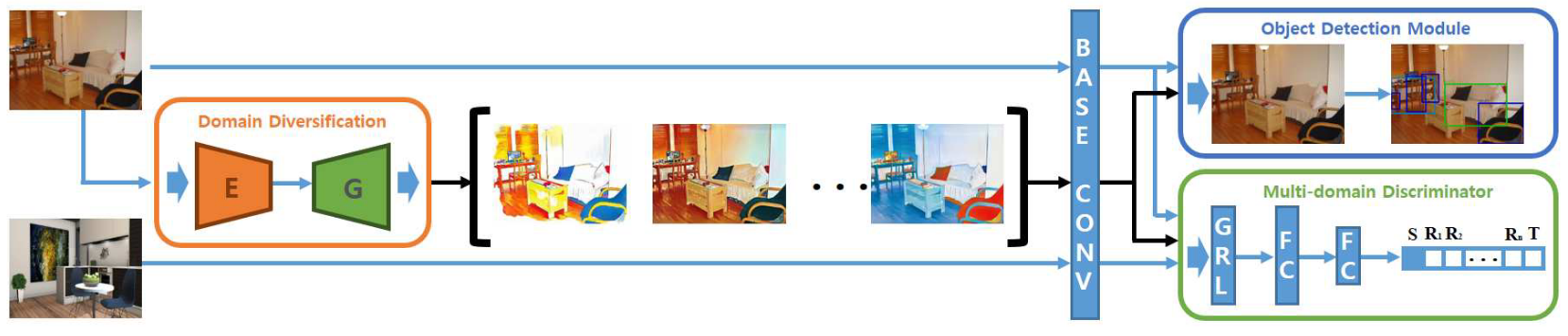}
	\end{center}
	\vskip -15.0pt
	\caption{Overview of the Diversify and Match methodology proposed by Kim \etal \cite{kim2019diversify}. The domain diversification module creates stylized version of source images with multiple distinct styles which are used to train the detection model, in addition to passing the source and target domain images. The detection model is then trained to produce style invariant features with the help of a multi-style domain discriminator and gradient reversal based training. Unlike previous methods which use binary classification, the multi-style domain discriminator performs multi-class classification into source-style, target-style and pre-defined styles created by domain diversification module.}
	\label{fig:diversify_match} 
\end{figure}

Kim \etal \cite{kim2019diversify} utilized domain randomization to adapt a Faster-RCNN based detection model. As illustrated in Fig.~\ref{fig:diversify_match}, the method has three components: domain diversification module, detection model, and multi-domain discriminator. The domain diversification module is based on generative adversarial networks \cite{goodfellow2014generative} and is tasked to take in the source domain images and shift the domain to derive a diverse set of visually distinct domains. Further, they enforce certain constraints on the output of domain diversification networks such as reconstruction, color preservation and cycle consistency. This ensures that the synthetically generated domains do not destroy the contents of the image that might negatively impact the model performance. Subsequently, the detection model is trained on these synthetically generated domains data along with the source and target domain data. Supervised detection loss, $\mathcal{L}_{det}^{frcnn}$, is applied on the source and synthetic domains to train the detector model. To ensure that the base network of detection model learns domain invariant feature representations, a multi-domain discriminator network with gradient reversal layer is employed at the end of the base network. Typically, domain discriminators are tasked to perform binary classification to identify any image as either from source or target. However, the multi-domain discriminator used in this work is tasked to perform multi-class classification to identify whether the feature representations belong to source, target, or one of the synthetically generated domains. Hence, unlike binary cross-entropy loss that is used in \cite{chen2018domain}, the multi-class cross-entropy loss is used for discriminator network. Here, the domain label for any $i^{th}$ input image are given as $y_d \in \{0, \dots, M+1\}$, where, $0$ and $M+1$ denote source and target domain, and rest of the labels denote each synthetic domains generated by domain diversification module.

\begin{figure}[b!] % BMVC paper
	\begin{center}
		\includegraphics[width=0.95\linewidth]{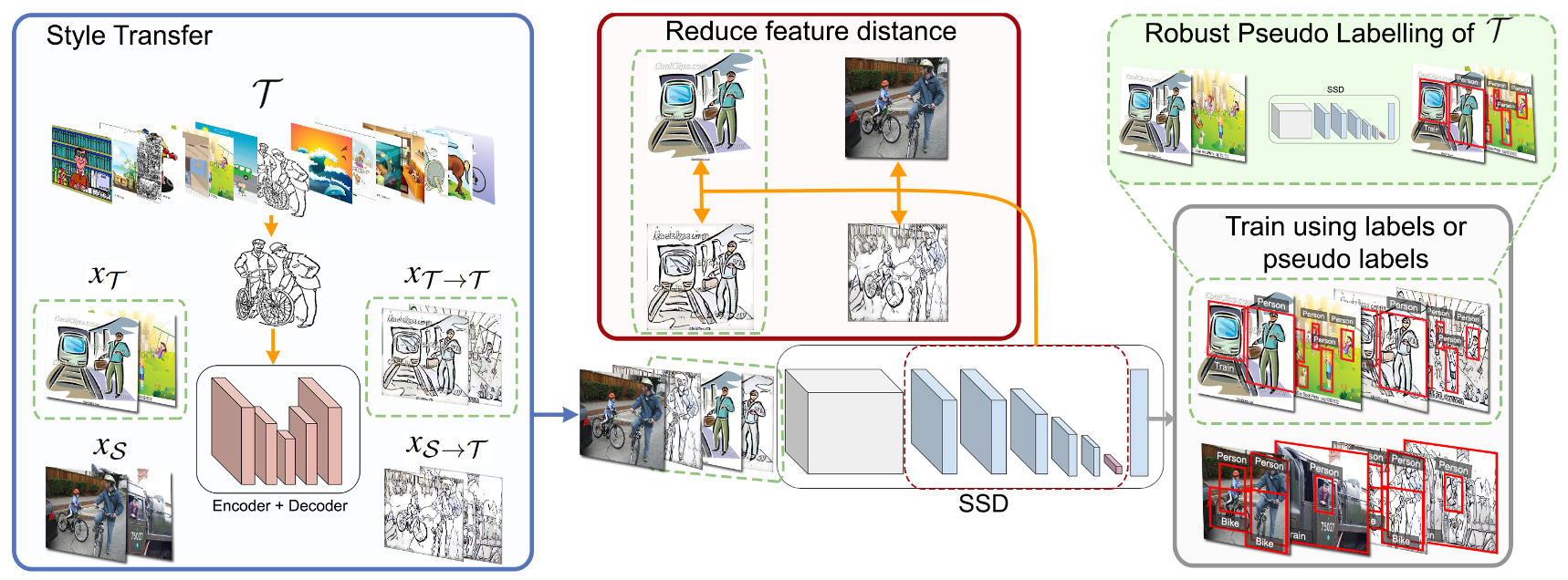}
	\end{center}
	\vskip -15.0pt
	\caption{Rodriguez \etal \cite{rodriguez2019domain} proposed to adapt SSD based detector model by training with synthetically generated dataset using image style transfer. As shown in the figure, the stylization is applied to source domain images. Subsequently, supervised source, supervised stylized-source domain, and pseudo-label supervision from the target domain are used to train the detector model to learn style/domain invariant features for object detection in target domain data.}
	\label{fig:det_via_style} 
\end{figure}

Rodriguez \etal \cite{rodriguez2019domain} proposed a domain randomization approach for adapting SSD based one-stage object detectors. The key idea, illustrated in Fig.~\ref{fig:det_via_style}, is to utilize the style transfer network proposed by \cite{huang2017arbitrary} to create a source-derived dataset with multiple distinct predefined styles. Multiple stylized version of the source-derived dataset is created with annotations borrowed from the source domain dataset. Furthermore, the source-trained detector model is evaluated on the target domain images to extract pseudo-labels, which are then used for self-training. In order to extract robust pseudo-labels for effective self-training, a positive and negative threshold is utilized to extract high-quality positive and negative examples. The base network of the detector model, denoted as ${F}_b$, is encouraged to learn domain invariant features through feature consistency loss that minimizes the $L2$-norm between feature representations extracted from source data and stylized source data. This loss, termed as feature consistency loss, is defined as:
\begin{equation}
\mathcal{L}^{feat}_{const}=\frac{1}{N_s}\sum\limits_{i=1}^{N_s}\sum\limits_{m=1}^{M}\left\|F_b\left(X^i_{s}\right)-F_{b}\left(X^i_{s_m}\right)\right\|^{2}_2.
\end{equation}

Let us denote source domain as $ \mathcal{S}=\{X^i_s, Y^i_s\}_{i=1}^{N_s}$, source-derived $m^{th}$-style dataset as $ \mathcal{S}^{style_m}=\{X^i_{s_m}, Y^i_s\}_{i=1}^{N_{s}}$ with $m \in \{1,\dots,M\}$, and target domain dataset as  $\tilde{\mathcal{T}}=\{X^i_t, \tilde{Y}^i_t\}_{i=1}^{N_t}$. Here, $M$ is the pre-defined number of styles used to create distinct source-derived dataset, $Y_s$ and $\tilde{Y}_t$ denote ground-truth source annotations and pseudo-labels for an arbitrary source/$m^{th}$-style and target image, respectively. The final training loss for the object detector can be described as:
\begin{equation}
\begin{aligned}
\mathcal{L}^{final}_{det} \ = \ \lambda_s \ \sum\limits_{i=1}^{N_s} & \ \mathcal{L}^{ssd}_{det}(X^i_s, Y^i_s) \ \\
+ \ & \sum\limits_{m=1}^{M} \lambda_{s_m} \ \sum\limits_{i=1}^{N_s} \mathcal{L}^{ssd}_{det}(X^i_{s_m}, Y^i_s) \ \\
&+ \lambda_t \sum\limits_{i=1}^{N_t} \mathcal{L}^{ssd}_{det}(X^i_t, \tilde{Y}^i_t) + \lambda_{c} \ \mathcal{L}^{feat}_{const},
\end{aligned}
\end{equation}
where $\lambda_s$, $\{\lambda_{s_m}\}_{m=1}^{M}$, $\lambda_t$, and $\lambda_{c}$ are trade-off parameters for source supervised loss, stylized-source supervised loss, pseudo-label training loss on target domain, and feature consistency loss, respectively.

Apart from these methods, there is an interesting work available as pre-print \cite{munir2020thermal}, which also utilizes domain randomization approach for adaptation of object detectors.

%%%%%%%%%%%%%%%%%%%%%%%%%%%%%%%%%%%%%%%%%%%%%%%%%%%%%%%%%%%%%%%%%%%%% STUDENT TEACHER TRAINING %%%%%%%%%%%%%%%%%%%%%%%%%%%%%%%%%%%%%%%%%%%%%%%%%%%%%%%%%%%%%%%%%%%%%

\subsection{Mean teacher training}\label{subsec:student_teacher_training}

% He \etal \cite{he2020domain} asymmetric tri-way Faster-RCNN - classification student-teacher or adversarial feature learning?

\begin{figure}[h!]
	\begin{center}
		\includegraphics[width=0.95\linewidth]{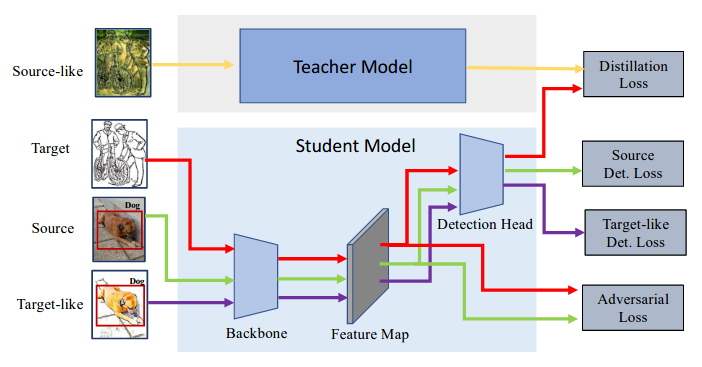}
	\end{center}
	\vskip -15.0pt
	\caption{Deng \etal \cite{deng2021unbiased} devised a strategy for cross-domain detection adaptation based on mean-teacher framework. As illustrated in the figure, it has three main components. First: image-to-image translation-based input processing on translating source into target-like images and vice versa. Second: distillation loss to transfer knowledge between teacher and student network training pipelines. Third: adversarial loss inspired from \cite{saito2019strong} to further reduce domain gap in feature space.}
	\label{fig:unbaised_mean_teacher} 
\end{figure}

Knowledge distillation has been demonstrated to be effective for exploiting unlabeled data in transfer learning \cite{hinton2015distilling, gupta2016cross, tarvainen2017mean}, semi-supervised learning \cite{berthelot2019mixmatch, sohn2020fixmatch, liuunbiased}, and domain adaptation \cite{french2018self, deng2019cluster, laine2016temporal, shu2018dirt}. Most of the domain adaptation work that utilize student-teacher training strategy has considered only the task of image classification. Their success has inspired many works which employ student-teacher training framework to perform unsupervised domain adaptation of object detector models. A recent and notable work based on this strategy is that of the unbiased mean-teacher strategy proposed by Deng \etal \cite{deng2021unbiased}, that specifically utilizes mean-teacher framework \cite{tarvainen2017mean} training for adapting object detector to the target domain. As it can be observed  from Fig.~\ref{fig:unbaised_mean_teacher}, Deng \etal \cite{deng2021unbiased} combine multiple strategies like image-to-image translation (discussed in Sec.~\ref{subsec:image_to_image_translation}) and adversarial feature learning (discussed in Sec.~\ref{subsec:adv_feat_learning}) with mean-teacher framework. First, the method trains a Cycle-GAN module to learn image mapping between source and target domain images, which is then used to create a source-like target and target-like source images. The student model pipeline is trained with the original source domain, target domain and target-like source images. Since both source and target-like source images are fully labeled,  they can be used for training using the supervised detection loss, shown in Fig.~\ref{fig:unbaised_mean_teacher} as source detection loss and target-like detection loss. Training the student pipeline with source and target domain images helps mitigate bias in the student model. Target-like source image training encourages student models to be more favorable towards target domain images. The model predictions of source-like target images are matched with model predictions of original target domain images to perform knowledge distillation. Further, the teacher parameters are updated with Exponential Moving Average as:
\begin{equation}
\Theta^i_{\mathcal{F}^{tch}}=\alpha \Theta^{i-1}_{\mathcal{F}^{tch}} \ + \ (1-\alpha) \Theta^i_{\mathcal{F}^{stu}},
\end{equation}
where $\Theta^i_{\mathcal{F}^{tch}}$ and $\Theta^i_{\mathcal{F}^{stu}}$ are network parameters for teacher and student model respectively, $\alpha$ denotes smoothing coefficient hyper-parameter that can be used to control the teacher updates, the super-script $i$ and $i-1$ denote the indices for the current and previous training iterations respectively. To further decrease the domain gap in the feature space of the student model, gradient reversal-based adversarial training involving strong local and weak global feature alignment \cite{saito2019strong}. Together with distillation, parameter updates supervised detection loss and adversarial feature training; the entire training is performed in an end-to-end fashion.
Other works that utilize the student-teacher framework include Liu \etal \cite{liuunbiased} which extended the mean-teacher framework to train an object detector in a semi-supervised fashion by exploiting large unlabeled data to improve the overall performance. Though Liu \etal focuses on semi-supervised detection with labeled and unlabeled data drawn from the similar distributions, it can be easily adopted to perform unsupervised domain adaptive training of object detectors. Cai \etal \cite{cai2019exploring} proposed cross-domain detection by utilizing a mean-teacher framework that relies on learning object relations through graph structures. Another interesting work \cite{tang2020learning}, available as a pre-print, utilizes mean-teacher framework is also available as pre-print.

%%%%%%%%%%%%%%%%%%%%%%%%%%%%%%%%%%%%%%%%%%%%%%%%%%%%%%%%%%%%%%%%%%%%% GRAPH REASONING %%%%%%%%%%%%%%%%%%%%%%%%%%%%%%%%%%%%%%%%%%%%%%%%%%%%%%%%%%%%%%%%%%%%%

\subsection{Graph-reasoning}\label{subsec:graph_reasoning}

\begin{figure}[h!]
	\begin{center}
		\includegraphics[width=0.95\linewidth]{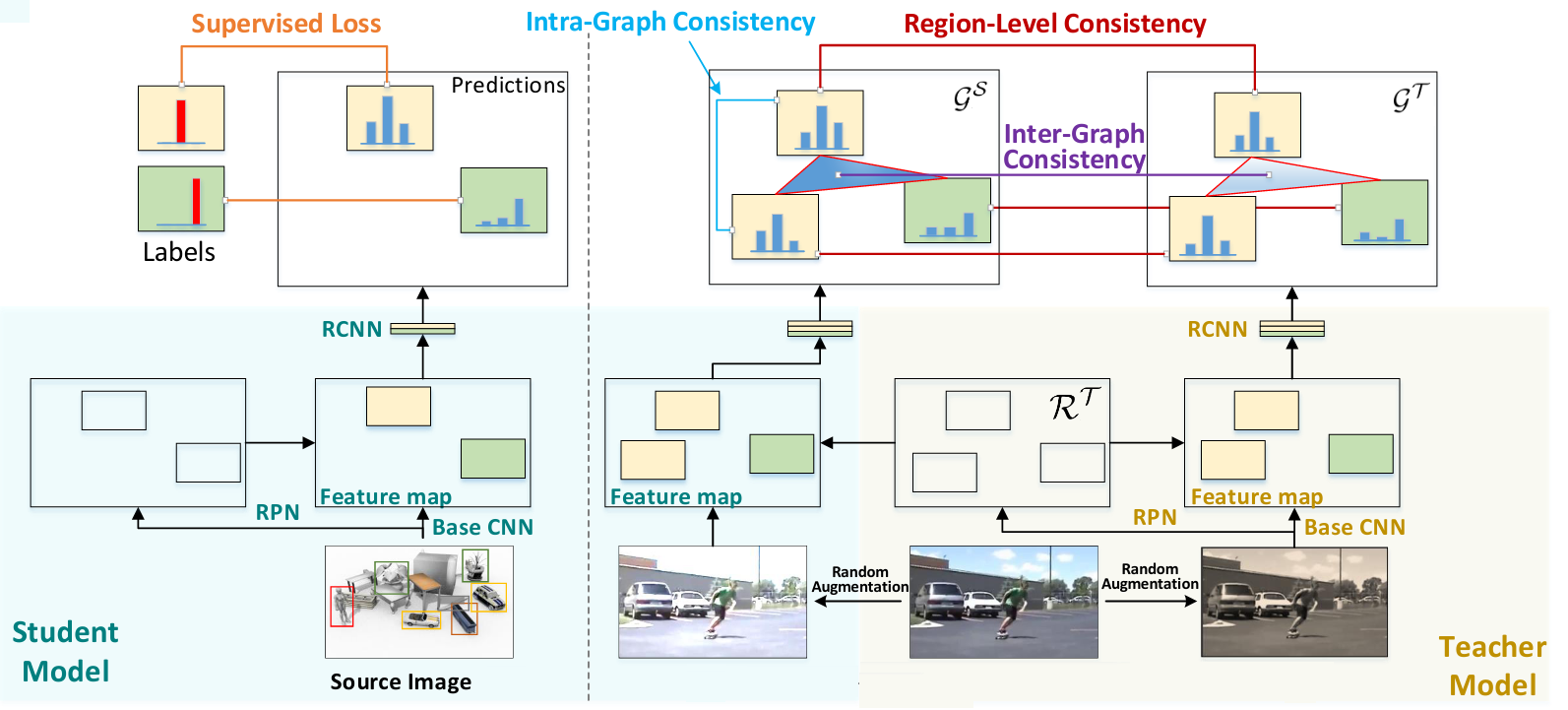}
	\end{center}
	\vskip -15.0pt
	\caption{Cai \etal \cite{cai2019exploring} explores graph reasoning on Faster-RCNN based detector model to control the information flow between source and target domain pipelines in a mean teacher framework. The source domain pipeline is used as student and target domain pipeline is used as teacher. Specifically, graph reasoning enforces \text{(1)} Regional-level consistency to align regional level predictions between teacher and student, \text{(2)} Inter-graph consistency to match the graph structures between between source and target pipelines, and \text{(3)} Intra-graph consistency that promotes the similarity between same categories within graph constructed from source domain model.}
	\label{fig:mtor} 
\end{figure}

Graph-reasoning is a widely popular technique in the computer vision community for visual recognition \cite{kipf2016semi, yan2019learning}, video understanding \cite{yan2018spatial, yan2019fine} etc. It is also utilized extensively for unsupervised adaptation of classification networks \cite{ma2019gcan, mancini2019adagraph}. The ability of graphs to model inter-image and intra-image relationships of underlying object categories can be extremely useful in the case of adapting object detectors. Furthermore, modeling graph structures is mutually exclusive and hence can be combined with other existing adaptation strategies. Motivated by these benefits, Cai \etal \cite{cai2019exploring} proposed a domain adaptive object detection approach, termed as Mean-Teacher Object Relations (MTOR), that utilizes graph-reasoning based strategy combined with student-teacher training. As illustrated in the Fig.~\ref{fig:mtor}, MTOR consists of four major components: \textit{1)} Regional-Lelev Consistency (RLC), \textit{2)} Inter-Graph Consistency (InterGC), \textit{3)} Intra-Graph Consistency (IntraGC), \textit{4)} Mean-Teacher Training (MTT). Here, the first three components are devoted to graph structures that model relationship between the objects within an image. Given a source domain dataset, $\mathcal{S}=\{X^i_s, Y^i_s\}_{i=1}^{N_s}$, target domain dataset, $\mathcal{T}=\{X^i_t\}_{i=1}^{N_t}$, detection model, $F$, the base network, $F_{base}$, teacher network, $F^{t}$ and student network $F^{s}$, a graph, $\mathcal{G}_{X_t} = \{\mathcal{V}_{X_t}, \mathcal{E}_{X_t}\}$ is constructed. Here $\mathcal{V}_{X_t}$ are vertices denoting set of region proposal predictions by detection network and $\mathcal{E}_{X_t}$ ($|\mathcal{V}_{X_t}| \times |\mathcal{V}_{X_t}|$) is an affinity matrix of the corresponding the graph. Note that $\mathcal{G}^t_{X_t}$ and $\mathcal{G}^s_{X_t}$ correspond to the graphs constructed on target domain images using teacher and student models, respectively. Each vertex in the graph corresponds to a region proposal and is assigned a probability vector denoting the probability of that region belonging to one of the $K_{X_t}$ categories. Before selecting the region proposal, they are filtered by eliminating all proposals having a maximum probability value falling below a threshold. For any arbitrary target image ${X_t}$, consider that there are $N_b$ such region proposals and for $i^{th}$ region proposal, the corresponding probability vector is $\textbf{p}^i_c$, the region level consistency loss is defined as:
\begin{equation}
\mathcal{L}_{t}^{R C L}=\frac{1}{N_b} \sum\limits_{i=1}^{N_b} \left\|\textbf{p}^{t_i}_c-\textbf{p}^{s_i}_c\right\|_{2}^{2},
\end{equation}
where $\textbf{p}^{t_i}_c$ and $\textbf{p}^{s_i}_c$ denote probability vectors extracted from teacher and student model and correspond to teacher graph $\mathcal{G}^t_{X_t}$ and student graph $\mathcal{G}^s_{X_t}$, respectively.

\begin{figure}[b!]
	\begin{center}
		\includegraphics[width=0.95\linewidth]{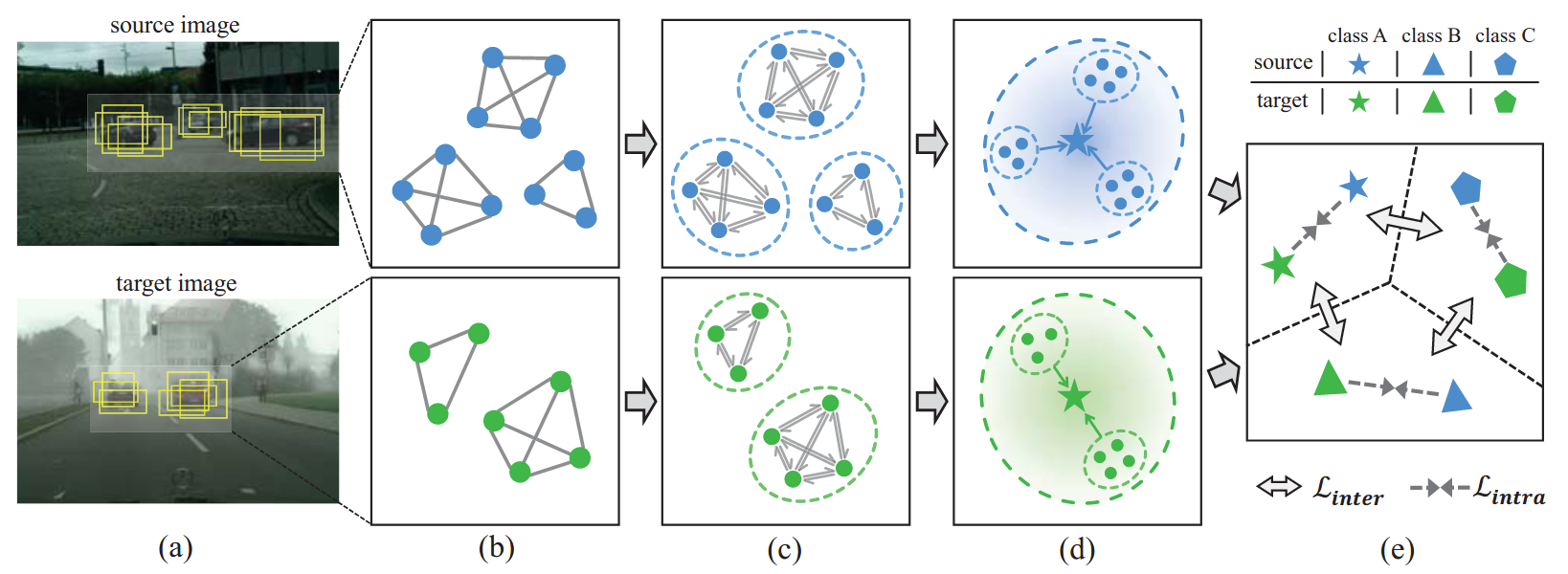}
	\end{center}
	\vskip -15.0pt
	\caption{The figure provides an overview of the method proposed by Xu \etal \cite{xu2020cross}. (a) Source and target domain images with respective object region proposals. (b) Utilizing the proposals to create a relation graph for both source and target image. (c) Information propagation between proposals of the same category to get accurate instance-level feature representations. (d) Merging category-wise information with confidence weighting to create a prototype for the respective object class. (e) Category-level adversarial feature alignment by enhancing intra-class compactness $\mathcal{L}_{intra}$ and inter-class separability $\mathcal{L}_{inter}$.}
	\label{fig:graph_prototype} 
\end{figure}

To compute the inter-graph consistency, InterGC, the affinity matrix of both student and teacher graph is calculated using cosine similarity. If we denote teacher graph as $\mathcal{G}^t_{X_t} = \{\mathcal{V}^t_{X_t}, \mathcal{E}^t_{X_t}\}$ for any arbitrary target image ${X_t}$ having $N_b$ region proposals, where any $i^{th}$ region proposal consists of the corresponding RoI-pooled feature $\textbf{f}^i_t$ extracted from the teacher model. The affinity matrix is calculated using the feature similarity between any two proposal features in a given image and is defined as:
\begin{equation}\label{eq:mtor_affinity}
\mathcal{E}^t_{X_t}(i,j)=\frac{\textbf{f}_{t}^{i} \cdot \textbf{f}_{t}^{j}}{\|\textbf{f}_{t}^{i}\|_{2} \|\textbf{f}_{t}^{j}\|_{2}}.
\end{equation}
Similarly, Eq.~\ref{eq:mtor_affinity} can be used to calculate affinity matrix for student model $\mathcal{E}^s_{X_t} = \{\mathcal{V}^s_{X_t}, \mathcal{E}^s_{X_t}\}$ for the same target domain image. Once we have both student and teacher model affinity matrix for a given target domain image, the inter-graph consistency loss is calculated as follows:
\begin{equation}
\mathcal{L}_{t}^{InterGC}=\frac{1}{N_b} \cdot\left\|\mathcal{E}^t_{X_t}-\mathcal{E}^{s}_{X_t}\right\|_{2}^{2}.
\end{equation}
This inter-graph consistency enforces alignment of the graph structure between student and teacher pipeline for a target image. Within each image, there can be multiple categories that are semantically related. Intra-graph consistency reinforces the similarity between proposals of the same category for student graphs with supervision from the teacher. Specifically, the teacher model is used to predict whether two region proposals contain an object of the same category. If $i^{th}$ and $j^{th}$ proposal contain same category, label $y^{graph}_{ij}$ is assigned a value of $1$ or $0$ otherwise. Given affinity matrix of student model for an arbitrary target image, $\mathcal{E}^s_{X_t}$, the intra-graph consistency loss is defined as:
\begin{equation}
\mathcal{L}_{t}^{IntraGC}=\frac{\sum\limits_{i,j} \ y^{graph}_{ij} \cdot (1 \ - \ \mathcal{E}^s_{X_t}(i, j))}{\max(1, \ \sum\limits_{i,j} \ y^{graph}_{ij})}.
\end{equation}
In addition to these graph-based consistency losses, the student network is trained with supervised detection loss. The teacher network parameter updates follow mean-teacher formulation discussed in Sec.~\ref{subsec:student_teacher_training}.

Recently, Xu \etal \cite{xu2020cross} proposed a graph-based adaptation approach where they construct a relational graph between object proposals to compute category prototypes that help alignment between source and target domain. The training strategy involves graph construction and alignment as illustrated in Fig.~\ref{fig:graph_prototype}. The proposed method is based on the Faster-RCNN detection framework and utilizes its two-stage detector training to compute relational graphs. The relational graphs are used to compute category-wise prototypes, which are then matched to enforce alignment between source and target domain at each stage of the detector model. The overall alignment strategy is termed as Graph-induced Prototype Alignment (GPA) in the paper \cite{xu2020cross}. Let us consider that, for an arbitrary image, the detection model produces $N_b$ number of region proposals with varying probabilities of containing an object of any given category in the dataset. Based on these proposals, a graph $\mathcal{G} = \{\mathcal{V}, \mathcal{E}\}$ is constructed with $\mathcal{V}$ as nodes and $\mathcal{E}$ denoting adjacency matrix/edge-values assigned to the graph. Each node represents a region proposal predicted by the detection model and the adjacency matrix values are computed by calculating Intersection over Union (IoU) between pairs of region proposals. Specifically, let's consider two region proposals for an arbitrary input image denoted as bounding boxes $\textbf{b}_i$ and $\textbf{b}_j$. The edge value between them is computed using the following equation:
\begin{equation}
\mathcal{E}_{ij}=\operatorname{IoU}(\textbf{b}_{i}, \textbf{b}_{j})=\frac{\textbf{b}_{i} \bigcap \textbf{b}_{j}}{\textbf{b}_{i} \bigcup \textbf{b}_{j}},
\end{equation}
where $\bigcap$ denotes intersection operation and $\bigcup$ denotes union operation. This creates an adjacency matrix of size $N_b \times N_b$. Once the relation graph is constructed, the adjacency matrix is used to aggregate the RoI-pooled features and their respective class probabilities for each region proposals. Let us denote each RoI-pooled feature having $d$ dimension as $\textbf{f} \in \mathbb{R}^{1 \times d}$ and respective probability vector as $\textbf{p} \in [0, \ 1]^{1 \times K}$. We can stack these features and probability values to create feature matrix $\textbf{F} \in \mathbb{R}^{N_b \times d}$ and probability matrix $\textbf{P} \in [0, \ 1]^{N_b \times d}$. The aggregated features $\tilde{\textbf{F}}$ and probabilities $\tilde{\textbf{P}}$ can be represented as:
\begin{equation}
\begin{array}{l}
\widetilde{\textbf{F}}=\textbf{D}^{-\frac{1}{2}} \ \mathcal{E} \ \textbf{D}^{-\frac{1}{2}} \textbf{F},\\
\widetilde{\textbf{P}}=\textbf{D}^{-\frac{1}{2}} \ \mathcal{E} \ \textbf{D}^{-\frac{1}{2}} \textbf{P},
\end{array}
\end{equation}
where matrix $\textbf{D}$ is a diagonal matrix with diagonal elements given as $\textbf{D}_{ii}=\sum_{j}\mathcal{E}_{ij}$. This aggregation updates the feature and probability vectors to represent the instance-level information accurately. Furthermore, the updated features and probabilities are used to calculate category-wise prototypes. The prototype feature for $k^{th}$ category can be computed as:
\begin{equation}
\textbf{v}_{k}=\frac{\sum\limits_{i=1}^{N_{b}} \tilde{\textbf{P}}_{ik} \cdot \tilde{\textbf{F}}_{i}^{T}}{\sum\limits_{i=1}^{N_{b}} \tilde{\textbf{P}}_{ik}},
\end{equation}
where $\textbf{v}_k \in \mathbb{R}^{1\times d}$ denotes the prototype corresponding to $k^{th}$ category and are calculated for both source and target domain. The alignment is performed by matching source and target prototypes by minimizing the distance between prototypes of the same category while maximizing the distance between different categories. Additionally, the same category prototypes are matched between source and target domain by minimizing respective category prototypes.

%%%%%%%%%%%%%%%%%%%%%%%%%%%%%%%%%%%%%%%%%%%%%%%%%%%%%%%%%%%%%%%%%%%%%%%%%%%%%%%%%%%%%%%%%%

\section{Evaluation protocols}
In this section, we discuss the details of the evaluation protocol used in evaluating the performance of the various domain adaptive detection approaches discussed earlier. First, we provide details of  various benchmark datasets that are used for the evaluation purpose, followed by a discussion of  various adaptation settings such as synthetic-to-real, clean-to-adverse weather, etc. Next, we define the metrics used for evaluation and comparison. Finally, we provide a detailed discussion of the results of various methods. 

\subsection{Datasets}

\begin{figure}[h!]
	\begin{center}
		\includegraphics[width=1\linewidth]{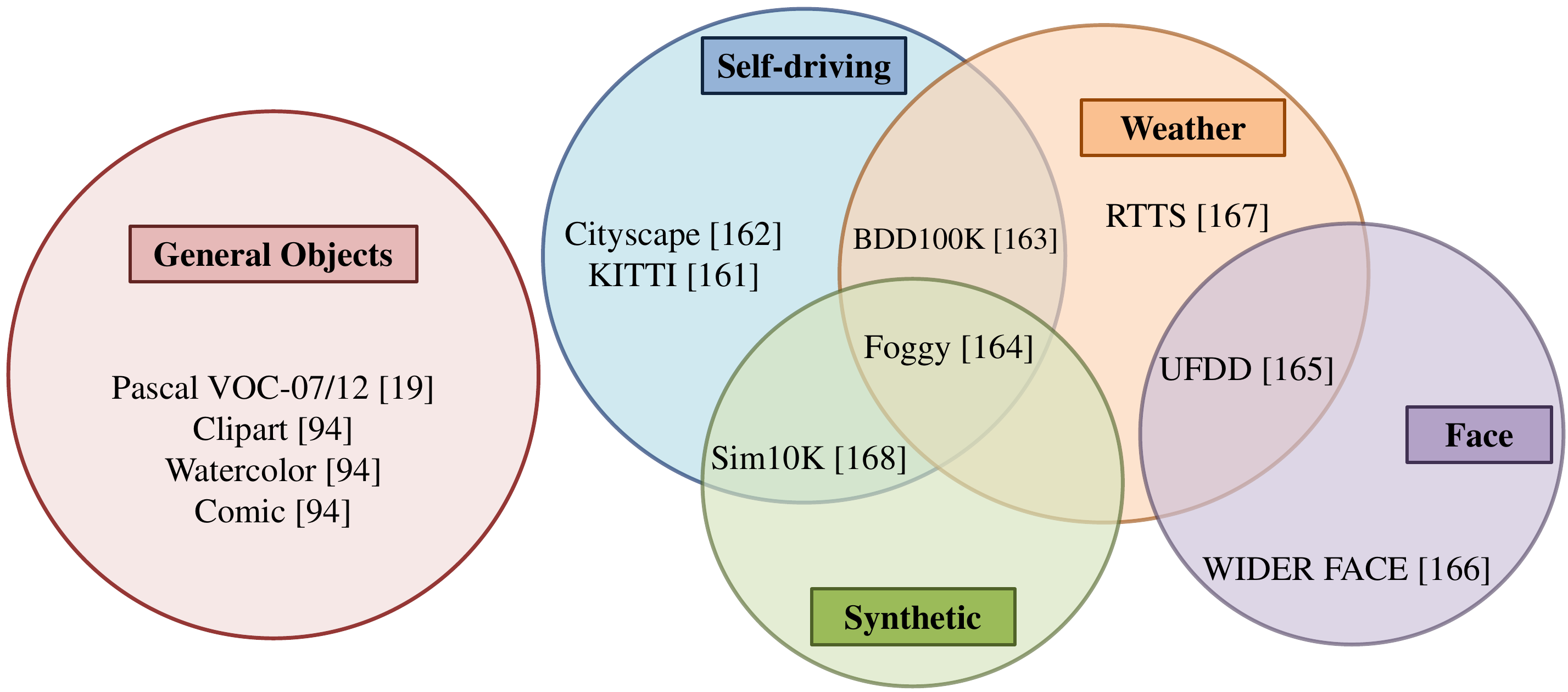}
	\end{center}
	\vskip -15.0pt
	\caption{ In unsupervised domain adaptive object detection, the benchmark datasets can be clustered into general objects, self-driving, face, weather degradation and synthetic categories. The dataset can be from purely one category, or it can be a combination of multiple categories. From the figure, we can infer that the Cityscapes dataset is purely from the self-driving category. In contrast, the FoggyCityscapes dataset is a combination of weather degradation, synthetic and self-driving.}
	\label{fig:dataset_venn}
\end{figure}

A variety of datasets have been used for evaluating the domain adaptive object detection approaches. Most of these datasets have been extensively used for evaluating object detection approaches, and hence, can naturally be used for the purpose of evaluating domain adaptive methods by simply defining different sets of data as source and target domain. These datasets can be classified into the following categories: (1) General objects \cite{everingham2010pascal}, \cite{inoue2018cross}, (2) Self-driving \cite{geiger2013vision, cordts2016cityscapes, yu2020bdd100k, Sakaridis2018SemanticFS}, (3) Face detection \cite{nada2018pushing, yang2016wider}, (4) Weather degradation \cite{li2019benchmarking, Sakaridis2018SemanticFS, nada2018pushing, yu2020bdd100k}, and (5) Synthetic datasets \cite{johnson2016driving}. Fig.~\ref{fig:dataset_venn} illustrates an overview of the different categories of datasets and Table ~\ref{tab:dataset_sum} provides summary of various datasets train and test samples. In what follows, we discuss the various datasets in detail.\\

\setlength{\tabcolsep}{15pt}
\renewcommand{\arraystretch}{1.25}
\begin{table*}[]
    \Huge
    \caption{Quantitative comparison (mAP) of existing domain adaptive object detection methods. CS: Cityscapes, F: FoggyCityscapes, S10K: Sim 10K, KI: KITTI, VOC: Pascal VOC, Clip: Clipart, WC: Watercolor, BDD: BDD100K, SS: Shorter side, LR: Learning Rate, Iter: Iteration, FRCNN: Faster-RCNN.  Red and blue color indicate best and second-best methods in respective adaptation scenario in terms of mAP. $^{\dagger}$ denotes that the corresponding method uses ResNet-50 backbone in the detection model, rest of the methods utilize VGG16 backbone.}
    \label{tab:comparison}
	\resizebox{1\linewidth}{!}{
	\begin{tabular}{|l|c|c|c|c|c|c|c|c|c|c|c|}
\hline
\textbf{Method}                                         & \textbf{CS $\rightarrow$ F} & \textbf{S10K $\rightarrow$ CS} & \textbf{KI $\rightarrow$ CS} & \textbf{CS $\rightarrow$ KI} & \textbf{VOC $\rightarrow$ Clip} & \textbf{VOC $\rightarrow$ WC} & \textbf{VOC $\rightarrow$ Comic} & \textbf{CS $\rightarrow$ BDD} & \textbf{Framework} & \textbf{LR} & \textbf{Additional Comments} \\ \hline
Faster-RCNN in the wild \cite{chen2018domain}           & 27.6                       & 38.9                         & 38.5                        & 64.1                        & -                      & -                         & -                    & -                   & FRCNN          &   0.001          &      SS: 500                        \\ \hline
Diversify and match \cite{kim2019diversify}             & 34.6                       & -                             & -                           & -                           & 41.8                   & 52.0                        & \textcolor{blue}{34.5}                 & -             & FRCNN                &   0.001           &  SS: 600                              \\ \hline
Multi-adversarial adaptation \cite{he2019multi}         & 34.0                         & 41.1                          & 41.0                          & 72.1                        & -                      & -                         & -                    & -              & FRCNN               &   0.001           &    -                          \\ \hline
Progressive adaptation \cite{hsu2020progressive}        & 36.9                       & -                             & \textcolor{blue}{43.9}                        & -                           & -                      & -                         & -                    & \textcolor{blue}{24.3}               & FRCNN           &    0.001          &   -                           \\ \hline
Strong weak distribution alignment \cite{saito2019strong}                   & 34.3                       & 40.7                          & -                           & -                           & 38.1                   & 53.3                      & -                    & -            & FRCNN                 &    0.001          &      SS: 600                        \\ \hline
%Domain contrast object detection \cite{liu2020domain}                 & -                          & -                             & -                           & -                           & 43.2                   & 53.7                      & 38.7                 & -                & FRCNN             &      0.001        &     Epochs: 7                        \\ \hline
Mean teacher with object relations \cite{cai2019exploring}                   & 35.1                       & -                             & -                           & -                           & -                      & -                         & -                    & -               & FRCNN              &     0.001        &     SS: 600  \\ \hline
Large-scale instance-level \cite{shen2021cdtd}                 & 40.3                      & -                            & -                           & -                           & -                      & -                         & -                    & -               & FRCNN              &     0.001        &     SS: 600  \\ \hline
%Scale-Aware DA \cite{chen2021scale}                 & 36.0                       & 49.0                             & 43.2                           & 40.7                           & 43.3                      & 56.0                         & -                    & -               & FRCNN              &     0.002        &     SS: 600, Epochs: 10  \\ \hline
Curriculum self-paced learning$^{\dagger}$ \cite{soviany2021curriculum}             & -                       & 47.6                     & 43.8                           & -                           & 37.8                      & -                        & -                    & -               & FRCNN              &     0.001        &     SS: 600  \\ \hline
Image-instance full alignment network \cite{zhuang2020ifan}                         & 36.2                     & 47.1                             & -                           & -                           & -                      & -                         & -                    & -                & FRCNN             &     0.001        &       SS: 600                       \\ \hline
Forward-Backward cyclic adaptation \cite{yang2020unsupervised}                     & 36.7                       & 42.7                          & -                           & -                           & 38.5                   & 53.6                      & -                    & -          & FRCNN                   &     0.001        &         SS: 600, Epochs: 10                    \\ \hline
Categorical regularization \cite{xu2020exploring}       & 37.4                       & -                             & -                           & -                           & 38.3                   & -                         & -                    & \textcolor{red}{26.9}                   & FRCNN          &    0.001         &      SS: 600                        \\ \hline
Collaborative training \cite{zhao2020collaborative}     & 35.9                       & 44.5                         & 43.6                        & -                           & -                      & -                         & -                    & -                     & FRCNN        &     0.001        &       SS: 600                       \\ \hline
%One-shot cross-domain adaptation \cite{d2020one}        & 31.9                         & -                             & 33.5                        & 75.4                         & 33.9                   & 52.0                      & 26.9                 & -              & FRCNN               &   0.001          &      SS: 600                        \\ \hline
Robust learning from noisy labels \cite{khodabandeh2019robust}        & 36.4                          & 42.5                             & 42.9                      & \textcolor{red}{77.6}                         & -                   & -                      & -                 & -              & FRCNN               &   0.001          &      SS: 600                        \\ \hline
Unbiased mean-teacher \cite{deng2021unbiased}        & 41.4                          & 43.1                             & -                       & -                        & \textcolor{blue}{42.7}                   & \textcolor{blue}{56.9}                      & -                 & -              & FRCNN               &   0.001          &      SS: 600                        \\ \hline
Prior-based DA Object Detection \cite{sindagi2020prior}                        & 39.3                       & -                             & -                           & -                           & -                      & -                         & -                    & -                & FRCNN             &     0.001        &       SS: 600                       \\ \hline
Coarse-to-fine adaptation \cite{zheng2020cross}         & 38.6                       & 43.8                          & -                           & 41.0                          & -                      & -                         & -                    & -                  & FRCNN           &      0.001       &     SS: 600                         \\ \hline
Dual Multi-Label Prediction \cite{zhao2020adaptive}                            & 38.8                       & -                             & -                           & -                           & -                      & 56.0                        & 33.5                 & -                & FRCNN             &     0.001        &         SS: 600, Epochs: 10                      \\ \hline
Instance-invariant progressive disentanglement \cite{wu2019instance}                         & 36.4                      & -                             & -                           & -                           & 42.1                      & 56.9                         & -                    & -                & FRCNN             &     0.001        &       SS: 600                       \\ \hline
Graph-induced prototype alignment$^{\dagger}$ \cite{xu2020cross}   & 39.5                      & 47.6                          & 47.9                        & -                           & -                      & -                         & -                    & -                & FRCNN          &     0.001        &  SS: 600, Epochs: 20                            \\ \hline   
Harmonizing transferability  \cite{chen2020harmonizing}   & 39.8                       & 42.5                          & -                           & -                           & 40.3                   & -                         & -                    & -                 & FRCNN            &      0.001       &      SS: 600                        \\ \hline
%Free lunch for source-free adaptation \cite{li2020free} & 33.5                       & 42.9                          & 44.6                          & -                        & -                      & -                         & -                    & 29.0                 & FRCNN           &     0.001        &        -                      \\ \hline
Asymmetric Tri-way Faster-RCNN \cite{he2020domain}                          & 38.7                       & 42.8                          & 42.1                        & 73.5                        & 42.1                   & 54.9                      & -                    & -                & FRCNN             &      0.001       &    SS: 600                         \\ \hline
Selective cross-domain alignment \cite{zhu2019adapting}                        & 33.8                       & 43.0                         & 42.5                        & -                           & -                      & -                         & -                    & -            & FRCNN                 &   0.0001           &   SS: 512, Epochs: 25                           \\ \hline
Spatial attention pyramid network \cite{li2020spatial}  & 40.9                       & 44.9                          & 43.4                        & 75.2                           & 42.2                   & 55.2                      & -                    & -                  & FRCNN           &     0.00001        &   SS: 600, Iter: 90K                           \\ \hline
Cycle-consistent adaptation \cite{zhang2019cycle}                          & 33.2                       & 41.5                          & 41.7                        & -                        & -                   & -                      & -                    & -                & FRCNN             &      0.002       &    SS: 600                         \\ \hline
Multi-level adaptation \cite{xie2019multi}                 & 36.0                       & 42.8                             & -                           & -                           & -                      & -                         & -                    & -               & FRCNN              &     0.002        &     SS: 600, Epochs: 10  \\ \hline
Multi-scale robust discrimination \cite{pan2020multi}                 & 37.0                       & 43.4                             & -                           & -                           & -                      & -                         & -                    & -               & FRCNN              &     0.002        &     SS: 600, Iter: 90K  \\ \hline
Domain invariant region proposal \cite{yang2020domain}                        & 39.2                       & 45.5                     & 44.0                           & -                           & -                      & -                         & -                    & -                & FRCNN             &     0.002        &       SS: 600, Epochs: 10                       \\ \hline
Memory-guided category-wise adaptation \cite{vs2021mega}  & \textcolor{blue}{41.8}                       & 44.8                          & 43.0                          & \textcolor{blue}{75.5}                        & -                      & -                         & -                    & -               & FRCNN              &     0.002        &    SS: 600, Epochs: 10                          \\ \hline
Conditional normalization network \cite{su2020adapting} & 36.6                       & \textcolor{blue}{49.3}                          & \textcolor{red}{44.9}                        & -                           & -                      &                           & -                    & -                  & FRCNN           &     0.00625        &    SS: 512, Epochs:12                          \\ \hline
Self-training for one-stage detector \cite{kim2019self} & -                          & -                             & -                           & -                           & 35.7                   & 49.9                      & 26.8                 & -                & SSD             &   0.001           &     SS: 300, Iter: 120K                         \\ \hline
Implicit invariant one-stage network \cite{chen20213net} & -                          & -                             & -                           & -                           & 37.8                   & 51.5                      & 30.1                 & -                 & SSD            &    0.001         &     SS: 300                       \\ \hline
%Cross-domain weakly supervised adaptation \cite{inoue2018cross}             & -                          & -                             & -                           & -                           & 34.9                   & 54.3                      & 37.2                 & -              & SSD               &    0.00001          &   SS: 300, Epochs: 20                          \\ \hline
Object detection via style consistency \cite{rodriguez2019domain}              & 34.3                          & -                             & -                           & -                           & \textcolor{red}{44.8}                   & \textcolor{red}{57.3}                      & \textcolor{red}{39.4}                 & -              & SSD               &    0.00001          &   SS: 300, Epochs: 20                          \\ \hline
Every pixel matters \cite{hsu2020every}                 & 36.0                       & 49.0                             & 43.2                           & -                           & -                      & -                         & -                    & -               & FCOS              &     0.005        &     SS: 800, Epochs: 10  \\ \hline
Adaptive transformer-based detector \cite{jiang2020blockchain}        & \textcolor{red}{43.5}                          & \textcolor{red}{55.3}                             & -                        & -                         & -                   & -                     & -                & -              & DETR               &   0.0002          &      Epochs: 50                        \\ \hline
\end{tabular}
}
\end{table*}

\setlength{\tabcolsep}{15pt}
\renewcommand{\arraystretch}{1.25}
\begin{table*}[]
    \Huge
    \caption{Quantitative comparison ($\Delta\text{mAP}$) of existing domain adaptive object detection methods. CS: Cityscapes, F: FoggyCityscapes, S10K: Sim 10K, KI: KITTI, VOC: Pascal VOC, Clip: Clipart, WC: Watercolor, BDD: BDD100K, SS: Shorter side, LR: Learning Rate, Iter: Iteration, FRCNN: Faster-RCNN. Red and blue color indicate best and second-best methods in respective adaptation scenario in terms of $\Delta\text{mAP}$. $^{\dagger}$ denotes that the corresponding method uses ResNet-50 backbone in the detection model, rest of the methods utilize VGG16 backbone.}
    \label{tab:comparison_diff}
	\resizebox{1.0\linewidth}{!}{
	\begin{tabular}{|l|c|c|c|c|c|c|c|c|c|c|c|}
\hline
\textbf{Method}                                         & \textbf{CS $\rightarrow$ F} & \textbf{S10K $\rightarrow$ CS} & \textbf{KI $\rightarrow$ CS} & \textbf{CS $\rightarrow$ KI} & \textbf{VOC $\rightarrow$ Clip} & \textbf{VOC $\rightarrow$ WC} & \textbf{VOC $\rightarrow$ Comic} & \textbf{CS $\rightarrow$ BDD} & \textbf{Framework} & \textbf{LR} & \textbf{Additional Comments} \\ \hline
Faster-RCNN in the wild \cite{chen2018domain}           & 8.80                       & 8.85                         & 8.30                        & 10.6                        & -                      & -                         & -                    & -                   & FRCNN          &   0.001          &      SS: 500                        \\ \hline
Diversify and match \cite{kim2019diversify}             & 16.7                       & -                             & -                           & -                           & \textcolor{red}{16.9}                   & 12.2                        & 13.1                 & -             & FRCNN                &   0.001           &  SS: 600                              \\ \hline
Multi-adversarial adaptation \cite{he2019multi}         & 15.2                         & 11.0                          & 10.8                          & 18.6                        & -                      & -                         & -                    & -              & FRCNN               &   0.001           &    -                          \\ \hline
Progressive adaptation \cite{hsu2020progressive}        & 17.3                       & -                             & \textcolor{red}{15.1}                        & -                           & -                      & -                         & -                    & \textcolor{blue}{3.50}               & FRCNN           &    0.001          &   -                           \\ \hline
Strong weak distribution alignment \cite{saito2019strong}                   & 14.0                       & 6.10                          & -                           & -                           & 10.3                   & 8.7                      & -                    & -            & FRCNN                 &    0.001          &      SS: 600                        \\ \hline
%Domain contrast object detection \cite{liu2020domain}                 & -                          & -                             & -                           & -                           & 15.4                   & 3.30                      & 8.90                 & -                & FRCNN             &      0.001        &     Epochs: 7                        \\ \hline
Mean teacher with object relations \cite{cai2019exploring}                   & 8.20                       & -                             & -                           & -                           & -                      & -                         & -                    & -               & FRCNN              &     0.001        &     SS: 600  \\ \hline
Large-scale instance-level \cite{shen2021cdtd}                 & 20.0                      & -                            & -                           & -                           & -                      & -                         & -                    & -               & FRCNN              &     0.001        &     SS: 600  \\ \hline
%Scale-Aware DA \cite{chen2021scale}                 & 36.0                       & 49.0                             & 43.2                           & 40.7                           & 43.3                      & 56.0                         & -                    & -               & FRCNN              &     0.002        &     SS: 600, Epochs: 10  \\ \hline
Curriculum self-paced learning$^{\dagger}$ \cite{soviany2021curriculum}              & -                       & 16.9                     & 12.3                           & -                           & 11.7                      & -                        & -                    & -               & FRCNN              &     0.001        &     SS: 600  \\ \hline
Image-instance full alignment network \cite{zhuang2020ifan}                         & 15.2                     & 12.0                             & -                           & -                           & -                      & -                         & -                    & -                & FRCNN             &     0.001        &       SS: 600                       \\ \hline
Forward-Backward cyclic adaptation \cite{yang2020unsupervised}                     & 17.9                       & 11.5                          & -                           & -                           & 10.7                  & 9.00                     & -                    & -          & FRCNN                   &     0.001        &         SS: 600, Epochs: 10                    \\ \hline
Categorical regularization \cite{xu2020exploring}       & 15.4                       & -                             & -                           & -                           & 11.3                  & -                         & -                    & \textcolor{blue}{3.50}                   & FRCNN          &    0.001         &      SS: 600                        \\ \hline
Collaborative training \cite{zhao2020collaborative}     & 9.70                       & 10.0                        & 8.7                        & -                           & -                      & -                         & -                    & -                     & FRCNN        &     0.001        &       SS: 600                       \\ \hline
%One-shot cross-domain adaptation \cite{d2020one}        & 5.10                          & -                             & 7.00                       & 0.30                         & 7.50                   & 5.20                      & 8.80                 & -              & FRCNN               &   0.001          &      SS: 600                        \\ \hline
Robust learning from noisy labels \cite{khodabandeh2019robust}        & 4.5                          & 11.5                             & 11.8                      & 21.4                         & -                   & -                      & -                 & -              & FRCNN               &   0.001          &      SS: 600                        \\ \hline
Unbiased mean-teacher \cite{deng2021unbiased}        & \textcolor{red}{19.6}                          & 8.80                             & -                       & -                        & 13.6                   & 8.00                      & -                 & -              & FRCNN               &   0.001          &      SS: 600                        \\ \hline
Prior-based DA Object Detection \cite{sindagi2020prior}                        & 14.9                       & -                             & -                           & -                           & -                      & -                         & -                    & -                & FRCNN             &     0.001        &       SS: 600                       \\ \hline
Coarse-to-fine adaptation \cite{zheng2020cross}         & 17.8                       & 8.80                          & -                           & 7.60                          & -                      & -                         & -                    & -                  & FRCNN           &      0.001       &     SS: 600                         \\ \hline
Dual Multi-Label Prediction \cite{zhao2020adaptive}                            & 15.4                       & -                             & -                           & -                           & -                      & \textcolor{blue}{11.4}                        & \textcolor{blue}{13.8}                 & -                & FRCNN             &     0.001        &         SS: 600, Epochs: 10                      \\ \hline
Instance-invariant progressive disentanglement \cite{wu2019instance}                         & 13.7                      & -                             & -                           & -                           & 14.3                      & \textcolor{red}{12.3}                         & -                    & -                & FRCNN             &     0.001        &       SS: 600                       \\ \hline
Graph-induced prototype alignment$^{\dagger}$ \cite{xu2020cross}  & 12.6                      & 13.0                          & 10.3                       & -                           & -                      & -                         & -                    & -                & FRCNN          &     0.001        &  SS: 600, Epochs: 20                            \\ \hline   
Harmonizing transferability  \cite{chen2020harmonizing}   & \textcolor{blue}{19.5}                      & 7.90                          & -                           & -                           & 12.5                   & -                         & -                    & -                 & FRCNN            &      0.001       &      SS: 600                        \\ \hline
%Free lunch for source-free adaptation \cite{li2020free} & 14.0                       & 9.20                          & 8.20                          & -                        & -                      & -                         & -                    & 7.20                 & FRCNN           &     0.001        &        -                      \\ \hline
Asymmetric Tri-way Faster-RCNN \cite{he2020domain}                          & 18.4                       & 8.20                          & 11.9                        & 20.0                        & 14.3                   & 10.3                      & -                    & -                & FRCNN             &      0.001       &    SS: 600                         \\ \hline
Selective cross-domain alignment \cite{zhu2019adapting}                        & 7.60                       & 9.10                         & 5.10                        & -                           & -                      & -                         & -                    & -            & FRCNN                 &   0.0001           &   SS: 512, Epochs: 25                           \\ \hline
Spatial attention pyramid network \cite{li2020spatial}  & 17.6                       & 10.3                          & \textcolor{blue}{13.2}                        & \textcolor{blue}{21.7}                           & 14.4                   & 10.6                      & -                    & -                  & FRCNN           &     0.00001        &   SS: 600, Iter: 90K                           \\ \hline
Cycle-consistent adaptation \cite{zhang2019cycle}                          & 8.78                       & 6.70                          &  4.00                       & -                        & -                  & -                      & -                    & -                & FRCNN             &      0.002       &    SS: 600                         \\ \hline
Multi-level adaptation \cite{xie2019multi}                 & 13.2                       & 8.50                             & -                           & -                           & -                      & -                         & -                    & -               & FRCNN              &     0.002        &     SS: 600, Epochs: 10  \\ \hline
Multi-scale robust discrimination \cite{pan2020multi}                 & 17.1                       & 9.90                             & -                           & -                           & -                      & -                         & -                    & -               & FRCNN              &     0.002        &     SS: 600, Iter: 90K  \\ \hline
Domain invariant region proposal \cite{yang2020domain}                        & 16.6                       & 11.3                     & 9.30                          & -                           & -                      & -                         & -                    & -                & FRCNN             &     0.002        &       SS: 600, Epochs: 10                       \\ \hline
Memory-guided category-wise adaptation \cite{vs2021mega}  & 17.4                       & 10.5                          & 12.8                          & \textcolor{red}{22.0}                        & -                      & -                         & -                    & -               & FRCNN              &     0.002        &    SS: 600, Epochs: 10                          \\ \hline
Conditional normalization network \cite{su2020adapting} & 10.5                       & \textcolor{blue}{15.0}                          & 7.80                        & -                           & -                      &   -                        & -                    & -                  & FRCNN           &     0.00625        &    SS: 512, Epochs:12                          \\ \hline
Self-training for one-stage detector \cite{kim2019self} & -                          & -                             & -                           & -                           & 9.00                   & 2.80                      & 4.90                 & -                & SSD             &   0.001           &     SS: 300, Iter: 120K                         \\ \hline
Implicit invariant one-stage network \cite{chen20213net} & -                          & -                             & -                           & -                           & 11.1                   & 4.4                      & 8.20                 & -                 & SSD            &    0.001         &     SS: 300                       \\ \hline
%Cross-domain weakly supervised adaptation \cite{inoue2018cross}             & -                          & -                             & -                           & -                           & 19.2                   & 4.70                      & 12.3                 & -              & SSD               &    0.00001          &   SS: 300, Epochs: 20                          \\ \hline
Object detection via style consistency \cite{rodriguez2019domain}              & 8.70                          & -                             & -                           & -                           & \textcolor{blue}{14.5}                   & 7.70                      & \textcolor{red}{14.5}                 & -              & SSD               &    0.00001          &   SS: 300, Epochs: 20                          \\ \hline
Every pixel matters \cite{hsu2020every}                 & 17.2                       & \textcolor{red}{18.9}                             & 13.0                           & -                           & -                      & -                         & -                    & -               & FCOS              &     0.005        &     SS: 800, Epochs: 10  \\ \hline
Adaptive transformer-based detector \cite{jiang2020blockchain}        & 9.5                          & 4.80                             & -                        & -                         & -                   & -                     & -                & -              & DETR               &   0.0002          &      Epochs: 50                        \\ \hline
\end{tabular}
}
\end{table*}

\noindent \textbf{Cityscapes.} Scene understanding of complex urban streets is an important problem statement for a wide range of applications.  To address this issue, the Cityscapes dataset was released in 2016 by a group of researchers in Daimler and TU Dresden \cite{cordts2016cityscapes}.  Cityscapes has 8 categories: car, truck, motorcycle/bike, train, bus, rider, and person. The dataset is collected in 50 cities and covers a variety of seasons (spring, summer, fall). The Cityscapes dataset contains 2975 images for training and 500 images for testing.\\

\noindent \textbf{FoggyCityscapes.} Analyzing scenes of urban streets under adverse weather is a challenging problem. To consider such scenarios, the FoggyCityscapes was introduced  \cite{Sakaridis2018SemanticFS}, by applying a fog filter over the Cityscapes dataset. The FoggyCityscapes has the same 8 categories of Cityscapes dataset. The FoggyCityscapes dataset contains 2975 images for training and 500 images for testing.\\

\noindent \textbf{Sim10K.} Considering that the collection of real-world data is time-consuming and tedious, advancements in computer graphics have been exploited to generate photo-realistic data which can be easily rendered and annotated. This alternative to real-world data collection is inexpensive and efficient.  In 2017, Sim10K \cite{johnson2016driving}, a synthetic dataset rendered by the gaming engine Grand Theft Auto was released. It has 10K images and only a car category having 58,701 car instances. The dataset provides annotations in the Pascal VOC format for ease of use.\\

\noindent \textbf{KITTI.} KITTI \cite{geiger2013vision} is one of the most popular datasets for self-driving and mobile robotics. The dataset contains hours of videos of traffic scenarios recorded with various high-quality sensors like RGB, grayscale and depth sensors. The KITTI object detection dataset consists of 7,481 training images and 7,518 test images, comprising of 80,256 labeled objects which span over 8 categories: Car, Van, Truck, Pedestrian, Person sitting, Cyclist, Tram, Misc.\\

\noindent \textbf{Pascal VOC2007/2012.}  PASCAL VOC \cite{everingham2010pascal} is a real-world dataset containing general objects. It was introduced for a variety of tasks such as large-scale classification, object detection, and segmentation. The PASCAL VOC image set has been updated multiple times between 2005 to 2012. Among these multiple versions, the most popular ones are 2007 and 2012 image sets with training and validation split containing a total of 15k images. The images in the dataset contain over 20 general object categories: airplane, bicycle, bird, boat, bottle, bus, car, cat, chair, cow, table, dog, horse, bike, person, plant, sheep, sofa, train, tv.\\

\noindent \textbf{Clipart, Watercolor, Comic.}  Clipart \cite{inoue2018cross}, Watercolor \cite{inoue2018cross} and Comic \cite{inoue2018cross} datasets contain abstract, artistic and comical images, respectively. Understanding these abstract-type images will allow for the direct study of how a model infer high-level semantic information. Consequently, these datasets are extensively used in dissimilar domain adaptation problems. The Clipart contains a total of 1K images with 20 categories same as Pascal VOC. The watercolor and comic dataset contain 6 categories: bike, bird, car, cat, dog, person. Moreover, both datasets contain 1K training images and 1K testing images, respectively.\\

\noindent \textbf{BDD100K.} The BDD100K dataset \cite{yu2020bdd100k} is an autonomous driving dataset contains large diversity in terms of scenes, geographical regions. It is the largest (till-date) driving video dataset with 100K videos containing scenes from New York, Berkeley, San Francisco and Bay Area. The dataset contains 70k training images and 10k validation images. The authors ensure that the images encompass different types of weather, six different scenes, three separate times of the day, and 10 object categories and bounding box annotations. For the domain adaptive detection task specifically, a subset of the BDD100k dataset that contains images labeled as daytime is used. This subset contains 36,728 training and 5,258 validation images.\\

\noindent \textbf{WIDER FACE.} Face detection is an important application in modern computer vision. However, due to lack of variance in the face detection dataset, in 2015, Shuo \cite{yang2016wider} released the WIDER FACE, which is the largest benchmark dataset consisting of 32,000 images and 199K labeled faces. The authors ensure high degree of variability in scale, expression, illumination, pose, makeup and occlusion.\\

\noindent \textbf{UFDD.} Face detection under adverse conditions is very important, especially for video surveillance.  Considering this, the UFDD dataset \cite{nada2018pushing} consisting of images collected under a variety of weather conditions was released. The authors attempt to specifically capture the following conditions: rain, snow, haze, lens impediments, blur, illumination. The dataset contains a total of 6424 images.\\

\noindent \textbf{RTTS.} REalistic Single Image DEhazing (RESIDE) \cite{li2019benchmarking} is a large-scale dataset consisting of both real and synthetic hazy images. RTTS is an object detection dataset and is a subset of the RESIDE. It contains 4,807 un-annotated and 4,322 annotated real-world hazy images covering mostly traffic and driving scenarios. RTTS has total 5 categories, namely motorcycle/bike, person, bicycle, bus and car.\\

\setlength{\tabcolsep}{5pt}
\renewcommand{\arraystretch}{1.15}
\begin{table}[!t]
	\caption{Summary of various datasets used in domain adaptive object detection experiments. All the datasets except RTTS contain labeled training images and test images. The RTTS dataset contains unlabeled training images and labeled test images.}
	\label{tab:dataset_sum}
	\centering
	\begin{tabular}{|l|c|c|c|c|}
		\hline
		\multirow{2}{*}{\textbf{Dataset}} & \multicolumn{2}{c|}{\textbf{Train}} & \multicolumn{2}{c|}{\textbf{Test}} \\ \cline{2-5} 
		& \textit{Images}     & \textit{Catagories}    & \textit{Images}    & \textit{Catagories}    \\ \hline
		Cityscapes               & 2975       & 8             & 500       & 8             \\ \hline
		VOC-2007                 & 5011       & 20            & 5011      & 20            \\ \hline
		VOC-2012                 & 11540      & 20            & 11540     & 20            \\ \hline
		SIM10K                   & 10000      & 1             & 10000     & 1             \\ \hline
		KITTI                    & 7481       & 1             & 7481      & 1             \\ \hline
		FoggyCityscapes          & 2975       & 8             & 500       & 8             \\ \hline
		Clipart                  & 1000       & 20            & 1000      & 20            \\ \hline
		Watercolor               & 1000       & 6             & 1000      & 6             \\ \hline
		BDD100K                  & 36728      & 10            & 5258      & 10            \\ \hline
		Comic                    & 1000       & 6             & 1000      & 6             \\ \hline
		WIDER FACE               & 32000      & 1             & 32000     & 1             \\ \hline
		UFDD                     & 442        & 1             & 442       & 1             \\ \hline
		RTTS                     & 4807       & 5             & 4322      & 5             \\ \hline		
	\end{tabular}
\end{table}

\subsection{Adaptation scenarios}

In this section, we describe the various adaptation scenarios and protocols followed by the existing approaches. 

\subsubsection{Adverse weather conditions}
Stable detection performance in different weather conditions is important for safety-critical applications like self-driving cars. Weather conditions introduce image artifacts which can negatively impact the detection performance. FoggyCityscapes and Cityscapes can be utilized as target and source domains to evaluate the effectiveness of adaptation methods in adverse weather. Moreover, under the real hazy condition, one can extend the domain adaptation setting for Cityscapes to the RTTS dataset, used as the source and target domain, respectively. Similarly, for face detection, such weather conditions prove challenging when the face detector is trained with clean weather data. This adaptation scenario can be explored with the source domain as WIDER-Face and target domain as UFDD-Haze. 
% The face detection task is closely related to the task of object detection. 

\subsubsection{Synthetic data adaptation}
Synthetic data offers an inexpensive alternative to real data collection as it is easier to collect, and with appropriate engineering, annotations can be made readily available for synthetic data with very little labor cost. In spite of the advancements in computer graphics, photo-realistic synthetic data generated using state-of-the-art rendering engines suffer from subtle image artifacts, which can result in sub-optimal performance on real-world data. The adaptation from Sim10K (source domain) to Cityscapes (target domain) is used in the literature to analyze this setting.
%the synthetic data can be auto-annotated

\subsubsection{Cross-camera adaptation}
Differences in the intrinsic and extrinsic camera properties like resolution, distortion, orientation, location result in images that capture the objects differently from each other in terms of quality, scale, and viewing angle. While the collected data can be real, these domain differences will potentially result in severe performance degradation. To evaluate the domain adaptation methods under cross-camera adaptation settings, most literature considers KITTI to Cityscapes and KITTI to Cityscapes.

\subsubsection{Adaptation between dissimilar domains}
In the shift from real-world to artistic images, underlying features, the texture of real-world images completely differ in artistic images. Hence, a lot of methods consider adaptation of dissimilar domains with three adaptation settings: PASCAL-VOC to Clipart, PASCAL-VOC to Watercolor and PASCAL-VOC to Comic. All these domain shifts are from real-world images to synthetic as well as artistic images.

%\subsubsection{Cross-city adaptation}
\subsubsection{Adaptation to Large-scale Dataset} 
The increasing availability of cheap and good quality cameras has made collecting large-scale datasets much easier. However, annotating them for the detection task is labor-intensive. Therefore, exploring the possibility of adopting from a smaller dataset to a larger dataset has significant real-world implications. To evaluate this adaptation setting, methods in most domain adaptation literature consider Cityscapes to BDD100k. In the BDD100K dataset, only the daylight subset is considered as the target domain and the Cityscapes dataset as the source domain.

%To examine the adaptation from a relatively smaller dataset to a large unlabeled domain-containing distinct attributes, one can harvest more from existing resources and adapt them too complicated environments. To this end, the Cityscapes and BDD100k datasets are used as the source and target domains, respectively. A subset of the BDD100k dataset annotated as daytime is chosen to be our target domain and consider the city scene as the adaptation factor since daytime data only exists in the Cityscapes dataset.

\subsection{Evaluation Metric}
Typically, object detectors are evaluated using the average precision (AP) that was introduced in VOC2007 \cite{everingham2010pascal}. The average precision is computed for each category by calculating the area under the precision-recall curve. Precision and recall are defined as follows:
\begin{equation}
\begin{aligned}
\text{Precision} &= \frac{\text{True Positive}}{\text{True Positive} + \text{False Positive}}\\
\text{Recall}    &= \frac{\text{True Positive}}{\text{True Positive} + \text{False Negative}}.
\end{aligned}
\end{equation}

\begin{figure}[h!]
    \begin{center}
		\includegraphics[width=0.50\linewidth]{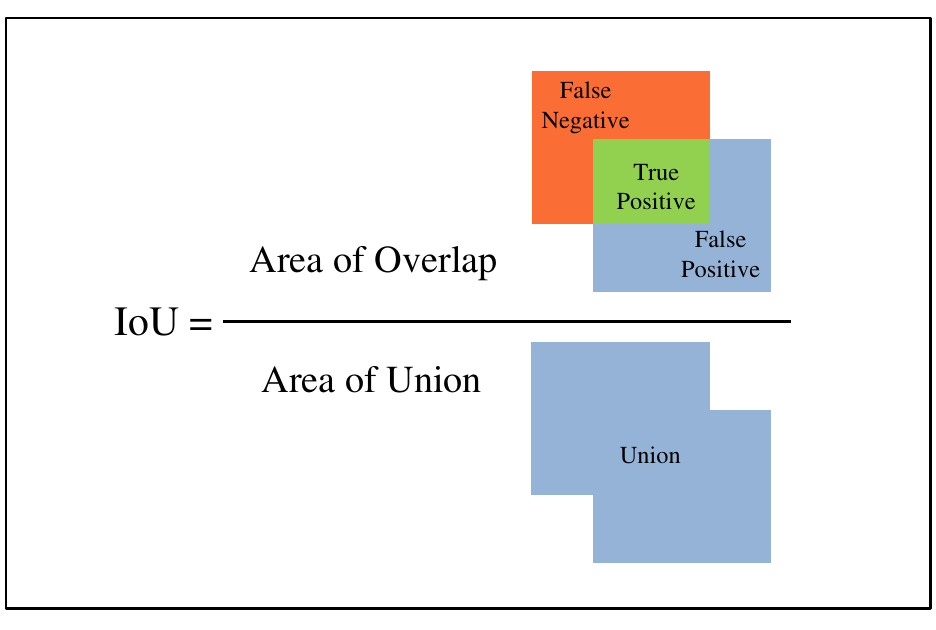}
	\end{center}
	\vskip -20.0pt
	\caption{Illustration of intersection over union (IoU).}
    \label{fig:iou}
\end{figure}

Precision is a measure of how accurate are the predictions of an object detector. Recall identifies how many of positive predictions of the object detector are correctly predicted. In order to determine if the prediction of a detector matches with the ground truth bounding box, the intersection over union (IoU) measure is used (see Fig.~\ref{fig:iou} for details). This measure is defined as follows: 
% if it is greater than a predefined threshold, the corresponding prediction is a true positive. If it is lesser than a predefined threshold, the corresponding prediction is a false positive.
\begin{equation}
\begin{aligned}
\operatorname{IoU}     &=    \frac{\text{Area of overlap}}{\text{Area of union}}. \\
\end{aligned}
\end{equation}
\noindent The IoU measure is then used to compute if a particular prediction is a true positives or false positive given a pre-defined threshold, using the following equation:
\begin{equation}
\begin{aligned}
TP_{ij} &= 1, \text{if}\operatorname{IoU}(t)_{ij}>\tau\\
FP_{ij} &= 1, \text{if}\operatorname{IoU}(t)_{ij}<\tau.\\
\end{aligned}
\end{equation}
Here, $i$ is the index of the $i$-th image while $j$ is the index of the $j$-th object and $\tau$ is the threshold value. Hence, by applying a threshold over IoU, one can measure the precision and recall by identifying all the object proposals as a true positive or false positive. Furthermore, to compare the overall performance among all categories, the mean Average Precision (mAP) metric is used, which involves calculating the average of all categorywise AP.
\begin{equation}
\text{mAP} = \frac{\sum_{i=1}^{k} \text{AP}_k}{k},
\end{equation}
where $k$ is the total number of category.
% In domain adaptation problem setting, the 0.5-IoU dependent mAP has become the standard for all domain adaptive object detection problems. 

%Moreover, mean Average Precision (mAP) is used to compare the overall performance among all categories by calculating the average category-wise AP.

% \section{Comprehensive performance results}
% \begin{table}[!htb]
% 	\centering
% 	\caption{Table for providing comprehensive comparison between major methods on all datasets with indicating hyper-parameters}
% 	\label{tab:exp}
% 	\begin{tabular}{|r|c|l|} \hline
% 		right & center & left  \\ \hline
% 		right & center & left \\ \hline
% 	\end{tabular}
% \end{table}.

\subsection{Discussion of results}

In this section, we provide detailed comparison and discussion of the performance of various domain adaptive object detection methods. For comparison, we only consider adaptation scenarios used by two or more works in the literature.  Most of the approaches conduct adaptation experiments that consist of various scenarios created by carefully selecting a source and target domain from the datasets discussed in the previous section. To indicate an adaptation scenario, we follow dataset1$\rightarrow$dataset2 format, where dataset1 is used as a labeled source domain and dataset2 is used as an unlabeled target domain.

The performance comparison reported in Table~\ref{tab:comparison} uses the absolute mAP performance as reported in the respective papers. It is important to note that this comparison is not necessarily fair for the following reasons:  Each method uses different version of the baseline detector. This is true even for the methods that use Faster-RCNN as the baseline detector. These methods use different hyperparameters for training the baseline. For example, Chen \etal \cite{chen2018domain} use Faster-RCNN with ROI pooling and images with a shorter side of 500 pixels, whereas Cai \etal \cite{cai2019exploring} uses Faster-RCNN with ROI-Align and images with a shorter side of 600 pixels. Additionally, there are several other differences like the number of epochs, learning rate, etc across different approaches. Hence, to provide a fair, comprehensive comparison, we consider relative mAP improvement, denoted as $\Delta\text{mAP}$, which is the difference between the performance of a particular approach and that of the corresponding source-only baseline as reported in the respective paper. Though $\Delta\text{mAP}$ would still not be a perfect comparison metric, we argue that it is better than comparing absolute mAP of methods that are trained with different hyperparameters. Since all methods in the literature train both source-only baseline and final method with same hyperparameters, $\Delta\text{mAP}$ would obviate most of the influences on the performance caused by hyperparameters of the respective methods. The $\Delta\text{mAP}$ metric for all methods is reported in Table~\ref{tab:comparison_diff}.

Comparing these two tables provide some interesting insights into the performance trends over the years for any particular adaptation scenario. For example, consider the case of Cityscapes$\rightarrow$BDD100K, where, from Table.~\ref{tab:comparison}, it seems that the forward-backward cyclic adaptation \cite{yang2020unsupervised}  approach improves over the progressive adaptation \cite{hsu2020progressive} strategy. However, considering the performance in terms of $\Delta\text{mAP}$, both these methods obtain similar performance improvements. This suggests that the Cityscapes$\rightarrow$BDD100K scenario needs to be explored further by future works. The case of adverse weather adaptation case (Cityscapes$\rightarrow$FoggyCityscapes), and has observed $\sim$20 mAP improvement since the first work \cite{chen2018domain}. From the absolute mAP comparison in Table.~\ref{tab:comparison}, the transformer-based adaptation \cite{zhang2021detr} and the memory-guided adaptation \cite{vs2021mega} provide the best and second-best performance. Whereas from the $\Delta\text{mAP}$ comparison, it can be observed that the unbiased mean-teacher \cite{deng2021unbiased} and harmonizing transferability \cite{chen2020harmonizing} obtains the most improvement over their respective source-only baseline. A similar trend is observed in other adaptation scenarios as well, e.g., based on absolute mAP for Sim10K$\rightarrow$Cityscapes, the transformer-based approach \cite{zhang2021detr} performs best. In contrast, the approach based on every-pixel matters \cite{hsu2020every} performs best when $\Delta\text{mAP}$ is considered. Note that for Sim10K$\rightarrow$Cityscapes, conditional normalization \cite{su2020adapting} has second-best performance for both absolute mAP and $\Delta\text{mAP}$. Similar results can be observed for the PascalVOC$\rightarrow$Comic adaptation scenario where style-consistency \cite{rodriguez2019domain} is the best performing method across the absolute mAP and the  $\Delta\text{mAP}$ measures.  

\subsubsection{Is one category of adaptation methods better than the other category of approaches?}

After discussing all the class of methods (i.e. adversarial feature learning, pseudo-label based self-training etc) and comparing their performances, it is natural to wonder if one class of methods is better than the rest.

From Table.~\ref{tab:comparison} and Table.~\ref{tab:comparison_diff}, we can see that there is no clear winner among different class of methods. For example, robust learning \cite{khodabandeh2019robust} is a pseudo-label based self-training approach that outperforms adversarial feature learning-based approach adaptive faster-rcnn \cite{chen2018domain} on many adaptation experiments. However, many adversarial feature learning-based methods \cite{vs2021mega, hsu2020every, sindagi2020prior, su2020adapting} outperform robust learning \cite{khodabandeh2019robust} on a variety of adaptation scenarios. Each class of adaptation strategy has its benefit and drawbacks. For example, adversarial feature learning (discussed in Sec.~\ref{subsec:adv_feat_learning} based training is known to be unstable in practice and requires careful parameter tuning and regularization to stabilize the feature adaptation process. However, under appropriate hyperparameters and with correct regularization, adversarial feature learning can provide significant performance improvements. Most likely for this reason, the most popular strategy used in the literature is adversarial feature learning-based methods \cite{chen2018domain, saito2019strong, xu2020exploring, hsu2020every, vs2021mega}. On the contrary, self-training with pseudo-labels (discussed in Sec.~\ref{subsec:self_training_pseudo_labels} has a stable training curve as supervised detection loss can be used to train the network with supervision from pseudo-labels obtained through the source-trained model. However, self-training based methods need to be careful of the noise in the pseudo-labels as such noise can potentially get re-enforced into the model resulting in sub-optimal performance. For this reason, most self-training based approaches utilize only confident predictions \cite{khodabandeh2019robust, kim2019self}.

Image-to-image translation based methods (discussed in Sec.~\ref{subsec:image_to_image_translation} try to avoid such training related issues by addressing the domain gap in the input-space. Typically, a image-translation module is added that transforms source samples to target-like samples and vice versa. Since target-like samples will have supervision readily available, the model can be trained better to perform detection on original samples from the target domain. However, this strategy heavily relies on the efficacy of the image-translation module to learn a perfect mapping between source and target domain. For this reason, most image-to-image translation methods additionally utilize adversarial feature learning and/or self-training strategy that can compensate some of the mistakes made by the image-translation module \cite{zhang2019cycle, hsu2020progressive, chen2020harmonizing}. 

Domain randomization (discussed in Sec.~\ref{subsec:domain_randomization}) based approaches on the other hand, create multiple stylized version of the source domain in order to ensure that the detection model is devoid of style-bias from source domain and focus on other important discriminative features to detect the objects. However, to completely de-bias the detector from any specific domain, many random domains need to be included during training, which might not be feasible. To address this, domain randomization methods also additionally apply adversarial feature learning between source, stylized-source, and target domain. 

Mean-teacher training (discussed in \ref{subsec:student_teacher_training}) is another popular strategy used to adapt detectors to a  target domain through mean-teacher updates \cite{tarvainen2017mean}. Mean-teacher \cite{tarvainen2017mean} has been successfully used for transfer learning and was initially proposed to utilize unlabeled data of similar domains. These approaches utilize additional regularization or feature matching strategy to extend the mean-teacher for unsupervised domain adaptation \cite{cai2019exploring, deng2021unbiased}. 

Similarly, graph-reasoning (discussed in \ref{subsec:graph_reasoning}) based methods aim to understand object relations within any given sample of the source domain and try to leverage that information to make sure the target domain also follow similar relations. This helps the knowledge transfer between source and target domain by enforcing certain constraints on the target features. Most of these different categories of approaches are mutually exclusive and can be combined in different ways to leverage the benefits of each strategy while mitigating the drawbacks of other strategies. For example, unbiased mean-teacher \cite{deng2021unbiased} combines mean-teacher training, adversarial feature learning and image-to-image translation and demonstrated that it could outperform most other works in the literature. 

\section{Research directions}

As discussed in the earlier sections, various methods have been proposed to address the challenging problem of adapting deep object detectors to different scenarios. In this section, we explore outstanding issues along with potential solutions and future research directions. In the following subsections, we discuss two broad categories of issues and directions: (i) Comprehensive evaluation and (ii) Improving generalization with real-world constraints.

\subsection{Comprehensive evaluation}

We identify some of the issues concerning the evaluation protocol. We believe that addressing these issues would result in a more robust and rigorous evaluation process.
\begin{itemize}
\item \textbf{Generalizing to other detection frameworks}: Most of the existing domain adaptation approaches for object detection are evaluated only on the Faster-RCNN detection framework \cite{ren2015faster}. Further, some of these approaches, such as \cite{chen2018domain} are designed with the assumption of a two-stage detection process. However, considering the popularity of single shot detection approaches such as YOLO \cite{redmon2016you}, SSD \cite{liu2016ssd}, and more recent ones such as FCOS \cite{tian2019fcos} and DETR \cite{carion2020end}, it is important to evaluate the performance of the adaptation methods using other frameworks as well. Evaluating on this additional category of detection approaches will verify if the adaptation approaches are generalizable to different detectors.\\
\item \textbf{Complex real-world datasets}: Initial approaches such as DA-Faster \cite{chen2018domain}, SWDA \cite{saito2019strong} for domain adaptive object detection designed the evaluation protocol based on existing datasets such as FoggyCityscapes, Sim10K, Pascal VOC, \textit{etc} and subsequent approaches followed similar protocols. It is important to note that some of these datasets and protocols were not specifically constructed to reflect the real-world distribution gap. For example, consider the case of building a self-driving system where it is important to construct source and target datasets such that the distribution gap arises from a variety of factors such as differences in weather conditions, geographical locations, types of vehicles, types of backgrounds, densities of vehicles, \textit{etc}. Considering this, it is crucial to construct complex datasets reflecting real-world scenarios in order to enable a more rigorous and robust evaluation process.\\
\item \textbf{Considering other applications}: Existing methods in the literature mainly focus on applications such as autonomous navigation \cite{chen2018domain, kim2019diversify, vs2021mega, hsu2020every}, crowd surveillance \cite{sindagi2020prior}, and commonly found general objects \cite{xu2020exploring, saito2019strong, he2020domain}. Whereas multiple other real world applications are rarely explored in the context of domain adaptive detection, such as medical imaging \cite{yang2020unsupervised}, scene text \cite{chen2019cross}, document objects \cite{li2020cross}. For example, consider a lesion detection model trained on Optical Coherence Tomography (OCT) images captured from a particular device. As OCT image distribution shifts when the respective image is captured from another device. It becomes crucial to perform an unsupervised adaptation of the lesion detection model to improve cross-device performance in such cases. Hence, it becomes important to develop adaptation strategies for these rarely explored applications, as they have significant real-world implications.\\
\item \textbf{Consistent training and inference strategy}: Although most of the domain adaptation methods for detection demonstrated their effectiveness on the Faster-RCNN detection framework, we observed that the training and inference strategies are not uniform. As it can be observed from Table~\ref{tab:comparison}, hyperparameters like the learning rate and training duration varies from method to method. While one may argue that this results from hyperparameter tuning, it is important to note that these hyperparameters are more associated with the backbone network, and changes in them can potentially be the cause of the performance changes. Hence, it is important to maintain consistent hyperparameters related to the backbone network in order to perform a fair comparison of the methods. Similarly, different methods use a different resizing ratio of the input images, which directly affects the detection performance. Considering these observations, it is important to establish consistent and uniform training/inference strategies that will enable the reader to obtain a fair perspective of the evaluation results.\\
\end{itemize}

\subsection{Improving generalization with real-world constraints}

In this section, we focus on potential future research directions that consider some real-world settings typically ignored in the existing approaches. Most of the existing methods are based on assumptions that need not necessarily be true in the real-world. For example, it is often assumed that (i) the number of samples across classes in both source and target domain dataset is assumed to be balanced, (ii) all the classes existing in the source domain dataset are present in the target domain dataset and vice versa, (iii) source domain dataset is always available during target domain adaptation, (iv) all target domain samples are unlabeled, and (v) large number of unlabeled target domain samples are available at the training. We discuss various problem settings where such assumptions are relaxed, the resulting challenges and possible strategies to overcome them. Furthermore, we also discuss additional strategies such as multi-source domain adaptation, test-time adaptation, and continuous adaptation necessary for real-world practice settings.
\begin{itemize}
\item \textbf{Weak/semi-supervised domain adaptation}: Unsupervised domain adaptation is specifically useful because annotating the target domain samples is labor-intensive and costly for the object detection task. However, in some scenarios, it might be possible to obtain bounding-box annotations for a subset of target samples or provide weak image-level annotations indicating the presence/absence of the categories. In such cases, this additional knowledge can be leveraged to improve the performance on target domain further. For weakly supervised domain adaptation, each image is annotated at image-level to indicate which categories are present/absence in the image and no bounding box annotations are provided. Inoue \etal \cite{inoue2018cross} is the only work in the literature addressing this issue with the help of image-to-image translation and pseudo-label training guided by weak annotations. A subset of the target dataset is labeled fully with bounding-box and respective category annotation for semi-supervised adaptation. Liu \etal \cite{liuunbiased} proposed the use of annotated subset and a relatively large unlabeled subset to enhance the detection performance. Though Liu \etal \cite{liuunbiased} does not extensively evaluate the adaptive detection benchmarks, it provides a potentially useful strategy that can be easily adopted for the domain adaptation case. Hence, the use of both weak-supervision and semi-supervision is important to explore further to bridge the performance gap between fully supervised and adaptive training.\\
\item \textbf{One/few-shot domain adaptation}: Conventional unsupervised domain adaptation setting assumes availability of a large number of unlabeled samples from the target domain. However, this might not hold true in many realistic conditions where data is an incoming stream of images \cite{d2020one} resulting in either one or few unlabeled samples available from the target domain during adaptation training. To address this sample scarcity in the target domain-specific adaptation strategy is needed to get the best out of the available dataset. D'Innocente \etal \cite{d2020one} is the only approach to explore the one-shot adaptation problem by adding an auxiliary self-supervision loss and few-shot adaptation is not yet explored in the domain adaptive detection literature. A straightforward strategy to address one/few-shot adaptation for object detectors would be to apply image stylizing to increase the pool of unlabeled target domain samples and apply a conventional feature-matching strategy by making detector domain invariant \cite{luo2020adversarial}. Another strategy would be to label the one or few images available in the target domain and perform a supervised domain adaptation with annotated sample-rich source domain and sample-scarce target domain \cite{motiian2017few, zhang2019few, zhang2019few}.\\
\item \textbf{Imbalanced classes}: Object frequency in the real world often follows a power law, and this imbalance class problem is typically ignored by existing domain adaptive detection methods which assume aligned class spaces. This assumption limits the methods' performances on imbalanced tasks encountered frequently in the real world, especially in the object detection task. Hence, it is important to avoid the dominance of any particular class while performing the domain alignment. An obvious solution to this problem is to re-weight the classes based on the frequency \cite{zou2018unsupervised, jamal2020rethinking}. However, this strategy is infeasible in the case of the target dataset where samples are unlabeled. Hence, it is important to develop methods that can estimate the class distribution in the target dataset, which can then be used for re-weighting the classes.\\
\item \textbf{Partial domain adaptation}: A related problem to the class imbalance issue is that of partial domain adaptation. A typical assumption in the domain adaptation literature is that the label space between the source and target dataset is same. That is, the source and target dataset have the same $K$ classes. However, a real-world setting might not completely reflect this scenario. It might often be the case that the target dataset has much fewer classes than the ones in the source dataset. Aligning the source domain completely with the target domain might result in negative transfer in such cases. Hence, it is important to mitigate the problem of negative transfer by down-weighting the data samples belonging to the outlier source classes, thereby promoting feature alignment only in the shared label spaces (positive transfer) \cite{cao2018partial, kim2020associative, cao2019learning}.\\
\item \textbf{Open-set domain adaptation}: Similar to the issue of partial domain adaptation, open-set DA deals with the issue where the label space is not completely shared between the source and the target domain. However, in this particular case, the target domain contains unknown classes that are not observed in the source domain. This issue also results in the negative transfer; however, identify the unknown class samples in the target dataset are challenging and we need to borrow principles from the open set recognition task in order to automatically down-weight such samples during the domain alignment \cite{panareda2017open, liu2019separate}.\\
\item \textbf{Source-free domain adaptation}: A common assumption in the existing literature is that the samples from the source domain are available during domain adaptation training. However, in real-world scenarios, gaining access to source data might not be practical due to privacy concerns, legal issues, and inefficient data transmission. To this end, we must tackle the problem of source-free domain adaptive object detection, where, there is no access to the source data but only the source trained model. An obvious approach would be based on the pseudo-label based self-training, which involves first generating pseudo-labels on the target and re-using them to supervise the network on the target data. However, it is important to generate highly reliable pseudo-labels by filtering out incorrect labels \cite{kundu2020universal, nelakurthi2018source, li2020free}. \\
\item \textbf{Multi-source domain adaptation}:  All the existing approaches for domain adaptive object detection assume that the labeled training data are sampled from a single domain. This neglects a more practical scenario where training data are collected from multiple sources. For example, consider the case of a self-driving system where source data is available from multiple cities, and the overall goal is to adapt to a new city. Since the source dataset comes from multiple cities, these data points might potentially correspond to multiple sub-distributions. Aligning the target data with all these sub-distributions might not necessarily result in optimal performance. Hence, it would be essential to perform a selective adaptation where the only relevant source samples are considered \cite{zhao2018adversarial, peng2019moment}.\\
\item \textbf{Continuous and test time adaptation}: Existing domain adaptive detection approaches consider the world to be separated into stationary domains. However, this may not be the case in many real-world applications, where samples arise from a continuously evolving underlying process. Examples include videos with gradually changing lighting/weather conditions. Hence, a more practical setting for domain adaptation is to consider the lifelong learning problem of adapting a pre-trained model to dynamically changing environmental conditions. This setting reflects a real-world scenario where we encounter unlabeled images from new target environments that are not observed during training. Hence, it is important to develop strategies to adapt at test time where we have access only to a single or a few instances of target domain data \cite{royer2015classifier, fredericks2014towards, wang2021tent}. Note that these strategies should consider additional constraints such as computationally inexpensive adaptation, restricted access to source domain data, and the issue of catastrophic forgetting, which is a prominent issue that results from continuous learning \cite{hoffman2014continuous, wulfmeier2018incremental, mancini2019adagraph,wu2019ace}. Such constraints are required to adapt to dynamically changing environments.\\
\end{itemize}

\section{Conclusion}
In this paper, we considered the unsupervised domain adaptation of deep object detectors and presented an extensive survey of existing approaches for this task. We have reviewed several
%in detail a total of \textcolor{black}{[fifty]} 
approaches that were published in the last few years. 
We have provided a comprehensive taxonomy of the existing approaches, followed by a detailed analysis of the various methods along with their merits and demerits. We have also provided a detailed discussion on the existing datasets,  evaluation protocols and comprehensive performance comparison.  Finally, we have identified some of the outstanding issues and promising directions to drive future research.

% or


%%%%%%%\appendices
%%%%%%%\section{Proof of the First Zonklar Equation}
%%%%%%%Appendix one text goes here.

%%%%%%%% you can choose not to have a title for an appendix
%%%%%%%% if you want by leaving the argument blank
%%%%%%%\section{}
%%%%%%%Appendix two text goes here.

% % use section* for acknowledgment
% \ifCLASSOPTIONcompsoc
%   % The Computer Society usually uses the plural form
%   \section*{Acknowledgments}
% \else
%   % regular IEEE prefers the singular form
% \fi
% The authors would like to thank......

\ifCLASSOPTIONcaptionsoff
  \newpage
\fi



%\bibliographystyle{IEEEtran}
%%\begin{thebibliography}{1}
%%
%%\bibitem{IEEEhowto:kopka}
%%H.~Kopka and P.~W. Daly, \emph{A Guide to \LaTeX}, 3rd~ed.\hskip 1em plus
%%  0.5em minus 0.4em\relax Harlow, England: Addison-Wesley, 1999.
%%
%%\end{thebibliography}

\bibliographystyle{IEEEtran}
\bibliography{da_detect_refs}


%%%%%%%%%%%%%%%%%%%%%\begin{IEEEbiography}{John Doe}
%%%%%%%%%%%%%%%%%%%%%Biography goes here.
%%%%%%%%%%%%%%%%%%%%%\end{IEEEbiography}

\begin{IEEEbiographynophoto}{Poojan Oza}
is  a  PhD  student  in the Department Of Electrical \& Computer Engineering at The Johns Hopkins University. He graduated from IIIT-Delhi with a  Master's degree in Electronics and Computer Engineering. His research interests include deep learning based one-class methods, anomaly/novelty detection, open-set recognition, domain adaptation and object detection.
\end{IEEEbiographynophoto}

\begin{IEEEbiographynophoto}{Vishwanath A. Sindagi}
is  a  PhD  student  in the Department Of Electrical \& Computer Engineering at The Johns Hopkins University. Prior to joining Johns Hopkins University, he worked for Samsung R\&D Institute-Bangalore. He graduated from IIIT-Bangalore with a  Master's degree in Information Technology. His research interests include deep learning based crowd analytics, object detection, applications of generative modeling, domain adaptation and low-level vision.
\end{IEEEbiographynophoto}

\begin{IEEEbiographynophoto}{Vibashan VS}
is  a  PhD  student  in the Department Of Electrical \& Computer Engineering at The Johns Hopkins University. He graduated from NIT-Tiruchirappalli with a  Bachelor's degree in Instrumentation and Control. His research interests include deep learning based object detection, domain adaptation.
\end{IEEEbiographynophoto}

\begin{IEEEbiographynophoto}{Vishal M. Patel}
is an Assistant Professor in the Department of Electrical and Computer Engineering (ECE) at Johns Hopkins University. Prior to joining Hopkins, he was an A. Walter Tyson Assistant Professor in the Department of ECE at Rutgers University and a member of the research faculty at the University of Maryland Institute for Advanced Computer Studies (UMIACS). His current research interests include signal processing, computer vision, and pattern recognition with applications in bio-metrics and imaging. He has received a number of awards including the 2016 ONR Young Investigator Award, the 2016 Jimmy Lin Award for Invention, A. Walter Tyson Assistant Professorship Award, Best Paper Award at IEEE AVSS 2017, Best Paper Award at IEEE BTAS2015, Honorable Mention Paper Award at IAPR ICB 2018, two Best Student Paper Awards at IAPR ICPR 2018, and Best Poster Awards at BTAS 2015 and 2016. He is an Associate Editor of the IEEE Signal Processing Magazine, IEEE Biometrics Compendium, Pattern Recognition Journal, and serves on the Machine Learning for Signal Processing Technical Committee of the IEEE Signal Processing Society. He is serving as the Vice President (Conferences) of the IEEE Biometrics Council. He is a member of Eta Kappa Nu, Pi Mu Epsilon, and Phi Beta Kappa. 
\end{IEEEbiographynophoto}

%\newpage

%%%%%%%%%\begin{IEEEbiographynophoto}{Jane Doe}
%%%%%%%%%Biography text here.
%%%%%%%%%\end{IEEEbiographynophoto}

%\vfill


\end{document}